\documentclass{article}
\usepackage{arxiv}

\usepackage{graphicx} 
\usepackage{tabularx} 
\usepackage[table]{xcolor}
\usepackage[compact]{titlesec}
\usepackage[hidelinks]{hyperref}
\usepackage{amsmath}
\usepackage{amssymb}
\usepackage{mathtools}
\usepackage{amsthm}
\usepackage{algorithm}
\usepackage{comment}
\usepackage{float}
\usepackage{stfloats}

\usepackage{xcolor}

\usepackage{booktabs} 
\usepackage{multirow}
\usepackage{makecell}
\usepackage{subcaption}
\usepackage[capitalize,noabbrev]{cleveref}
\usepackage{threeparttable}
\usepackage{listings}
\usepackage{tcolorbox}

\newcommand{\subject}[1]{\vspace{3pt}\noindent\textbf{#1}}

\newcommand{\query}{x_{\text{query}}} 
\newcommand{\queryresp}{y_{\text{query}}} 
\newcommand{\iclmodel}{\mathcal{M}} 

\newcommand{\bnm}{\begin{newmath}}
\newcommand{\enm}{\end{newmath}}
\newcommand{\bne}{\begin{newequation}}
\newcommand{\ene}{\end{newequation}}

\newenvironment{newmath}{\begin{displaymath}%
\setlength{\abovedisplayskip}{4pt}%
\setlength{\belowdisplayskip}{4pt}%
\setlength{\abovedisplayshortskip}{6pt}%
\setlength{\belowdisplayshortskip}{6pt} }{\end{displaymath}}

\newenvironment{newequation}{\begin{equation}%
\setlength{\abovedisplayskip}{4pt}%
\setlength{\belowdisplayskip}{4pt}%
\setlength{\abovedisplayshortskip}{6pt}%
\setlength{\belowdisplayshortskip}{6pt} }{\end{equation}}

\newlength{\saveparindent}
\newlength{\saveparskip}
\newcounter{ctr}

\renewcommand{\paragraph}[1]{\vspace*{6pt}\noindent\textbf{#1}}

\usepackage[normalem]{ulem}

\usepackage{cite}

\title{Beyond One-Size-Fits-All: Personalized Harmful Content Detection with In-Context Learning} 

\author{
  Rufan Zhang \\
  University of Science and \\Technology of China \And Lin Zhang\\ University of Science and \\Technology of China \And
  Xianghang Mi \\ University of Science and \\Technology of China}
\date{}

\begin{document}

\thispagestyle{plain}
\pagestyle{plain}

\maketitle

The proliferation of harmful online content—e.g., toxicity, spam, and negative sentiment—demands robust and adaptable moderation systems. However, prevailing moderation systems are centralized and task-specific, offering limited transparency and neglecting diverse user preferences—an approach ill-suited for privacy-sensitive or decentralized environments. We propose a novel framework that leverages in-context learning (ICL) with foundation models to unify the detection of toxicity, spam, and negative sentiment across binary, multi-class, and multi-label settings. Crucially, our approach enables lightweight personalization, allowing users to easily block new categories, unblock existing ones, or extend detection to semantic variations through simple prompt-based interventions—all without model retraining.
Extensive experiments on public benchmarks (TextDetox, UCI SMS, SST2) and a new, annotated Mastodon dataset reveal that: (i) foundation models achieve strong cross-task generalization, often matching or surpassing task-specific fine-tuned models; (ii) effective personalization is achievable with as few as one user-provided example or definition; and (iii) augmenting prompts with label definitions or rationales significantly enhances robustness to noisy, real-world data.
Our work demonstrates a definitive shift beyond one-size-fits-all moderation, establishing ICL as a practical, privacy-preserving, and highly adaptable pathway for the next generation of user-centric content safety systems.
To foster reproducibility and facilitate future research, we publicly release our code on GitHub\footnote{\url{https://github.com/ChaseSecurity/personalizable\_harmful\_content\_detection}} and the annotated Mastodon dataset on Hugging Face\footnote{\url{https://huggingface.co/datasets/ChaseLabs/Harmful-Texts-On-Mastodon}}.

\section{Introduction}
\label{sec:intro}
The integrity of online ecosystems is persistently challenged by a spectrum of harmful content, from overtly toxic language and manipulative spam to subjectively negative speech. The societal impact of such content is severe, such as corroding healthy discourse, inflicting psychological harm, facilitating the spread of misinformation, and fracturing digital communities~\cite{citron2014hate, schmidt2017survey, fortuna2018survey, feldman2013techniques, almuhimedi2013tweets}. As a result, harmful content detection has become a critical task for both academic research and industry applications. Traditional supervised approaches, such as deep neural classifiers~\cite{badjatiya2017deep, roy2020deep, raffel2020sst2} have achieved high performance on harmful content benchmarks. However, these methods rely on large-scale annotated datasets that are domain- and task-specific. Once deployed, without costly retraining, they struggle to adapt to new categories of harmfulness, user preferences, or concept shifts.


The recent ascendancy of large language models (LLMs) offers a promising alternative through in-context learning (ICL), which adapts to new tasks dynamically using task descriptions and few-shot demonstrations~\cite{brown2020language,wei2022emergent,chung2022scaling}. While preliminary studies have explored ICL for tasks like toxicity detection~\cite{amin2023wideevaluationchatgptaffective}, they remain confined to narrow, single-task settings on curated benchmarks. Consequently, three critical questions remain largely unaddressed: \textbf{(1) Can ICL generalize effectively across heterogeneous harmful content tasks?} \textbf{(2) Can it be efficiently personalized to align with individual user preferences?} and \textbf{(3) How robust is ICL when faced with the noisy, ambiguous, and multi-faceted nature of real-world "wild" data?}

In this paper, we present the first systematic study of ICL for \textbf{personalizable and cross-task harmful content detection}. We unify three subtasks---binary harmfulness detection, multi-class categorization (\texttt{toxic}, \texttt{spam}, \texttt{negative}), and multi-label classification---into a single evaluation framework, enabling us to assess cross-task generalization. Beyond these canonical settings, we introduce three personalization scenarios inspired by real-world moderation: (i) \textit{blocking new categories}, (ii) \textit{unblocking existing categories}, and (iii) \textit{blocking perturbed variations}. To further validate robustness, we extend evaluation to \textit{wild data} collected from Mastodon, where content distributions are noisier and more ambiguous.

Specifically, we implement our framework by systematically comparing multiple foundation model families (e.g., LLaMA-3, Mistral, Qwen2) under zero-, few-, and many-shot ICL settings, combined with diverse demonstration retrieval strategies (random, lexical retrieval, and semantic retrieval) and prompt designs (with/without definitions or decision rationales). On benchmark datasets, our unified ICL formulation achieves strong cross-task generalization: for instance, in binary harmful text classification, LLaMA-3-8B with random 128-shot prompts attains 97.1\% F1 on TextDetox Toxicity, 97.0\% on UCI SMS Spam, and 95.9\% on SST2 sentiment, competitive with or surpassing task-specific baselines. In multi-task binary and multi-task multi-class tasks, LLaMA-3-8B with 196 shots achieves 91.8\% F1 and 93.8\% Weighted avg F1, respectively. 

For personalization, we design prompt-level interventions that allow users to modify filtering categories with minimal effort. Adding only a few user-provided examples or definitions enables the model to reliably block or unblock categories.

Finally, we evaluate robustness in a new Mastodon dataset annotated under binary, multi-class, and multi-label schemes. Results reveal a significant gap between curated benchmarks and real-world noisy data: for instance, in multi-task binary, LLaMA-3-8B with random 48-shot prompts' F1 drops from 88.7\% on benchmarks to 79.5\% in wild settings. However, replacing demos with wild data and incorporating reason-augmented prompts mitigates performance loss and yields more interpretable outputs(87.5\% F1).

Together, these advances move beyond the one-size-fits-all paradigm and highlight ICL as a practical and privacy-preserving solution for user-centric and cross-task harmful content detection. our main contributions are threefold:
\begin{itemize}
    \item \textbf{Unified Multi-Task Formulation:} We introduce a comprehensive evaluation framework that unifies binary, multi-class, and multi-label detection of toxicity, spam, and negative sentiment, demonstrating strong cross-task generalization by foundation models.
    \item \textbf{Lightweight Personalization Mechanisms:} We design and evaluate novel prompt-based interventions for users to effortlessly block new categories, unblock existing ones, and defend against adversarial variations, requiring minimal supervision.
    \item \textbf{Wild Data Evaluation and Robustness Analysis:} We construct and release a new annotated dataset from Mastodon, enabling the first large-scale evaluation of personalized detection in a decentralized context. Our analysis reveals the limitations of benchmark-only evaluation and identifies rationale-augmented prompts as a key technique for improving robustness.
\end{itemize}

The remainder of this paper is organized as follows: We review preliminaries in Section~\ref{sec:preliminary} and detail our experimental setup in Section~\ref{sec:tasks}. Sections~\ref{sec:single-task_binary_icl}, \ref{sec:multi-task_binary} and \ref{sec:multi-task_multi} present our results on single-task, multi-task binary, and multi-task multi-class detection, respectively. Section~\ref{sec:personalization} explores personalization scenarios, and Section~\ref{sec:wild_data} evaluates performance on wild data. We discuss implications and limitations in Section~\ref{sec:discussion} and conclude in Section~\ref{sec:conclusion}.

\section{Preliminaries}
\label{sec:preliminary}

\subsection{Defining Harmful Texts}
We define \emph{harmful texts} as content that can undermine individual well-being or platform integrity, typically manifested through offensive, manipulative, or otherwise negative language.

Toxicity encompasses language involving harassment, hate speech, profanity, or offensive expressions. Its societal impact is well-documented, with benchmarks like the Wikipedia Toxic Comments dataset~\cite{jigsaw-toxic-comment-classification-challenge} underlining its prevalence and importance for online communities.
Spam encompasses unsolicited or manipulative content, often characterized by promotional intent, repetitive links, or coordinated amplification. Classical spam detection research dates back to early work on email and web spam~\cite{alam2008spam,wang2010detecting}, and continues to be relevant for social media platforms where adversarial actors exploit open publishing mechanisms. Negative sentiment, while not overtly abusive, captures emotionally harmful expressions such as anger, frustration, or despair. Although often excluded from traditional harmfulness definitions, negative sentiment has psychosocial consequences for online well-being~\cite{medhat2014sentiment}. Widely adopted sentiment analysis datasets such as SST-2~\cite{socher2013recursive} provide a standardized basis for studying this dimension.

Together, these categories capture three major axes of online harm: abusive harms (toxicity), economic and functional disruption (spam), and psychosocial harms (negative sentiment). This tripartite scheme allows us to ground our experiments in well-established benchmarks while reflecting the heterogeneity of real-world harmful discourse.

\subsection{Limitations of Prior Work}
Traditional approaches to harmful text detection relied heavily on keyword-based filters or classical machine learning classifiers trained on handcrafted features~\cite{schmidt2017survey}. While straightforward, these methods suffer from low recall, brittleness to adversarial rephrasing, and poor adaptability to domain shifts. The advent of deep learning introduced neural text classifiers trained on large annotated datasets, achieving strong performance in within-domain settings~\cite{badjatiya2017deep,zhang2018detecting}. However, subsequent studies revealed that such models generalize poorly across domains or platforms~\cite{arango2019hate}, limiting their practical reliability.

Recent advances in large language models (LLMs) have spurred interest in adapting them for harmful content detection. Fine-tuned LLMs have shown strong results in toxicity~\cite{gururangan2020don}, spam~\cite{labonne2023spamt5benchmarkinglargelanguage} and sentiment detection~\cite{raffel2020sst2}, while prompting-based methods reduce the reliance on large labeled datasets~\cite{amin2023wideevaluationchatgptaffective, He2024sptoxic, zhang2024toxic}. Nevertheless, most existing works focus on narrow problem settings, such as binary harmful detection within a single task, while overlooking more comprehensive scenarios that span multiple tasks and involve binary, multi-class, or multi-label classification. Moreover, prior studies typically optimize for within-domain benchmarks, without explicitly testing generalization to real-world “wild data,” which is noisier, more diverse, and less neatly separable into canonical categories.

\subsection{In-Context Learning}
\label{subsec:icl}

In-context learning (ICL) is a technique that enables large language models to adapt to new tasks by conditioning on a task description and a small set of demonstrations, without the need for parameter updates. Mathematically, given a task description $t$ and a set of $k$ demonstrations $D_k=\{(x_i,y_i)\}_{i=1}^k$, where $x_i$ represents the input and $y_i$ is the corresponding label, ICL predicts the label $y_{query}$ for a query $x_{query}$ as follows:
\[
y_{query} = M(t, D_k, x_{query}),
\]
where $M$ denotes the large language model. This approach allows for rapid task adaptation and showcases the model's ability to learn from examples provided within the context.

A growing body of research highlights that in-context learning (ICL) performance is highly sensitive to design choices across multiple dimensions. Model choice~\cite{chung2022scaling,wei2022emergent,chowdhery2022palm} is one important factor: instruction-tuned and larger LLMs consistently exhibit stronger generalization and more robust few-shot behavior. Retrieval strategy~\cite{luo2023dricl, rubin2022learning} also matters, as demonstrations can be selected randomly, by lexical similarity (e.g., BM25~\cite{robertson2009probabilistic}), or by semantic similarity (e.g., Sentence-Transformers~\cite{reimers2019sentencebert}), with each approach shaping the inductive bias of ICL. The number of demonstrations~\cite{liu2021makes,min2022rethinking} is another critical variable: while increasing the number of shots generally improves stability, studies show diminishing returns and even saturation beyond certain thresholds. 

Equally important is the design of the task description and output format. Prior work demonstrates that subtle variations in prompt phrasing can drastically alter performance, and structured outputs help models adhere more reliably to task requirements~\cite{kojima2023largelanguagemodelszeroshot,ye2023compositional}. Finally, the addition of rationales---natural language explanations paired with labels---has been shown to enhance reasoning and reduce error propagation~\cite{lampinen2022can,wei2022chain, mishra2022cross}, providing interpretive signals that guide the model beyond surface-level pattern recognition.

Our work builds on these insights by systematically examining how these dimensions shape ICL for harmful content detection. Unlike prior research, we unify evaluation across binary, multi-class, and multi-label formulations; explore personalization strategies via three common personalization scenarios; and extend evaluation to Mastodon wild data, where domain shift, noise, and label overlap provide a realistic stress test for ICL robustness.

\section{Experimental Setup: Tasks, Datasets, and In-Context Learning Configuration}
\label{sec:tasks}
Building upon the foundation laid in Section~\ref{sec:preliminary}, we now detail our experimental setup for evaluating multi-task and personalized harmful content detection via in-context learning (ICL). This section is organized as follows: we first describe the datasets (Section~\ref{subsec:datasets}), then formally define the learning tasks (Section~\ref{subsec:tasks}), and finally specify the ICL configurations and parameters (Section~\ref{subsec:icl-setting}).
\subsection{Datasets}
\label{subsec:datasets}
\begin{table}
    \centering
    \caption{Overview of datasets.}
    \label{tab:overview_of_datasets}
    \begin{tabular}{cccccc}
        \toprule
        Task & Dataset & Language &Label & Train & Test \\
        \midrule
        \multirow{2}{*}{Toxicity} & \multirow{2}{*}{Textdetox} & \multirow{2}{*}{English} & Toxic & 2,000 & 500 \\
        & & & Non-toxic & 2,000 & 500 \\
        \midrule
        \multirow{2}{*}{Spam} & \multirow{2}{*}{UCI} & \multirow{2}{*}{English} & Spam & 586 & 161 \\
        & & & Ham & 3,873 & 954 \\
        \midrule
        \multirow{2}{*}{Sentiment} & \multirow{2}{*}{SST2} & \multirow{2}{*}{English} & Negative & 29,780 & 428\\
        & & & Positive & 37,569 & 444 \\
        \bottomrule
    \end{tabular}
\end{table}


Building upon the categories of harmful content—toxicity, spam, and negative sentiment—introduced in Section~\ref{sec:preliminary}, we have meticulously curated representative, publicly available datasets to facilitate model development and evaluation. These datasets form the cornerstone for constructing and benchmarking our detection models. For datasets with predefined train/test splits, we adhere to these standard divisions to ensure comparability with prior research. For datasets lacking official splits, we employ an 80/20 random stratified split, allocating 80\% of the data for training (used to select demonstration examples for prompting) and reserving 20\% for testing. This approach ensures a balanced and representative evaluation framework.

Table~\ref{tab:overview_of_datasets} provides a summary of the selected datasets. Next, we delve into the specifics of each dataset, highlighting their collection methodologies, quality attributes, and the rationale for their selection over alternative datasets.

\subject{TextDetox~\cite{dementieva2024overview}.} TextDetox is a meticulously curated dataset tailored for binary toxicity classification across multiple natural languages. As of this writing, it supports 15 languages. For each languages, the dataset comprises 5,000 samples per language, evenly distributed between toxic and non-toxic content. TextDetox is constructed by carefully selecting high-quality subsets from existing toxicity corpora, ensuring its utility as a reliable resource for training and evaluating toxicity detection models. 

In our experiments, we utilize only the English subset of TextDetox, which is derived from the Jigsaw Toxic Comment Classification Challenge (Jigsaw~\cite{jigsaw2018}), a widely recognized benchmark for toxicity detection. The Jigsaw dataset contains over 200,000 comments, manually annotated by multiple human raters for six toxicity categories: toxic, severe toxic, obscene, threat, insult, and identity hate. Its large scale, multi-label structure, and real-world relevance make it an invaluable resource for developing and benchmarking toxicity detection models. Furthermore, its extensive adoption in academic research \cite{deepak2021toxic, dale-etal-2021-text, spirosv2018cnntoxic} underscores its status as a standard benchmark for toxicity classification tasks.

The decision to use TextDetox instead of directly employing Jigsaw is motivated by two key considerations. First, TextDetox enhances data quality through rigorous manual screening when selecting samples from Jigsaw, ensuring a cleaner and more reliable dataset. Second, the inclusion of non-English samples in TextDetox provides a valuable opportunity to evaluate our detection models on out-of-domain datasets, thereby enabling a more comprehensive assessment of their generalization capabilities in the future. This dual advantage makes TextDetox an ideal choice for our experiments.

\subject{UCI~\cite{sms_spam_collection_228}.} The UCI SMS Spam Collection is a widely recognized dataset curated for SMS spam detection research. It comprises 5,574 English SMS messages, each annotated as either spam or ham (legitimate). 
The dataset aggregates messages from multiple reputable sources, including the Grumbletext website (425 manually extracted spam messages from a UK forum), the NUS SMS Corpus (3,375 randomly selected legitimate messages from a large collection at the National University of Singapore), Caroline Tag's PhD thesis (450 legitimate messages), and the SMS Spam Corpus v.0.1 Big (1,002 legitimate and 322 spam messages). This diverse origin ensures that the dataset captures a wide range of real-world texting behaviors and spam types, enhancing its representativeness and utility for spam detection research.

The manual selection process employed in curating the dataset further ensures high data quality, making it a reliable benchmark for evaluating spam detection models. Since its public release in 2012, the dataset has been extensively used in academic research, serving as a standard benchmark for SMS spam detection tasks. Its adoption in numerous studies\cite{almeida2012uci, sjarif2019sms, roy2020deep, abayomi2022deep, liu2021spam} underscores its significance in the field and ensures that findings based on this dataset are comparable and reproducible across different works. 



\subject{SST2~\cite{socher2013recursive}.} The Stanford Sentiment Treebank (SST2) is a benchmark dataset widely utilized for binary sentiment classification. It comprises 70,042 movie review sentences annotated as either positive or negative, with 67,349 samples designated for training, 872 for validation, and 1,821 for testing. In our experiments, we use the original training split for generating demonstration examples and the validation split for testing. This ensures that the test set size for sentiment analysis aligns with the other two categories, thereby avoiding over-representation when calculating the performance of multi-task detection models.

The dataset originates from 11,855 movie reviews, with 215,154 phrases within their parse trees manually annotated with sentiment labels. This granular annotation enables models to capture the compositional nature of sentiment within sentence structures, making SST2 a high-quality resource for sentiment analysis. Compared to other sentiment datasets, SST2 distinguishes itself through its detailed phrase-level annotations and syntactic structure information, which facilitate the learning of complex linguistic patterns. Furthermore, as part of the GLUE benchmark, SST2 has been extensively adopted for evaluating NLP models\cite{raffel2020sst2, jiang-etal-2020-smart, Lan2020albert, muppet2021sst2}, ensuring that research findings are both comparable and reproducible across studies.

To further assess the generalization capabilities of our models, we incorporate out-of-domain test data sourced from Mastodon, a widely recognized federated social network platform. This data corresponds to the three categories of harmful content under investigation. The selection of this dataset aims to rigorously challenge the models' robustness and adaptability when confronted with unseen data distributions. Comprehensive details regarding the dataset and the associated evaluation methodologies will be presented in Section~\ref{sec:wild_data}.

\subsection{Learning Tasks}
\label{subsec:tasks}
Building upon the datasets introduced earlier, we now provide a formal definition of the learning tasks designed to enable multi-task and personalized harmful content detection using in-context learning (ICL). The evaluation of these tasks will be detailed in subsequent sections.

Let $\mathcal{X}$ denote the input space, $\mathcal{Y}$ the label space, and $t$ a natural-language task instruction. We define $D_k^{r} = \{ (x_1, y_1), (x_2, y_2), \dots, (x_k, y_k) \} \subseteq D$ as a set of $k$ demonstration examples retrieved from dataset $D$ under retrieval strategy $r$. Let $\query \in \mathcal{X}$ denote a new input under evaluation (i.e., the query sample), and let $\iclmodel$ denote the foundation model employed for in-context learning. The predicted output $\queryresp$ is defined as:
\[
    \queryresp = \iclmodel(t,\,D_k^{r},\, \query).
\]
This formulation represents the general ICL process. For classification tasks, we typically employ \textbf{greedy decoding} to select the most probable label from the model's vocabulary that matches a label in \(\mathcal{Y}\). This can be formalized as:
\[ y_{\text{query}} = \operatorname*{arg\,max}_{y \in \mathcal{Y}} P_{\mathcal{M}}\left(y \mid t,\, D_{k}^{r},\, x_{\text{query}}\right), \]
where \(P_{\mathcal{M}}\) represents the model's conditional probability distribution over its vocabulary.

\subject{Remark.} In our experiments, we primarily adopt demonstrations of the form $(x_i, y_i)$, i.e., input--label pairs. 
Only in the wild data experiments do we extend demonstrations with reasons $(x_i, y_i, e_i)$, where $e_i$ provides a textual rationale for the assigned label.

\subject{Single-Task Harmful Content Classification.} Single-task harmful content classification independently detect each category of harmful content. In this approach, the label space $\mathcal{Y}$ corresponds to the specific harm category under consideration (e.g., \textit{spam} and \textit{ham} for spam detection), and the demonstration set $D_k^{r}$ is exclusively sampled from the relevant training dataset (e.g., toxicity training datasets for toxicity detection). These independent ICL detectors are collectively referred to as single-task harmful content classifiers.

\subject{Multi-Task Harmful Content Classification.} In contrast to single-task classification, multi-task harmful content classification aims to simultaneously detect multiple categories of harmful content. Here, the label space $\mathcal{Y}$ is defined as the union of all harm categories under consideration. The demonstration set $D_k^{r}$ is sampled from all relevant training datasets, encompassing examples from each harm category. Depending on the structure of the label space $\mathcal{Y}$, this task can be further divided into two subcategories: multi-task \textit{binary} ICL and multi-task \textit{multi-class} ICL.

In multi-task binary ICL, the model distinguishes harmful content from benign content, with the label space defined as $\mathcal{Y} = \{\textit{harmful}, \textit{benign}\}$, where $\textit{harmful}$ represents harmful(toxic, spam, negative) content and $\textit{benign}$ represents non-harmful content. Conversely, in multi-task multi-class ICL, the model assigns harmful content to one of the specific harm categories, with the label space defined as $\mathcal{Y} = \{\textit{toxic}, \textit{spam}, \textit{negative}, \textit{benign}\}$.

As mentioned in preliminaries(section~\ref{sec:preliminary}), the performance of an ICL classifier is influenced by various hyperparameters, including but not limited to the number and retrieval strategy of demonstration examples ($k$), the task instruction ($t$), and the choice of foundation model ($\iclmodel$). In the following section, we provide a comprehensive overview of the ICL hyperparameters to be systematically explored for each harmful content detection task.

\subsection{General ICL Settings}
\label{subsec:icl-setting}

\subject{Foundation Models.} In our experiments, we evaluate three state-of-the-art open-weight large language model (LLM) families: Llama~\cite{touvron2023llamaopenefficientfoundation}, Mistral~\cite{jiang2023mistral7b}, and Qwen~\cite{bai2023qwentechnicalreport}. The selection of specific model configurations (e.g., instruction-tuned variants, model sizes and versions) is guided by multiple criteria, including performance on diverse NLP benchmarks, support for long-context inferences, and computational efficiency on our available hardware resources. 

For our study, we adopt the latest instruction-tuned versions available as of July 2024: Llama-3.1-8B-Instruct\cite{meta_llama_3_1_8B_Instruct}, Mistral-7B-Instruct-v0.3\cite{mistral_7B_Instruct_v0_3}, and Qwen2-7B-Instruct\cite{qwen2_7B_Instruct}. These models support maximum context lengths of 128K, 32K, and 128K tokens, respectively. To ensure compatibility across all tasks, we constrain the prompt length to remain within 32K tokens. This selection balances cutting-edge performance with practical considerations for efficient evaluation on our computing infrastructure. 

As of July 2025, newer releases of the Llama, Mistral, and Qwen families have become available. However, these models are either not well aligned with the requirements of our current evaluation setup or show no significant performance gains on our tasks compared to the versions we adopt in this study. We therefore retain the above models as our primary evaluation suite. Further empirical details are provided in \ref{appendix:model_selection}.




\subject{Prompt Template.} 
As illustrated in Figure~\ref{fig:example_prompt} and discussed in Section~\ref{subsec:tasks}, our prompt consists of three components: the task description $t$, the demonstration set $D_k^{r}$, and the query input $x_{\text{query}}$. 
Varying the formulation of $t$ and $D_k^{r}$ plays a critical role in shaping ICL performance.

\subject{Demonstration-Relevant Parameters.} 
We systematically study demonstration-related hyperparameters. 
First, we vary the number of demonstrations $k$ (e.g., $ k \in \{0,1,2,4,8,16,32,64,128\}$ in single-task binary tasks), thereby covering the full spectrum from zero-shot to many-shot settings. 
Unless otherwise noted, demonstrations are balanced across all classes in $\mathcal{Y}$.

Second, we evaluate the effect of different retrieval strategies for selecting $D_k^{r}$. 
We consider five methods:  
\textbf{Random}, which uniformly samples an equal number of examples per label from the training set, averaging over three random seeds for robustness;  
\textbf{Semantic}, which retrieves semantically similar examples for each query using dense sentence embeddings from \texttt{sentence-transformers}~\cite{reimers2019sentencebert}, implemented with the \texttt{retriv} toolkit\footnote{\url{https://github.com/AmenRa/retriv}};  
\textbf{Balanced Semantic}, which combines semantic similarity with label balancing;  
\textbf{Lexical}, which selects examples based on surface-level overlap using BM25~\cite{robertson2009probabilistic}, also implemented in \texttt{retriv}; and  
\textbf{Balanced Lexical}, which enforces class balance on top of BM25 retrieval. 

Finally, we extend demonstrations beyond input–label pairs $(x_i,y_i)$ to \textbf{reason-augmented demonstrations} $(x_i, y_i, e_i)$, where $e_i$ is a natural-language explanation of why $x_i$ belongs to class $y_i$. 
In our study, such reason-augmented demonstrations are employed only in wild data experiments, where interpretability and robustness to noisy annotations are most critical.

\subject{Task Description.} 
We design two levels of task descriptions. 
\textbf{Level 1} provides a concise statement of the task and its label space, while \textbf{Level 2} augments this with detailed definitions for each label. 
This design allows us to evaluate whether richer task information improves ICL performance. 
Unless otherwise stated, Level~1 descriptions are used.

\subject{Inference Setting.} 
For efficient deployment of foundation models, we employ \textit{vLLM}~\cite{kwon2023efficient}, a high-performance inference engine optimized for large-scale LLMs. 
We adopt GuidedDecodingParams to constrain models to \textbf{structured outputs}, i.e., predictions restricted to the label space without generating irrelevant natural-language content. 
To ensure deterministic and reproducible results, we set the temperature to zero and disable stochastic sampling by configuring top-p to 1 and top-k to -1. 
Given the cumulative logits for candidate labels, we normalize them with a softmax function to obtain prediction probabilities.

\begin{figure}[ht]
    \centering
    \begin{lstlisting}
<Task Description>
==
Query: <sentence 1>
Answer: <label>
==
Query: <sentence 2>
Answer: <label>
==
...
==
Query: <query text>
Answer:
    \end{lstlisting}
    \caption{Prompt template for in-context learning (ICL).}
    \label{fig:example_prompt}
\end{figure}

\section{Single-Task ICL}
\label{sec:single-task_binary_icl}

We begin our evaluation by establishing strong single-task baselines. This allows us to assess the fundamental capability of ICL-equipped foundation models for each harm category independently, before investigating more complex multi-task and personalized settings.

\subsection{Experimental Setup}

\subject{Models.} In single-task binary classification scenarios, we assess the performance of three instruction-tuned foundation models: \textit{Llama-3.1-8B-Instruct}, \textit{Mistral-7B-Instruct-v0.3}, and \textit{Qwen2-7B-Instruct}. Formally, let $\mathcal{M}$ denote the set of models under evaluation, where $\mathcal{M} = \{\text{\textit{Llama-3.1-8B-Instruct}}, \:\text{\textit{Mistral-7B-Instruct-v0.3}},\:\text{\textit{Qwen2-7B-Instruct}}\}$.

\subject{Datasets.} 
As summarized in Table~\ref{tab:overview_of_datasets}, each dataset $D$ consists of training data $D_{\text{train}}$ and testing data $D_{\text{test}}$. 
Demonstrations $D_k^{r} \subseteq D_{\text{train}}$ are retrieved to construct prompts, while $D_{\text{test}}$ is directly used for evaluation. 
Accordingly, the input space in each single-task setting is defined as $\mathcal{X} = D_{\text{test}}$.

\subject{Label Names.} 
The label spaces are defined as follows:  
\textbf{Toxicity}: $\mathcal{Y} = \{\textit{toxic}, \textit{benign}\}$;  
\textbf{Spam}: $\mathcal{Y} = \{\textit{spam}, \textit{ham}\}$;  
\textbf{Sentiment}: $\mathcal{Y} = \{\textit{negative}, \textit{positive}\}$.

\subject{Number of Shots.} 
We experiment with $k \in \{0, 2, 4, 8, 16, 32, 64, 128\}$ demonstrations, selected according to retrieval strategy $r$, to analyze the impact of shot number.

\subject{Retrieval Strategies.} 
As described in Section~\ref{subsec:icl-setting}, we evaluate five retrieval strategies: \textit{Random}, \textit{Semantic}, \textit{Balanced Semantic}, \textit{Lexical}, and \textit{Balanced Lexical}.

\subject{Prompt Design.} 
Figure~\ref{fig:description_of_textdetox} illustrates the task description for Toxicity. 
Descriptions for Spam and Sentiment follow the same format and are provided in \ref{appendix:task_description}.

\subject{Metrics.} Typical classification Metrics are adopted, including recall, precision, F1-score, and accuracy. Also, as fake alarm rate is a critical metric in content moderation, we also measure the metric of false positive rate. 

\subject{Baselines.} 
Prior studies~\cite{devlin2018bert, koroteev2021bert, gonzalez2020comparing} have demonstrated that fine-tuned BERT models remain strong baselines for a variety of NLP classification tasks. In particular, BERT-based approaches have achieved competitive or even state-of-the-art results on the datasets we use in this work: TextDetox for toxicity detection~\cite{Detoxify}, SST-2 for sentiment classification~\cite{munikar2019finegrainedsentimentclassificationusing}, and the UCI SMS Spam Collection for spam detection~\cite{li2024spamdamprivacypreservingadversaryresistantsms}. To provide a fair and informative comparison, we adopt \texttt{BERT-base-uncased} as our supervised baseline. The model is fine-tuned separately on each dataset using its respective training set and evaluated on the same test sets as used in our ICL experiments. This allows us to directly contrast the performance of traditional fine-tuning against in-context learning under identical data splits, thereby highlighting the efficiency and adaptability of ICL relative to established supervised methods.

\begin{figure}
    \centering
    \begin{lstlisting}
#### Positioning
- Assistant Name: Toxic text Detector
- Primary Task: Classify whether a given text is toxic or benign.

#### Capabilities
- Message Analysis: Accurately analyze the content and characteristics of text.
- Toxicity Classification: Determine whether a text is 'toxic' or 'benign'.

#### Knowledge Base
- Message Categories:
  - toxic
  - benign

#### Instructions for Use
- Input: A piece of text.
- Output: Only output the category of the text ('toxic' or 'benign'), without any additional explanation.
    \end{lstlisting}
    \caption{The task description of single-task ICL (Toxicity).}
    \label{fig:description_of_textdetox}
\end{figure}

\subsection{Results}

\begin{figure}
    \centering
    \begin{subfigure}{0.32\textwidth}
        \centering
        \includegraphics[width=\textwidth]{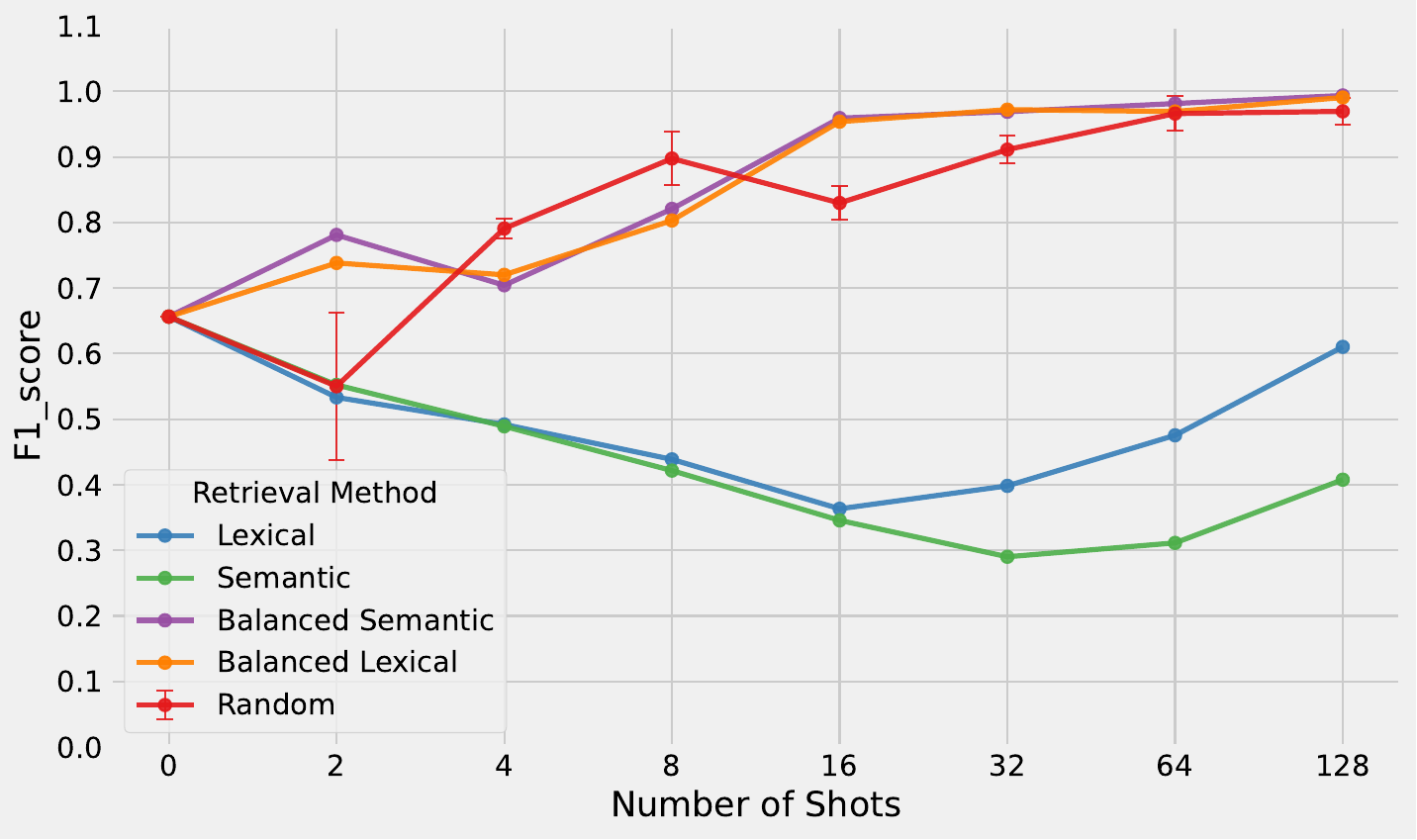}
        \caption{Llama}
    \end{subfigure}
    \begin{subfigure}{0.32\textwidth}
        \centering
        \includegraphics[width=\textwidth]{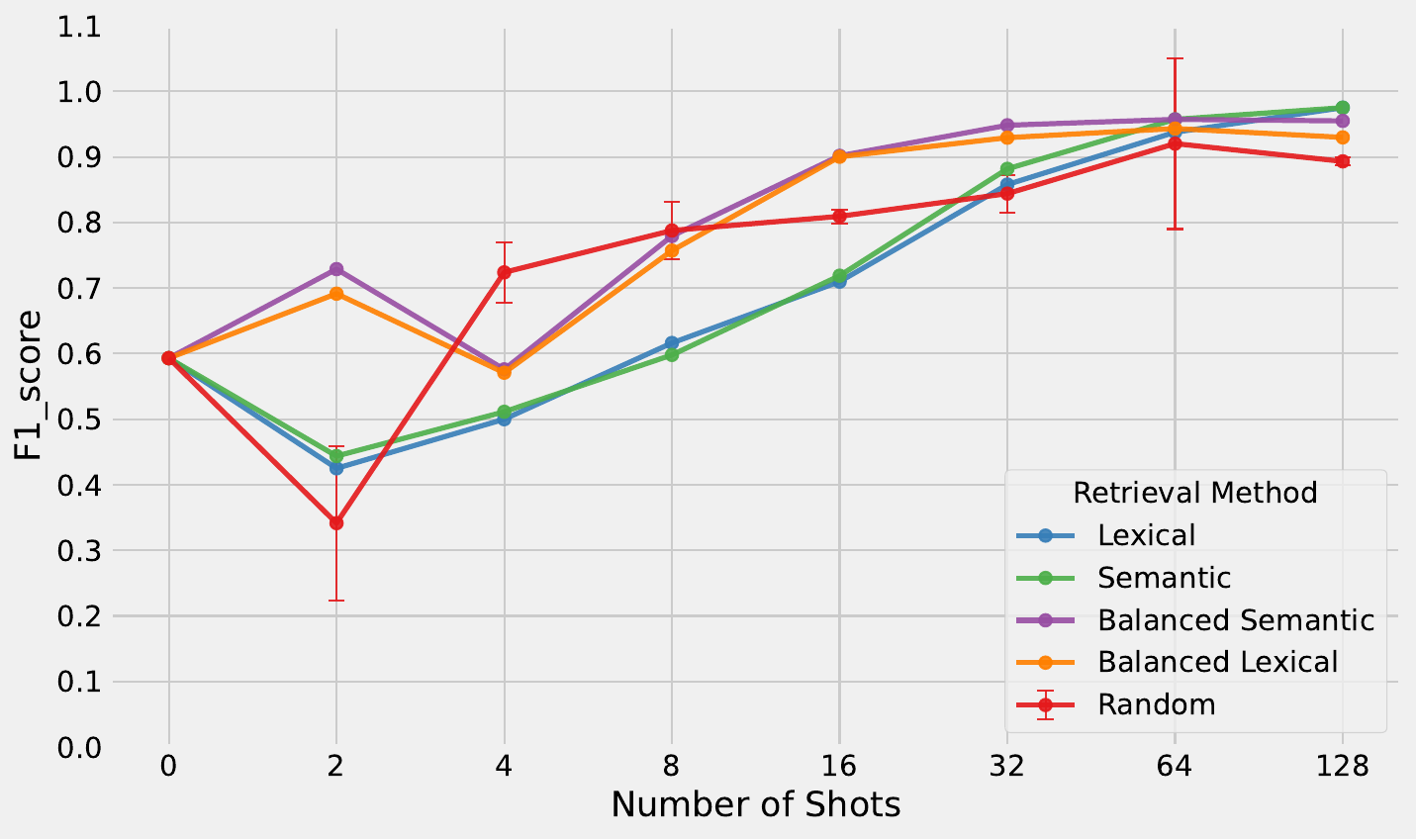}
        \caption{Mistral}
    \end{subfigure}
    \begin{subfigure}{0.32\textwidth}
        \centering
        \includegraphics[width=\textwidth]{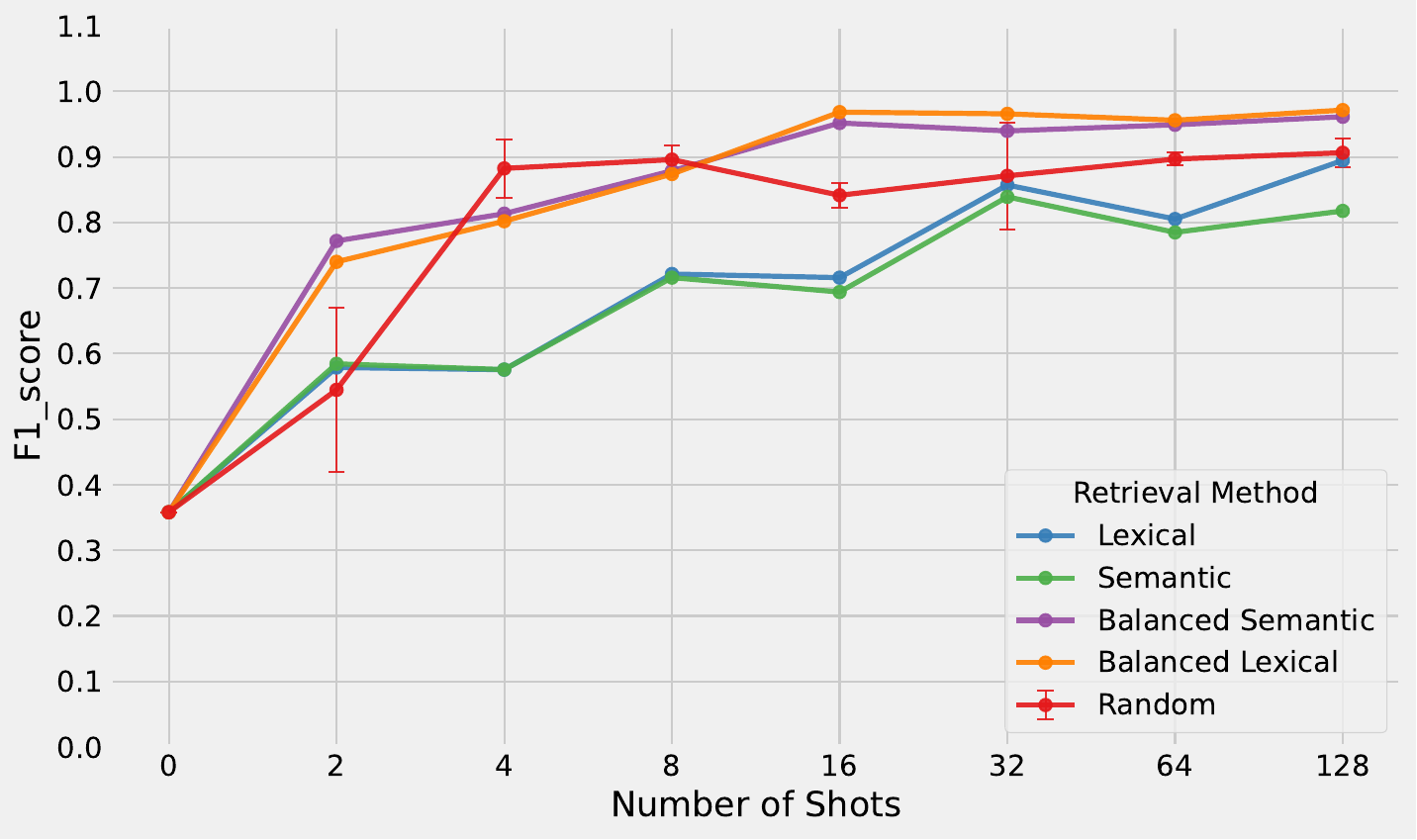}
        \caption{Qwen}
    \end{subfigure}
    \caption{The performance of ICL on Binary Spam Classification}
    \label{fig:single-task_uci}
\end{figure}

\subject{Spam.} 
As shown in Figure~\ref{fig:single-task_uci}, ICL performance on the UCI spam dataset generally improves with more demonstrations, though saturation occurs beyond $k=32$. 
Balanced Semantic and Balanced Lexical retrieval consistently yield the best results across all models, closely followed by Random. 
In contrast, Lexical and Semantic retrieval without label balancing exhibit instability, especially for Llama. 
For model-specific results, Llama achieves the best F1-score of 0.991 (Balanced Lexical), Qwen reaches 0.971 (Balanced Lexical), and Mistral achieves 0.957 (Balanced Semantic).

\begin{figure}
    \centering
    \begin{subfigure}{0.32\textwidth}
        \centering
        \includegraphics[width=\textwidth]{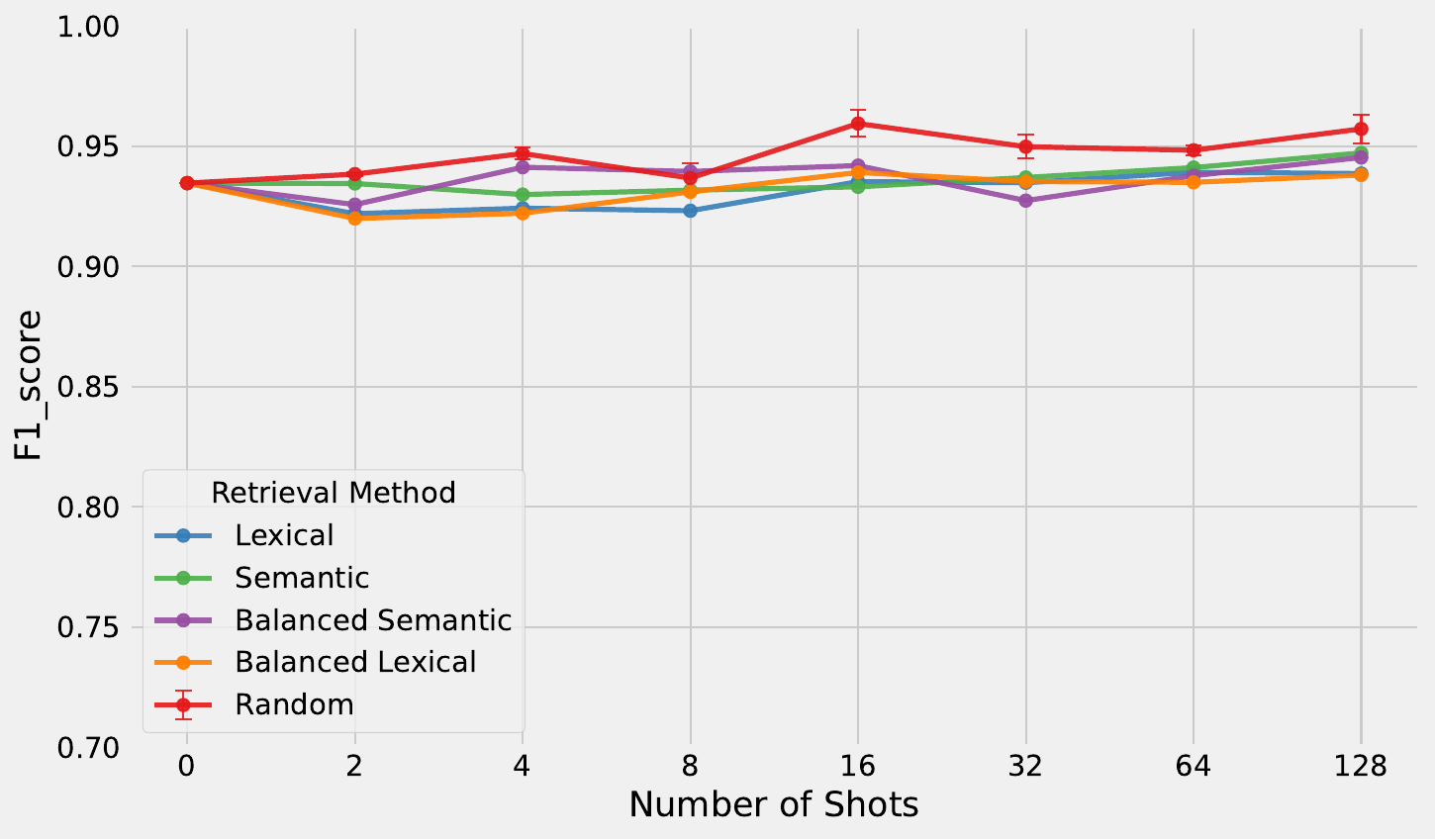}
        \caption{Llama}
    \end{subfigure}
    \begin{subfigure}{0.32\textwidth}
        \centering
        \includegraphics[width=\textwidth]{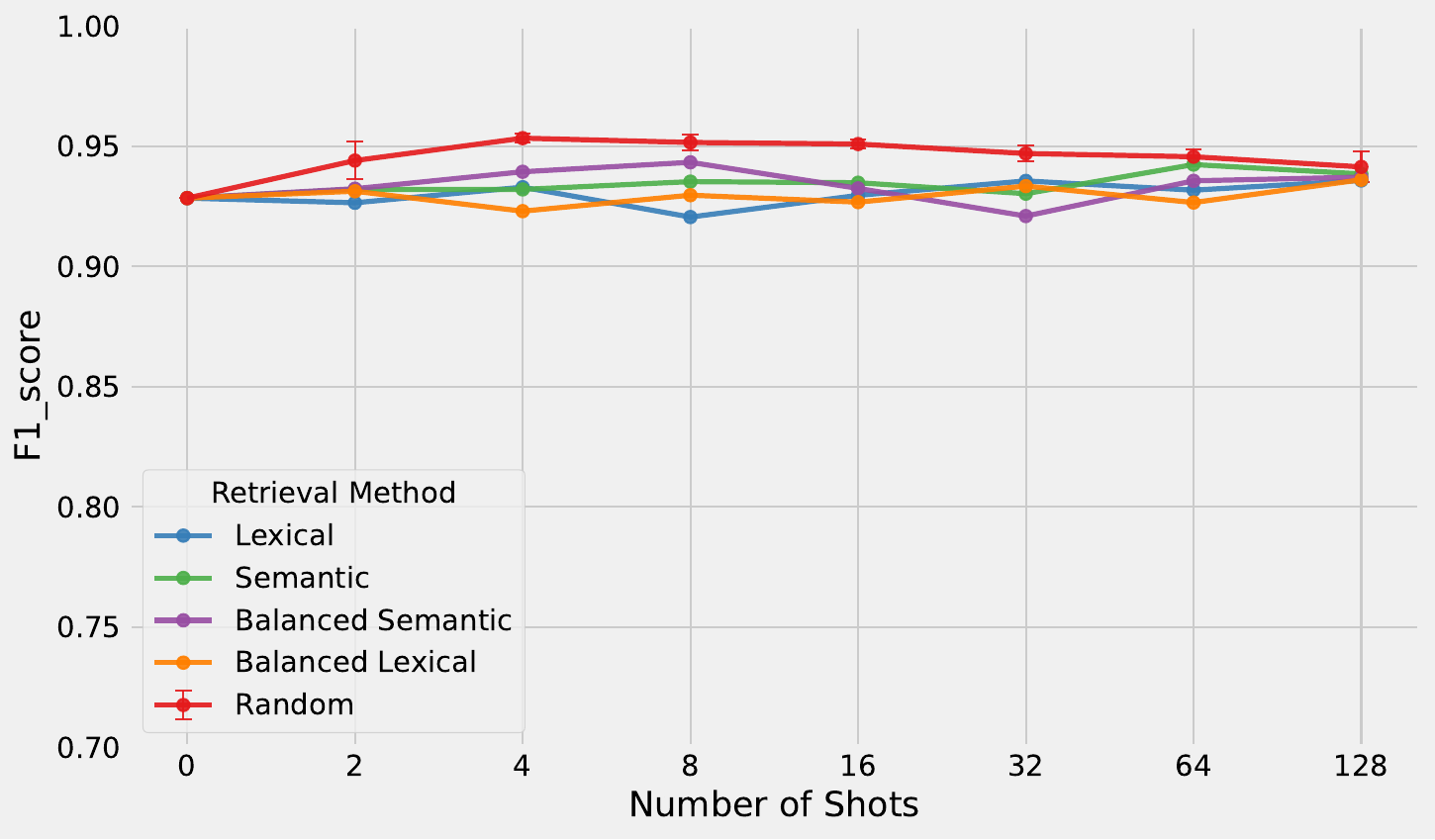}
        \caption{Mistral}
    \end{subfigure}
    \begin{subfigure}{0.32\textwidth}
        \centering
        \includegraphics[width=\textwidth]{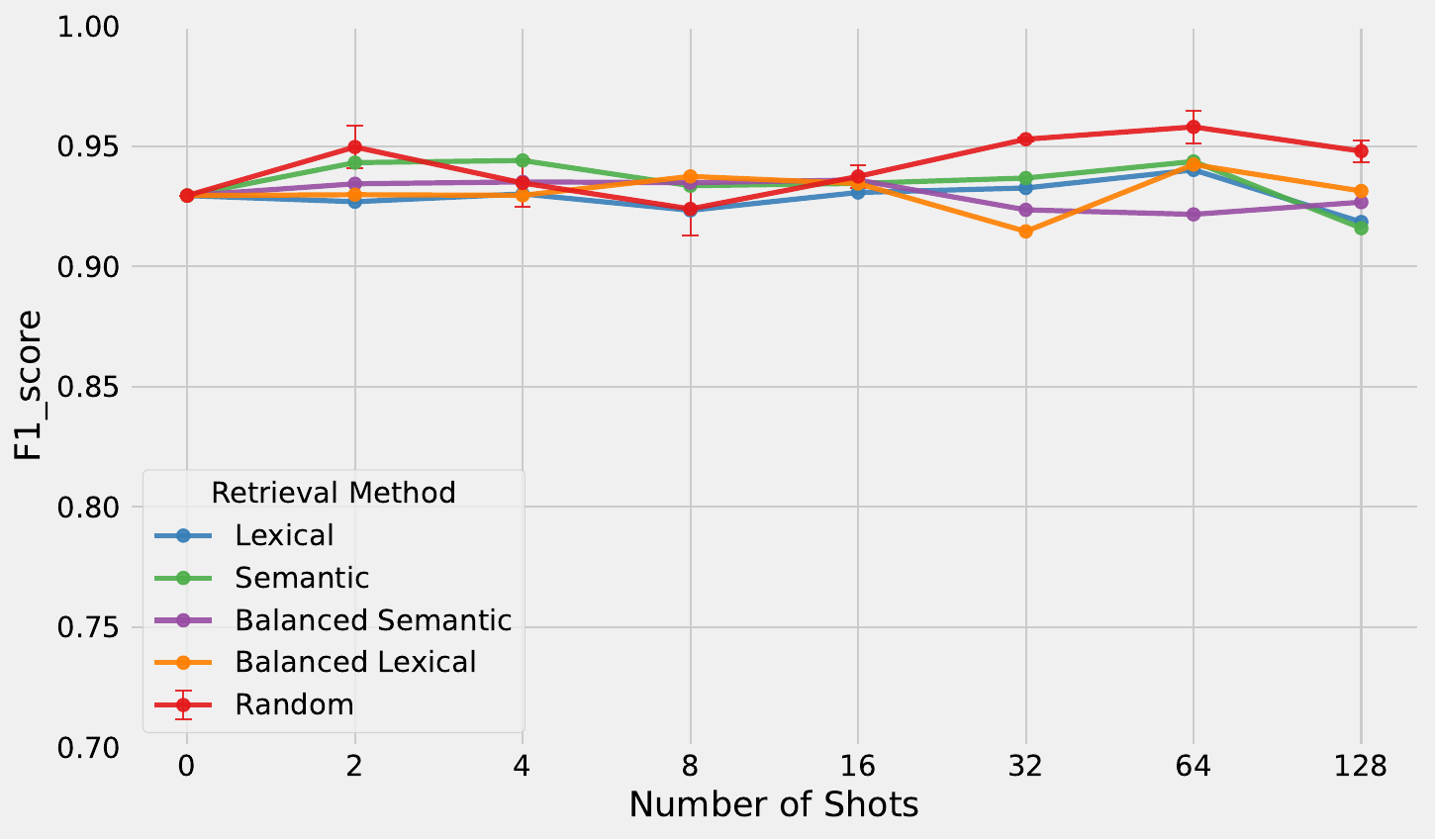}
        \caption{Qwen}
    \end{subfigure}
    \caption{The performance of ICL on Binary Sentiment Analysis }
    \label{fig:single-task_SST2}
\end{figure}

\subject{Sentiment.} 
Figure~\ref{fig:single-task_SST2} shows that all models achieve near-ceiling performance on SST-2, with F1-scores consistently above 0.9 regardless of retrieval strategy or number of shots. 
This suggests that sentiment classification is relatively easy for instruction-tuned models. 
Random retrieval shows a slight advantage as $k$ increases, but overall differences between strategies are marginal.

\begin{figure}
    \centering
    \begin{subfigure}{0.32\textwidth}
        \centering
        \includegraphics[width=\textwidth]{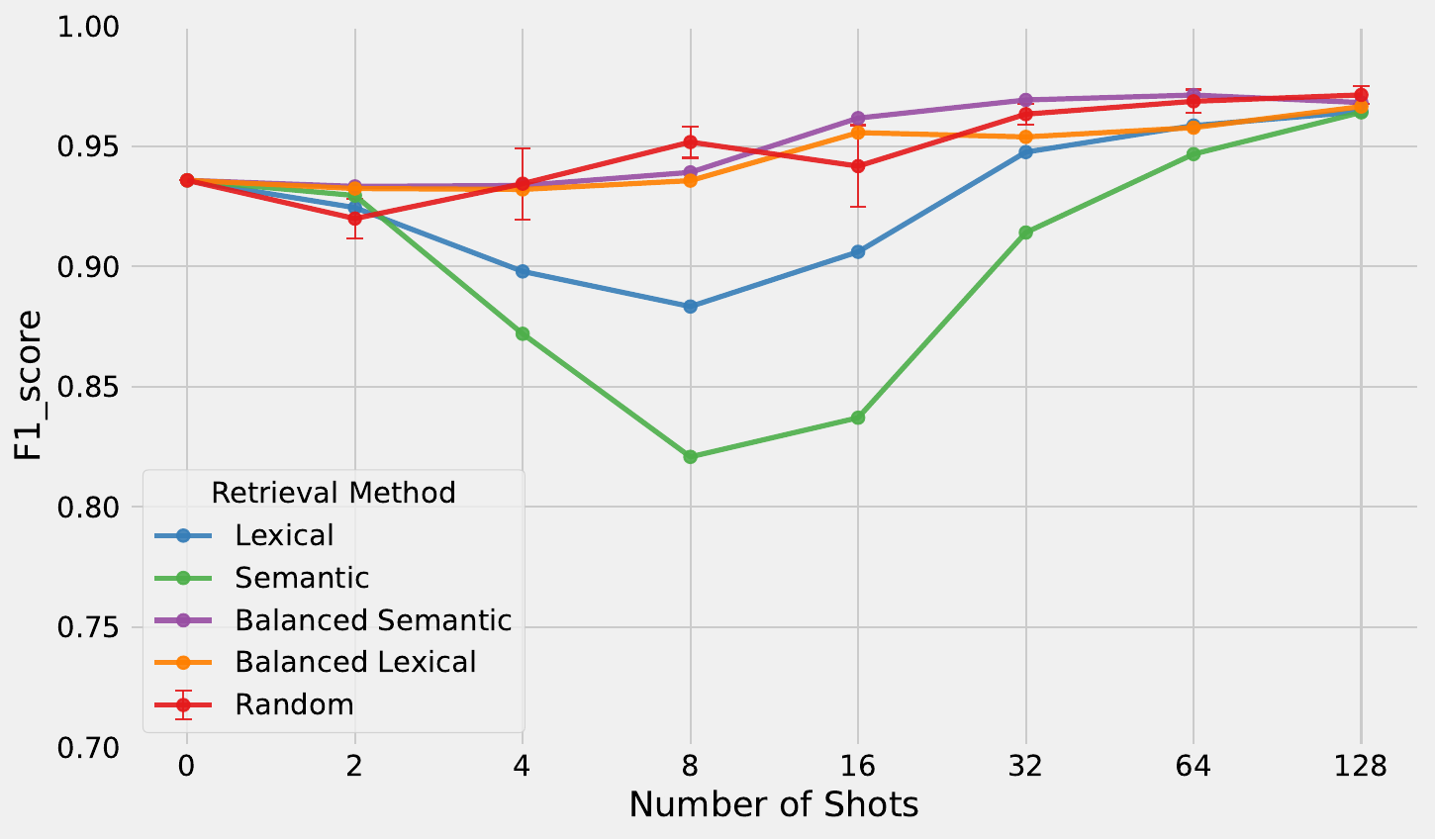}
        \caption{Llama}
    \end{subfigure}
    \begin{subfigure}{0.32\textwidth}
        \centering
        \includegraphics[width=\textwidth]{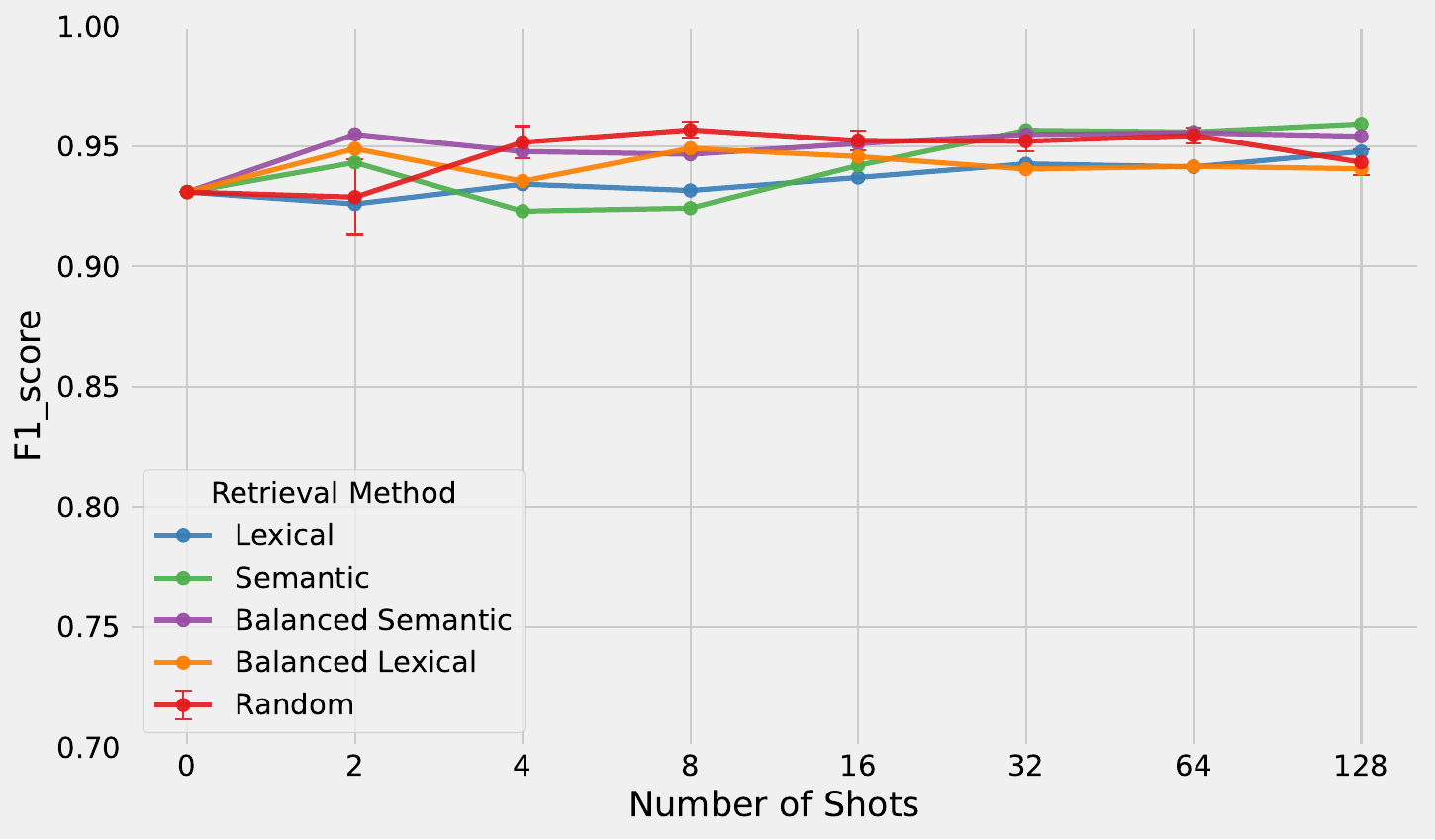}
        \caption{Mistral}
    \end{subfigure}
    \begin{subfigure}{0.32\textwidth}
        \centering
        \includegraphics[width=\textwidth]{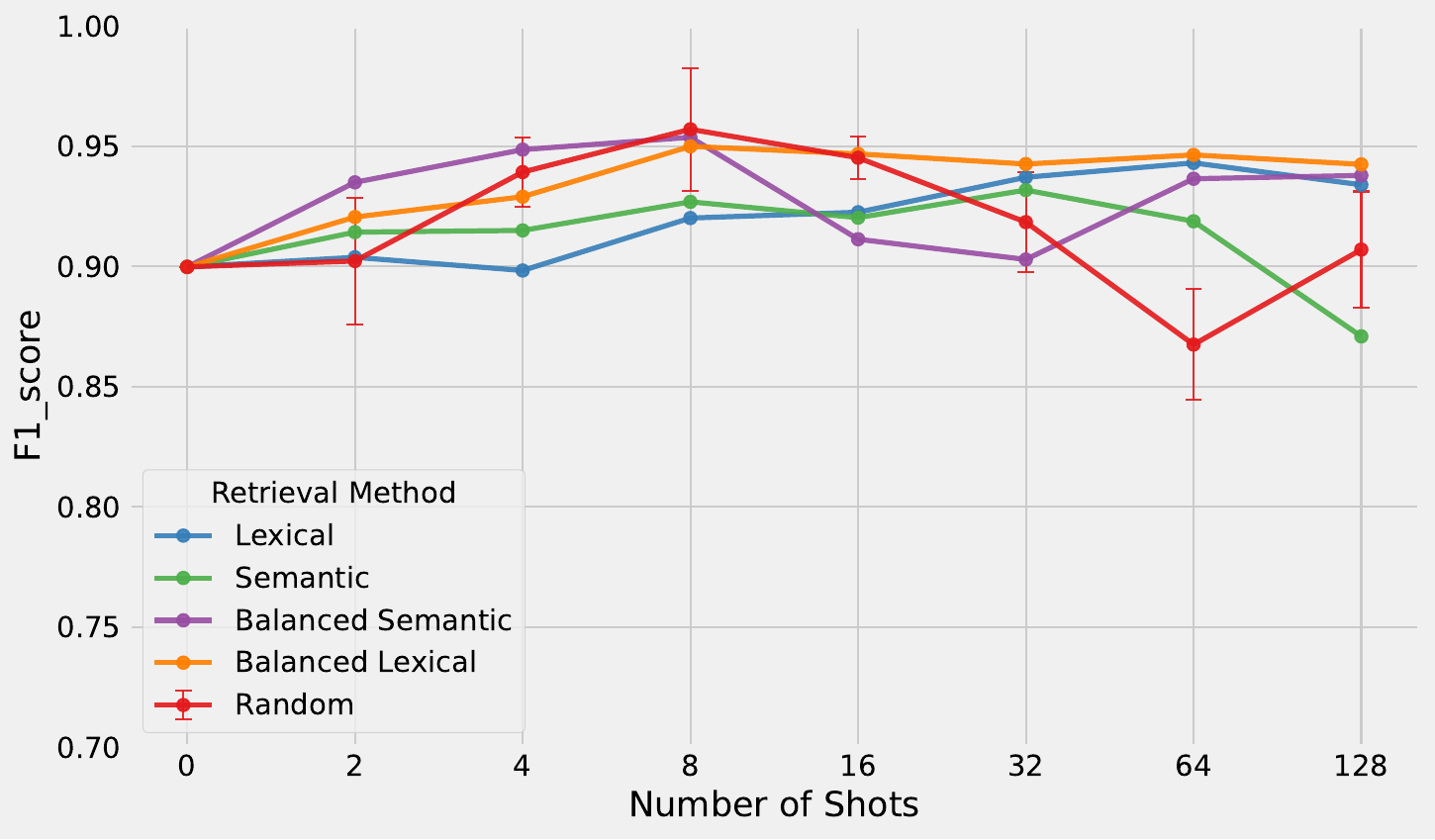}
        \caption{Qwen}
    \end{subfigure}
    \caption{The performance of ICL on binary toxicity classification}
    \label{fig:single-task_Textdetox}
\end{figure}

\subject{Toxicity.} 
As shown in Figure~\ref{fig:single-task_Textdetox}, performance on TextDetox improves with more demonstrations but exhibits less stability than UCI. 
Some retrieval strategies degrade at higher $k$, e.g., Mistral–Random and Qwen–Semantic. 
Top-performing strategies vary across models, but Random and Balanced Semantic are consistently competitive. 
Similar to UCI, Llama displays poor performance under Lexical and Semantic retrieval, whereas Random and Balanced Semantic achieve the strongest results, with Llama reaching 0.971 F1.

\subject{Comparison with Baselines.} Table~\ref{tab:single-task_model_comparison} summarizes the best-performing ICL classifiers for all three tasks along with a direct comparison with the baselines (fine-tuned BERT models). Key observations show that ICL exhibits strong performance across different tasks, with certain setups outperforming the BERT baseline. For instance, Llama with Balanced Semantic retrieval achieves an average F1-score of 0.970, which is higher than BERT's average of 0.959. 

Specifically, in the spam classification task, most of Llama's top ICL configurations surpass BERT's performance. Similarly, for the toxicity task, Llama's best ICL results match or exceed BERT's performance. However, for the sentiment analysis task, BERT's F1-score of 0.923 is relatively lower compared to the near-ceiling performance of ICL models, which all achieve F1-scores above 0.9. This comparison indicates that ICL can be a competitive alternative to traditional fine-tuned models like BERT, showcasing its efficiency and adaptability in binary classification tasks.

\subject{Summary.} Our results establish that ICL adapts effectively across diverse binary classification tasks for harmful content. Crucially, ICL models achieve performance comparable to or surpassing a fine-tuned BERT baseline across all three tasks, despite requiring no task-specific gradient updates and far less explicit supervision. This demonstrates ICL's strong potential as an efficient and highly adaptable alternative to traditional fine-tuning pipelines.

\begin{table}
    \centering
    \begin{threeparttable}
        \caption{Comparison of Best F1-score for Different Model-Retrieval Pairs Across Tasks}
        \label{tab:single-task_model_comparison}
        \begin{tabular}{c c c c c c}
            \toprule
            Model & Retrieval Method & Average & Spam & Sentiment & Toxicity \\
            \midrule
            Bert(Baseline) & - & 0.959 & 0.985 & 0.923 & 0.969 \\
            \midrule
            \multirow{3}{*}{Llama} 
                & Random             & 0.967 & 0.970 (128)\tnote{1} & 0.959 (16) & 0.971 (128) \\
                & Balanced Lexical   & 0.965 & 0.991 (128) & 0.939 (16) & 0.966 (128) \\
                & Balanced Semantic  & 0.970 & 0.994 (128) & 0.945 (128) & 0.971 (64) \\
            \midrule
            \multirow{3}{*}{Qwen}  
                & Random             & 0.940 & 0.906 (128) & 0.958 (64) & 0.957 (8) \\
                & Balanced Lexical   & 0.954 & 0.971 (128) & 0.942 (64) & 0.950 (8) \\
                & Balanced Semantic  & 0.950 & 0.961 (128) & 0.936 (16) & 0.954 (8) \\
            \midrule
            \multirow{3}{*}{Mistral} 
                & Random             & 0.943 & 0.920 (64) & 0.953 (4) & 0.957 (8) \\
                & Balanced Lexical   & 0.943 & 0.943 (64) & 0.936 (128) & 0.949 (8) \\
                & Balanced Semantic  & 0.952 & 0.957 (64) & 0.943 (8) & 0.955 (64) \\
            \bottomrule
        \end{tabular}
        \begin{tablenotes}
            \item[1] \textbf{F1-score (Number of Shots)}: The F1-score achieved by the model-retrieval pair, followed by the number of shots used in that configuration.
        \end{tablenotes}
    \end{threeparttable}
\end{table}

\section{Multi-Task Binary ICL}
\label{sec:multi-task_binary}
Having established strong single-task baselines, we now investigate the feasibility of ICL for \textbf{multi-task binary} harmful content classification. This setting tests the model's ability to learn a unified concept of "harmfulness" from heterogeneous categories (toxicity, spam, negative sentiment), a crucial step toward a general-purpose moderation system.
\subsection{Experimental Setup}

We only report experimental configurations that differ from the single-task setting (Section~\ref{sec:single-task_binary_icl}).

\subject{Datasets.} 
As shown in Table~\ref{tab:overview_of_datasets}, we merge the training data from all three binary datasets into a single training pool for retrieving demonstrations. 
Similarly, the testing data are consolidated into a unified evaluation set: 

\[\mathcal{X} = D_{\text{test}}^{\text{TextDetox}} \cup D_{\text{test}}^{\text{UCI}} \cup D_{\text{test}}^{\text{SST2}}, \quad D_{k}^{r} \subseteq D_{\text{train}}^{\text{TextDetox}} \cup D_{\text{train}}^{\text{UCI}} \cup D_{\text{train}}^{\text{SST2}}.\]

\subject{Label Names.} To construct a binary label space, we group $\{\textit{toxic}, \textit{spam}, \textit{negative}\}$ as \textbf{harmful}, and $\{\textit{non-toxic}, \textit{ham}, \textit{positive}\}$ as \textbf{benign}, yielding $\mathcal{Y} = \{\textit{harmful}, \textit{benign}\}$. 

\subject{Number of Shots.} 
To ensure balanced coverage of categories in the prompt, $k$ is set as a multiple of the number of classes, i.e., 
$k \in \{0, 6, 12, 24, 48, 96, 192\}$.

\subject{Retrieval Strategies.} 
Based on single-task results, we retain the two strongest strategies: Random and Balanced Semantic. 
Although Balanced Lexical also performed competitively, its mechanism closely resembles Balanced Semantic, which generally achieved better results; therefore, we omit it for efficiency. 
In the multi-task setting, Balanced Semantic is further extended to \textbf{Fine-grained Balanced Semantic}, which balances across six subcategories instead of the two superclasses. 

\subject{Task Description.} 
~\ref{appendix:task-description-mutli-task-binary} shows two prompt levels: 
\textbf{Level~1} mirrors the single-task binary description and only specifies the binary label space, while \textbf{Level~2} additionally provides explicit definitions and subcategory examples. 
Unless otherwise specified, we default to Level~1 in this section. 

\subsection{Results}

\begin{figure}
    \centering
    \begin{subfigure}{0.32\textwidth}
        \centering
        \includegraphics[width=\textwidth]{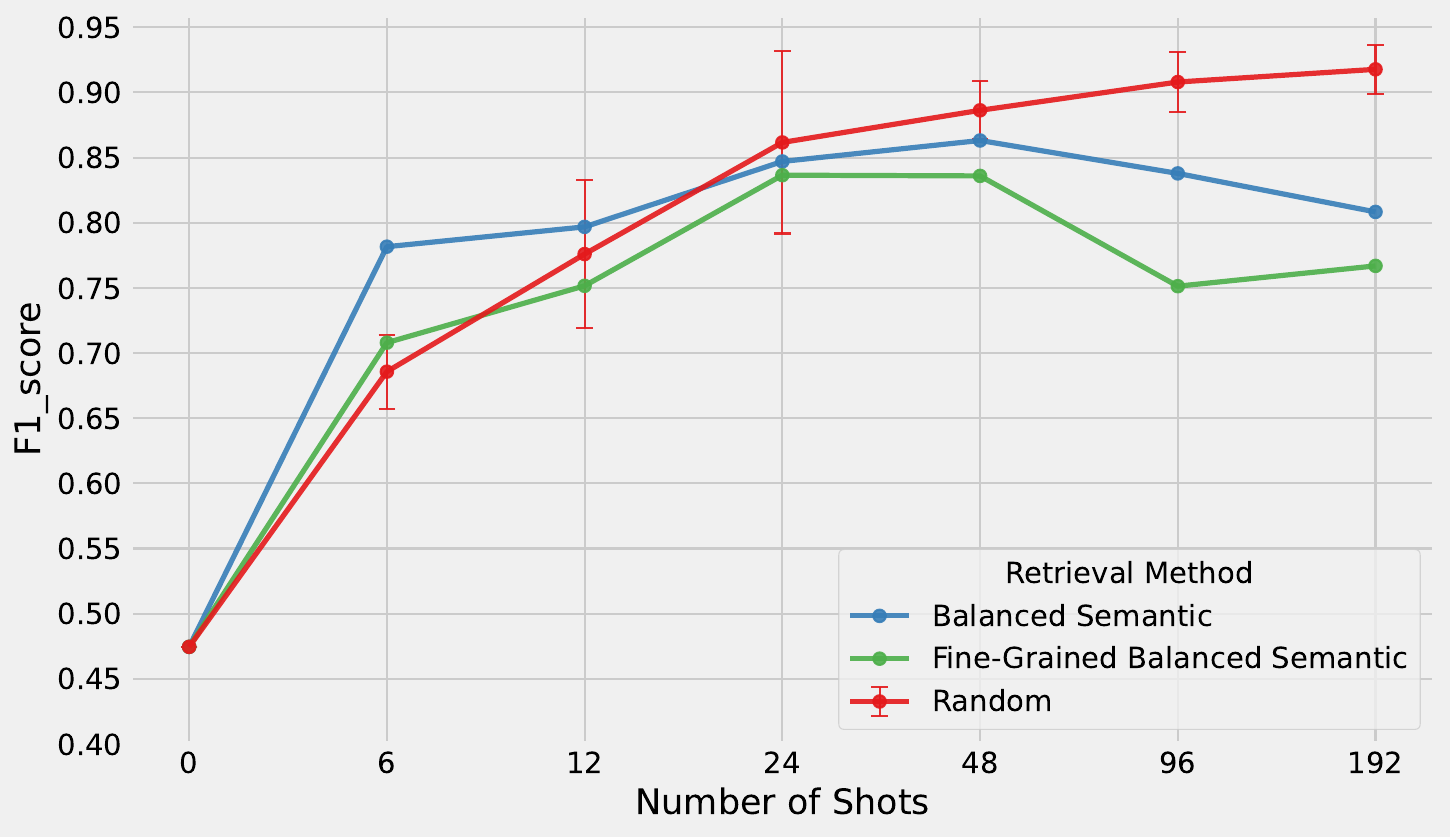}
        \caption{Llama}
    \end{subfigure}
    \begin{subfigure}{0.32\textwidth}
        \centering
        \includegraphics[width=\textwidth]{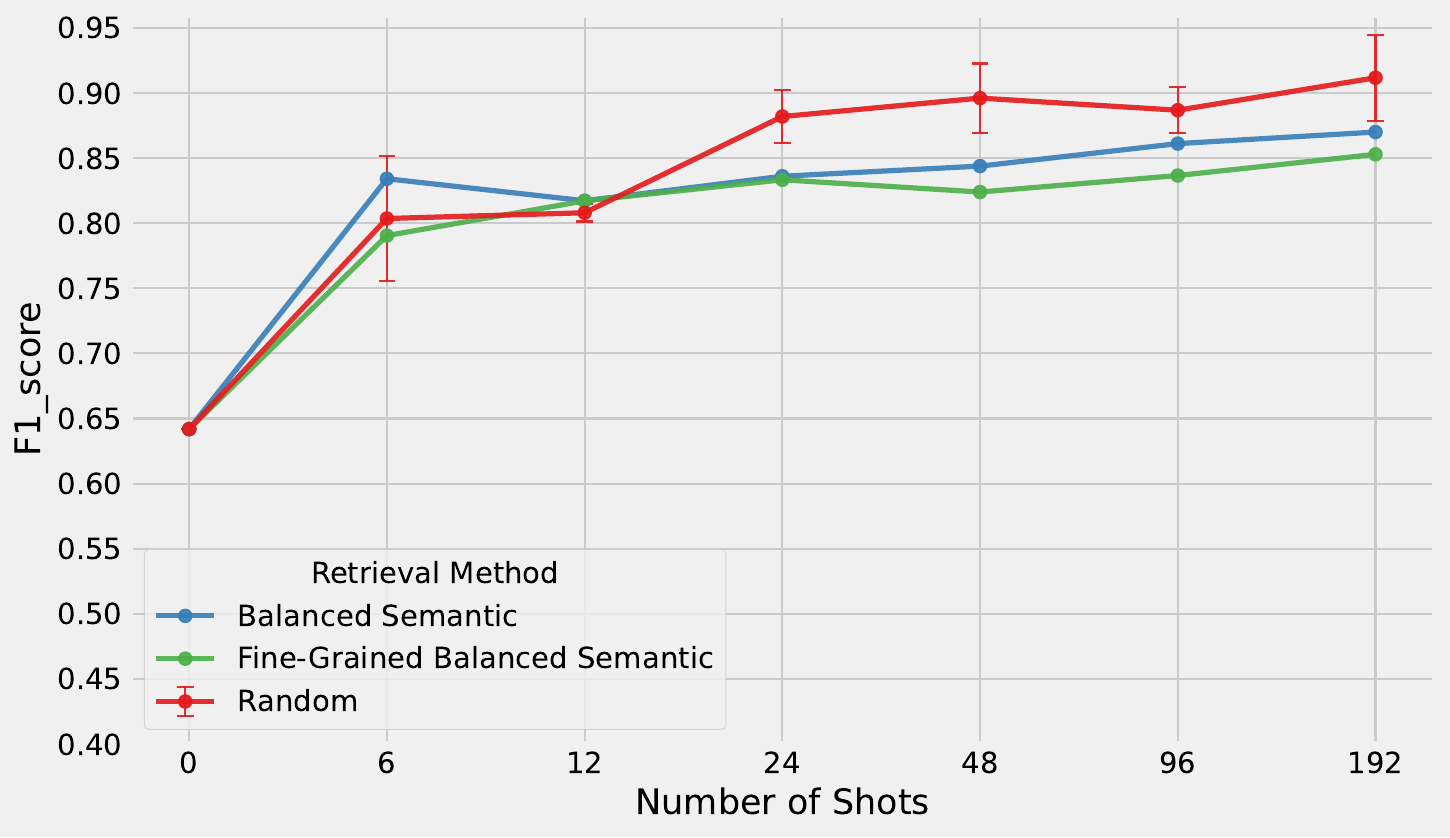}
        \caption{Mistral}
    \end{subfigure}
    \begin{subfigure}{0.32\textwidth}
        \centering
        \includegraphics[width=\textwidth]{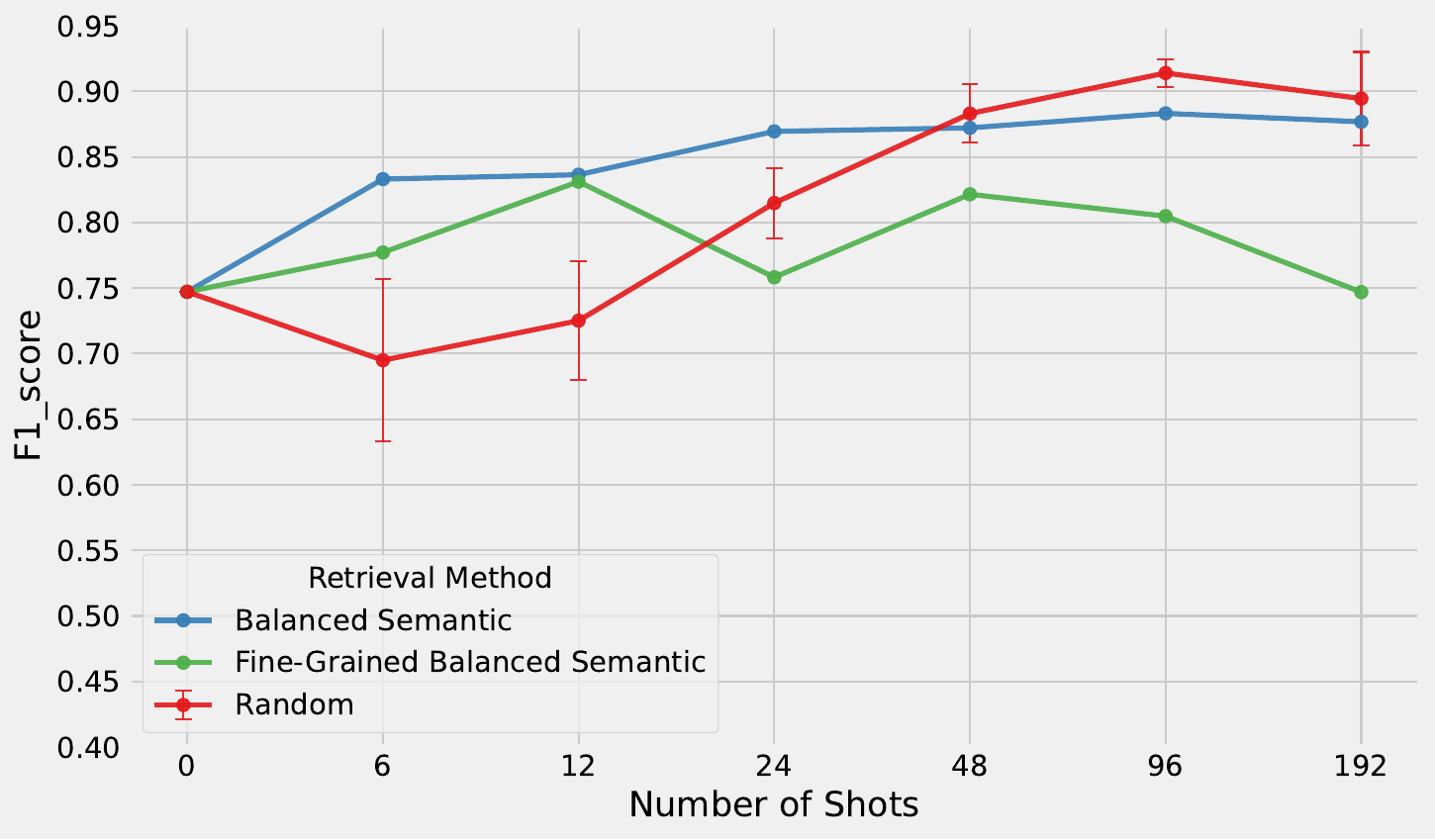}
        \caption{Qwen}
    \end{subfigure}
    \caption{The performance of multi-task binary ICL, as measured by F1-score.}
    \label{fig:multi-task_binary}
\end{figure}

\subject{Comparison of Retrieval Strategies and Models.} 
As shown in Figure~\ref{fig:multi-task_binary}, several consistent patterns emerge in different models.  
(1) Balanced Semantic generally outperforms Fine-grained Balanced Semantic, suggesting that enforcing balance at the superclass level is more beneficial than over-constraining to subcategories.  
(2) At small $k$, semantic-based strategies (Balanced and Fine-grained) outperform Random, but their advantage diminishes as $k$ increases. 
At larger $k$, Random consistently surpasses semantic-based methods, indicating that diversity may be more important than semantic closeness when sufficient demonstrations are available.  
(3) The best overall results are achieved under the Random strategy.  

Regarding performance difference among LLMs, 
as presented in Figure~\ref{fig:multi-task_binary} and Table~\ref{tab:multi-task_binary_model_comparison}, Llama-Random achieves the strongest result with an F1-score of 0.918 at 192 shots. 
Qwen and Mistral follow closely, achieving best F1-scores of 0.914 and 0.912, respectively.

\begin{figure}
    \centering
    \begin{subfigure}{0.32\textwidth}
        \centering
        \includegraphics[width=\textwidth]{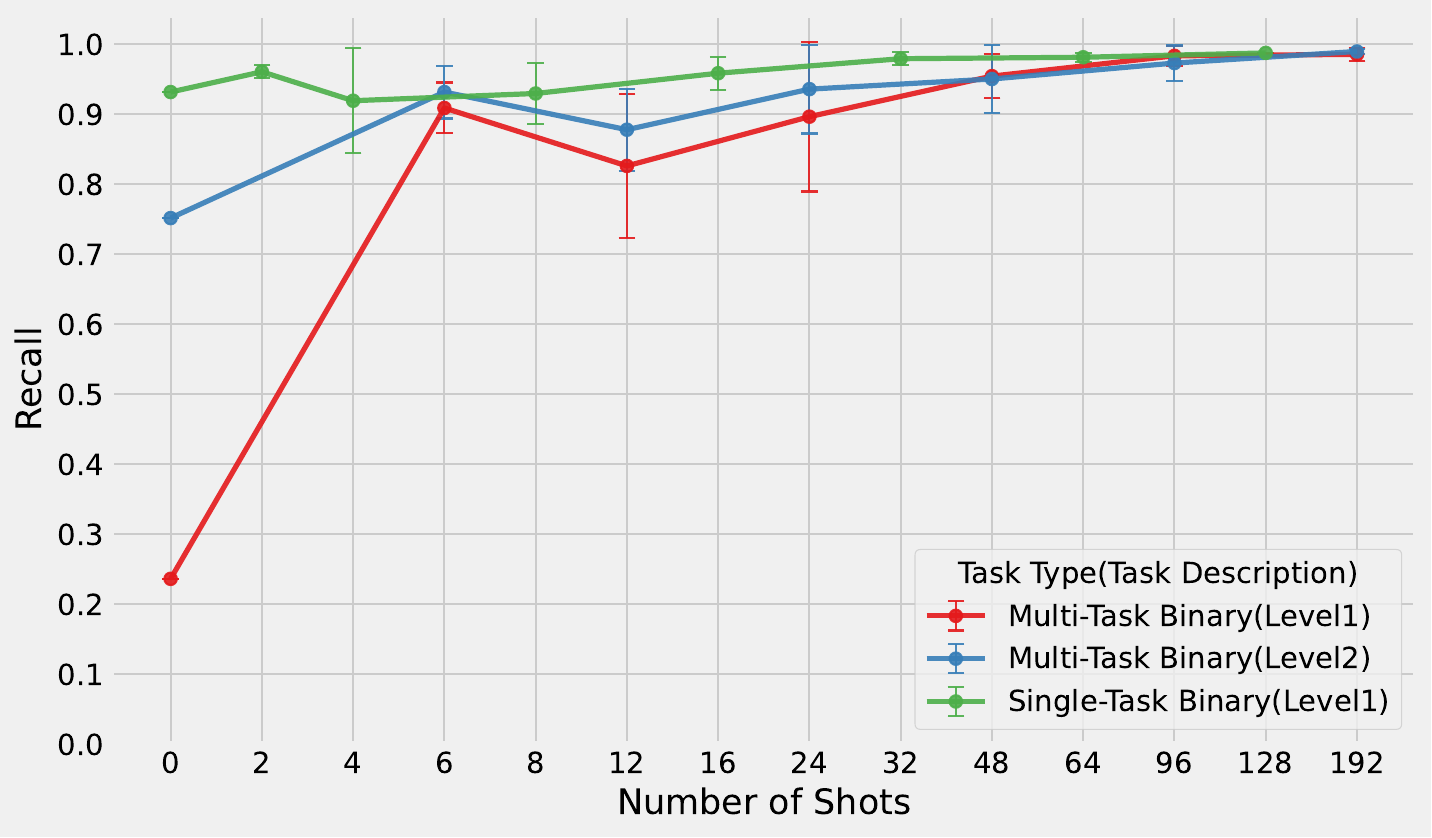}
        \caption{UCI}
    \end{subfigure}
    \begin{subfigure}{0.32\textwidth}
        \centering
        \includegraphics[width=\textwidth]{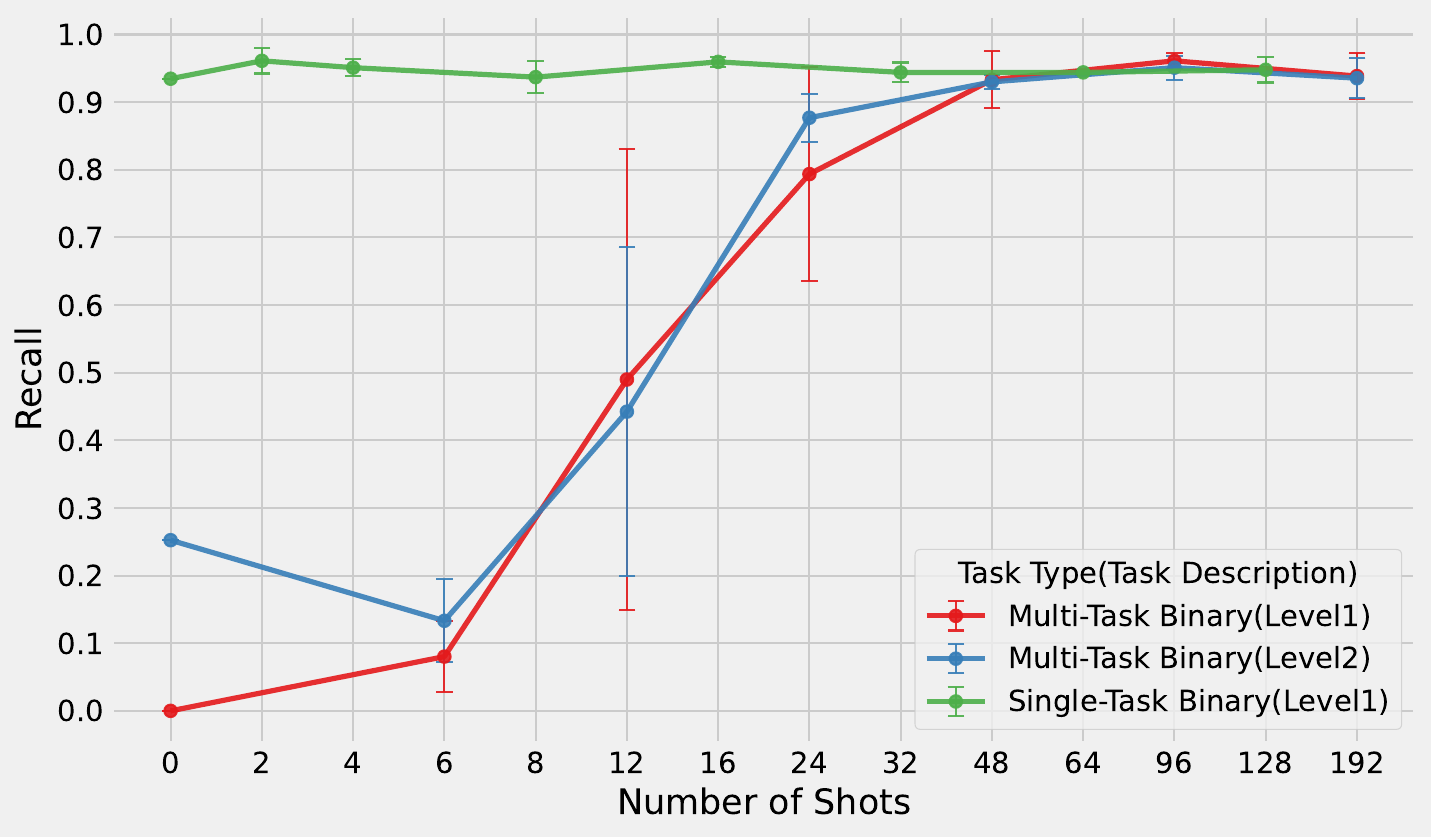}
        \caption{SST2}
    \end{subfigure}
    \begin{subfigure}{0.32\textwidth}
        \centering
        \includegraphics[width=\textwidth]{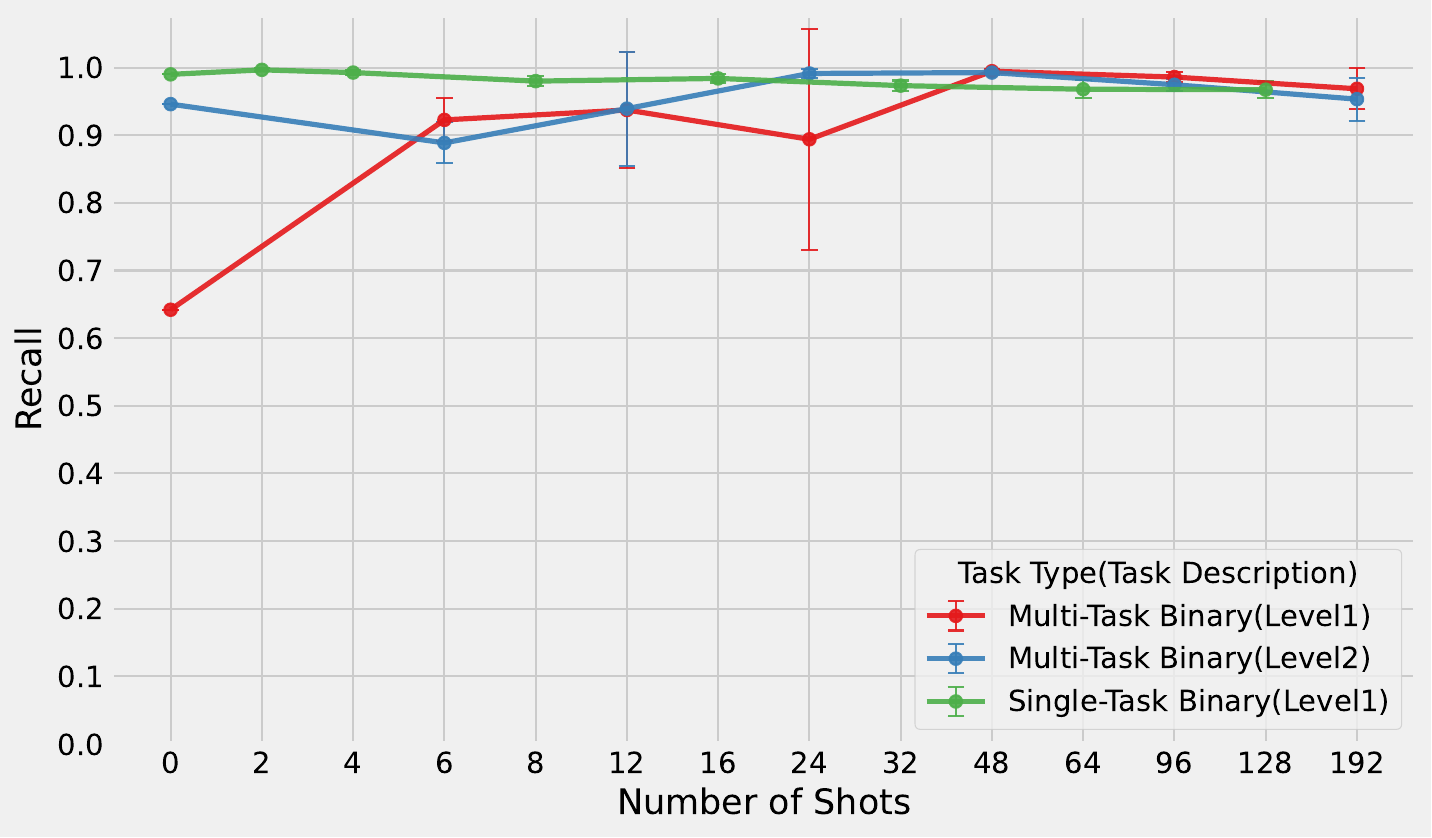}
        \caption{Textdetox}
    \end{subfigure}
    \caption{The performance comparison between multi-task binary ICL and single-task binary ICL, as measured by recall.}
    \label{fig:multi-task_binary_recall}
\end{figure}

\begin{figure}[ht]
    \centering
    \begin{subfigure}{0.32\textwidth}
        \centering
        \includegraphics[width=\textwidth]{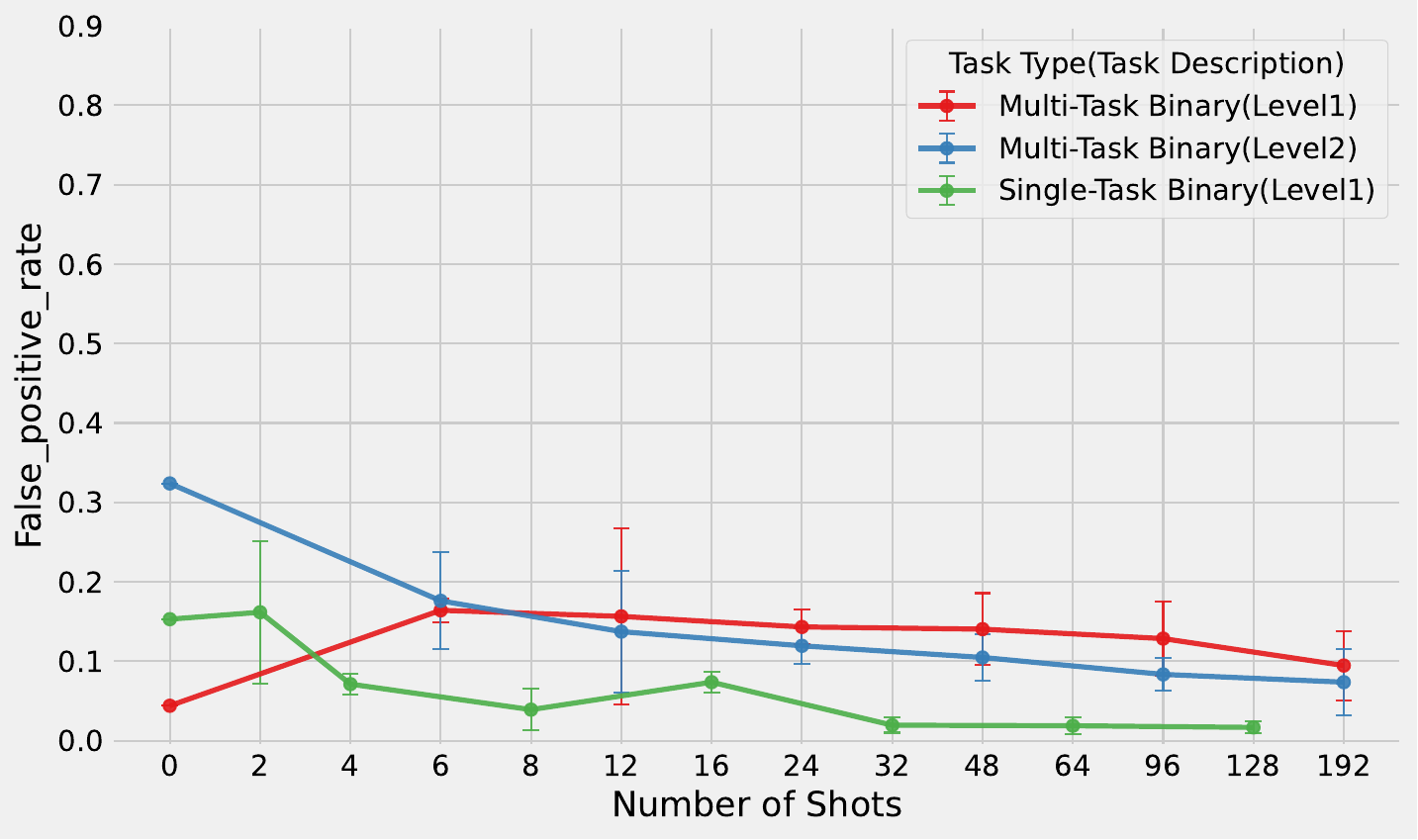}
        \caption{UCI}
    \end{subfigure}
    \begin{subfigure}{0.32\textwidth}
        \centering
        \includegraphics[width=\textwidth]{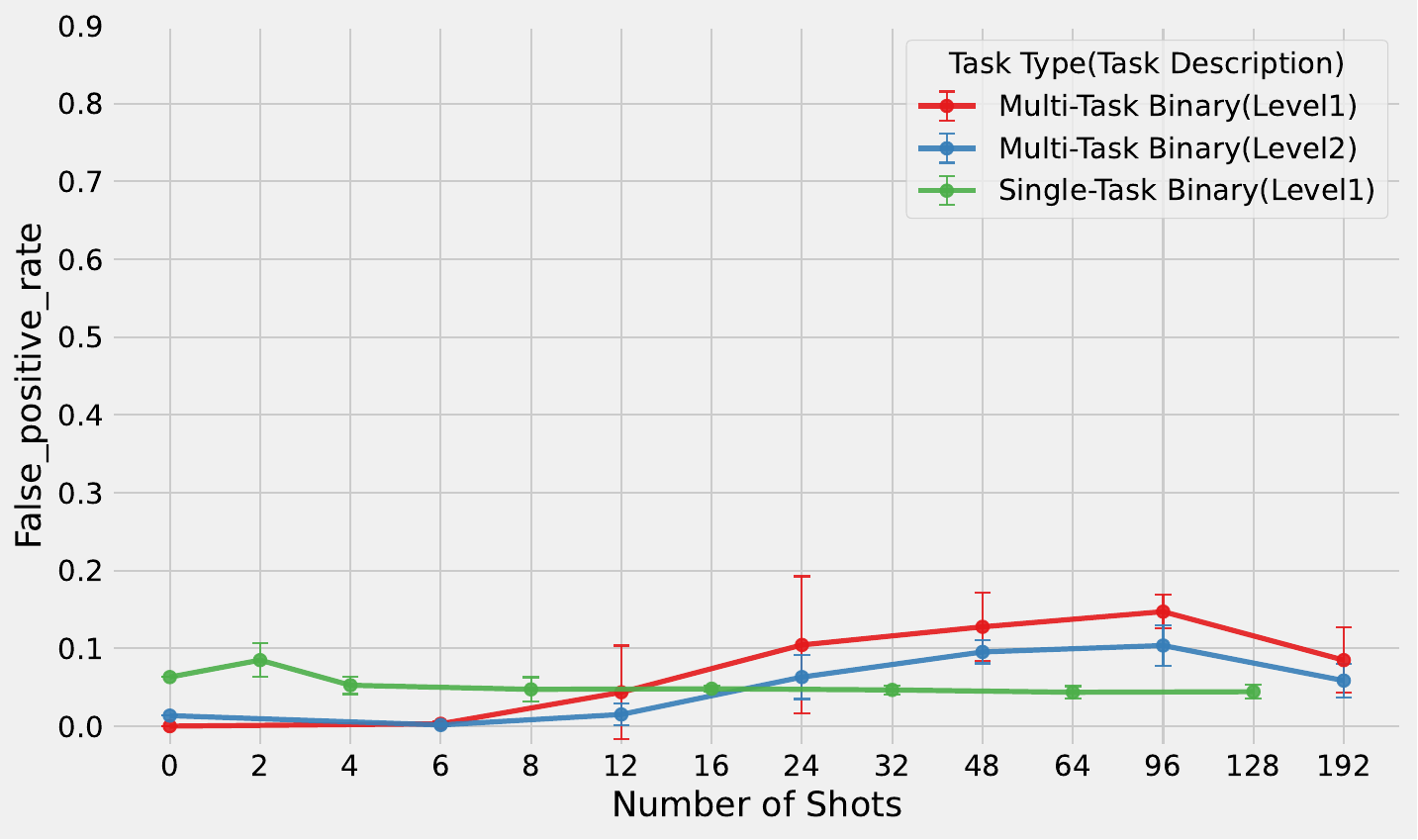}
        \caption{SST2}
    \end{subfigure}
    \begin{subfigure}{0.32\textwidth}
        \centering
        \includegraphics[width=\textwidth]{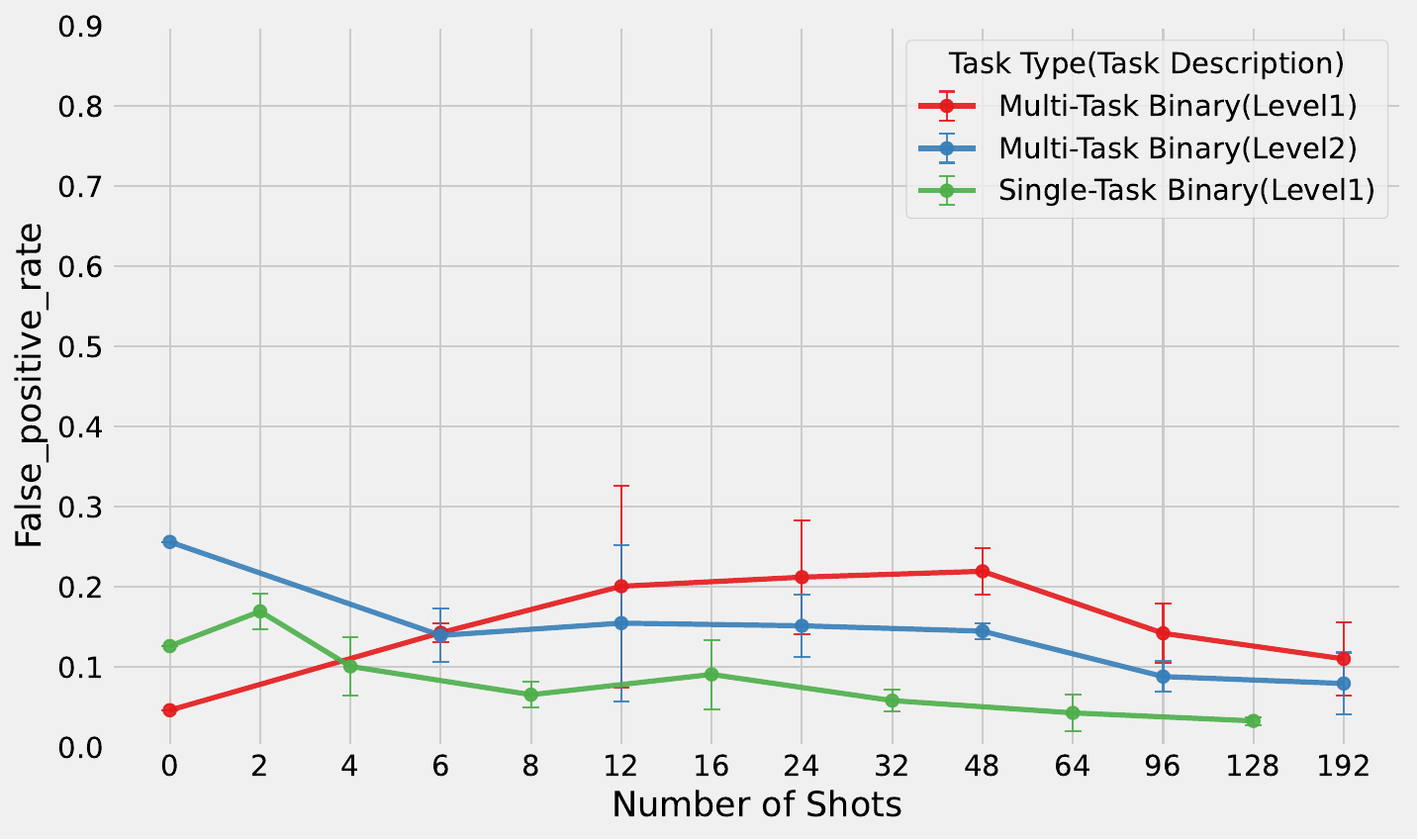}
        \caption{Textdetox}
    \end{subfigure}
    \caption{The false positive rate for Llama-Random on multi-task binary ICL and single-task ICL}
    \label{fig:multi-task_binary_fpr}
\end{figure}

\subject{Performance Gap: Multi-Task vs. Single-Task.} 
Although Llama-Random is the best-performing combination, its peak performance in multi-task ICL remains below that of single-task ICL. 
To investigate this discrepancy, we compare recall and false positive rates (FPR) between the two settings. 
As shown in Figure~\ref{fig:multi-task_binary_recall}, recall is comparable or even higher in multi-task ICL, particularly on TextDetox, indicating that reduced performance is not due to missed harmful samples. 
Instead, Figure~\ref{fig:multi-task_binary_fpr} shows that the FPR is substantially higher in multi-task ICL. 
This outcome is expected: when exposed to heterogeneous harmful categories, the model learns a broader notion of harmfulness, increasing the likelihood of misclassifying benign texts as harmful.

\subject{Mitigating False Positives with Enhanced Task Descriptions.} 
To mitigate elevated FPR, we introduce Level~2 task descriptions with explicit definitions and examples of harmful and benign categories (Figure~\ref{fig:description_of_multi-task_binary}). 
As shown in Figures~\ref{fig:multi-task_binary_recall} and \ref{fig:multi-task_binary_fpr}, Level~2 task descriptions significantly reduce FPR while preserving recall. 
Consequently, the best F1-score of Llama-Random improves from 0.918 (192 shots) to 0.934 (192 shots). 

These findings demonstrate that richer task descriptions are crucial for helping models disambiguate harmfulness in multi-task contexts, significantly mitigating the false positive problem. While a performance gap with single-task models remains, this work establishes a strong foundation for unified harmful content detection.

\begin{table}
    \centering
    \begin{threeparttable}
        \caption{Best-performing ICL setups for multi-task binary harm text classification, where each cell shows the best F1-score followed by the number of demonstration shots in parentheses.}
        \label{tab:multi-task_binary_model_comparison}
        \begin{tabular}{c c c c}
            \toprule
            \multirow{2}{*}{Model} & \multicolumn{3}{c}{Retrieval Strategy} \\
            \cmidrule(lr){2-4}
             & Random & Balanced Semantic & Fine-grained Balanced Semantic \\
            \midrule
            Llama &\textbf{0.918(192)} & 0.863(48) & 0.836(24) \\
            Mistral & 0.912(192) & 0.870(192) & \textbf{0.853(192)} \\
            Qwen & 0.914(96) & \textbf{0.883(96)} & 0.831(12) \\
            \bottomrule
        \end{tabular}
    \end{threeparttable}
\end{table}

\section{Multi-Task Multi-Class ICL}
\label{sec:multi-task_multi}
While binary classification identifies whether content is harmful, effective moderation and \textbf{personalization} often require understanding the \textit{type} of harm. To this end, we extend our evaluation to \textbf{multi-task multi-class} ICL, where the model must distinguish between specific harmful categories (toxic, spam, negative) and benign content.

\subsection{Experimental Setup}

We only report the experimental configurations that differ from the Multi-Task Binary ICL (Section~\ref{sec:multi-task_binary}).

\subject{Label Names.} We extend the binary setting into a four-class setting by keeping harmful categories distinct while merging non-toxic, ham, and positive into a single benign category. Formally, $\mathcal{Y} = \{\textit{toxic}, \textit{spam}, \textit{negative}, \textit{benign}\}$.  

\subject{Retrieval Strategies.} We follow the retrieval setting from Multi-Task Binary ICL. However, since this is a multi-class task with four distinct labels (\(\mathcal{Y}=\{\textit{toxic}, \textit{spam}, \textit{negative}, \textit{benign}\}\)), we retain the \textit{Fine-grained Balanced Semantic} strategy, which enforces balance across all four classes, and discard the binary-level \textit{Balanced Semantic} strategy.

\subject{Task Descriptions.} Based on the findings in Multi-Task Binary ICL, where Level 2 prompts consistently improved performance, we also evaluate both Level 1 and Level 2 prompts here. To ensure efficiency, Level 2 prompts are only applied to the best-performing model-strategy pair.

\subject{Metrics.} For overall performance, we report the weighted average F1-score across all categories. For per-class analysis, we adopt precision, recall, and F1-score, consistent with Single-Task Binary ICL (Section~\ref{sec:single-task_binary_icl}). We also examine confusion matrices for error analysis.

\subsection{Results}

\begin{figure}[ht]
    \centering
    \begin{subfigure}{0.32\textwidth}
        \centering
        \includegraphics[width=\textwidth]{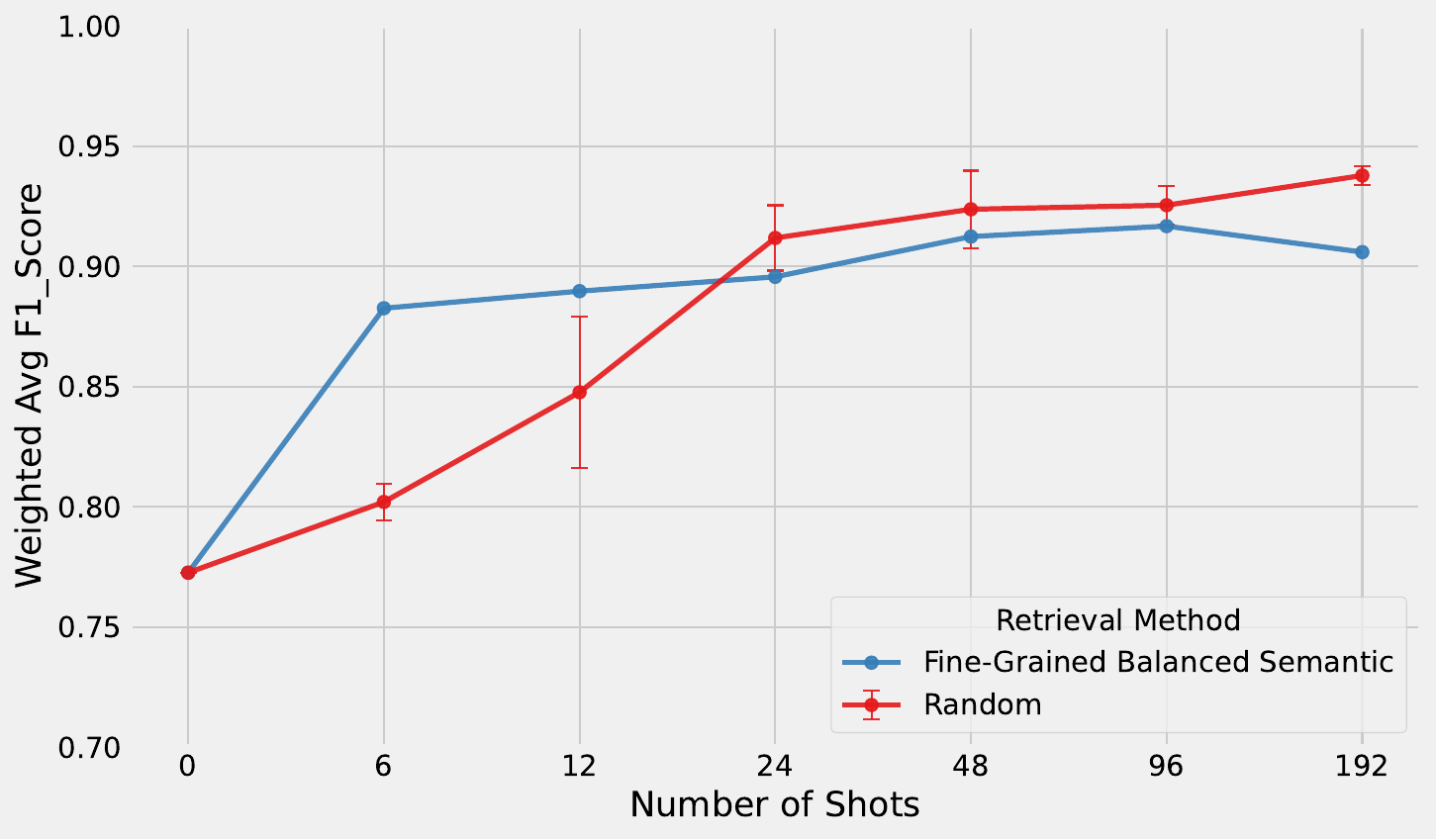}
        \caption{Llama}
    \end{subfigure}
    \begin{subfigure}{0.32\textwidth}
        \centering
        \includegraphics[width=\textwidth]{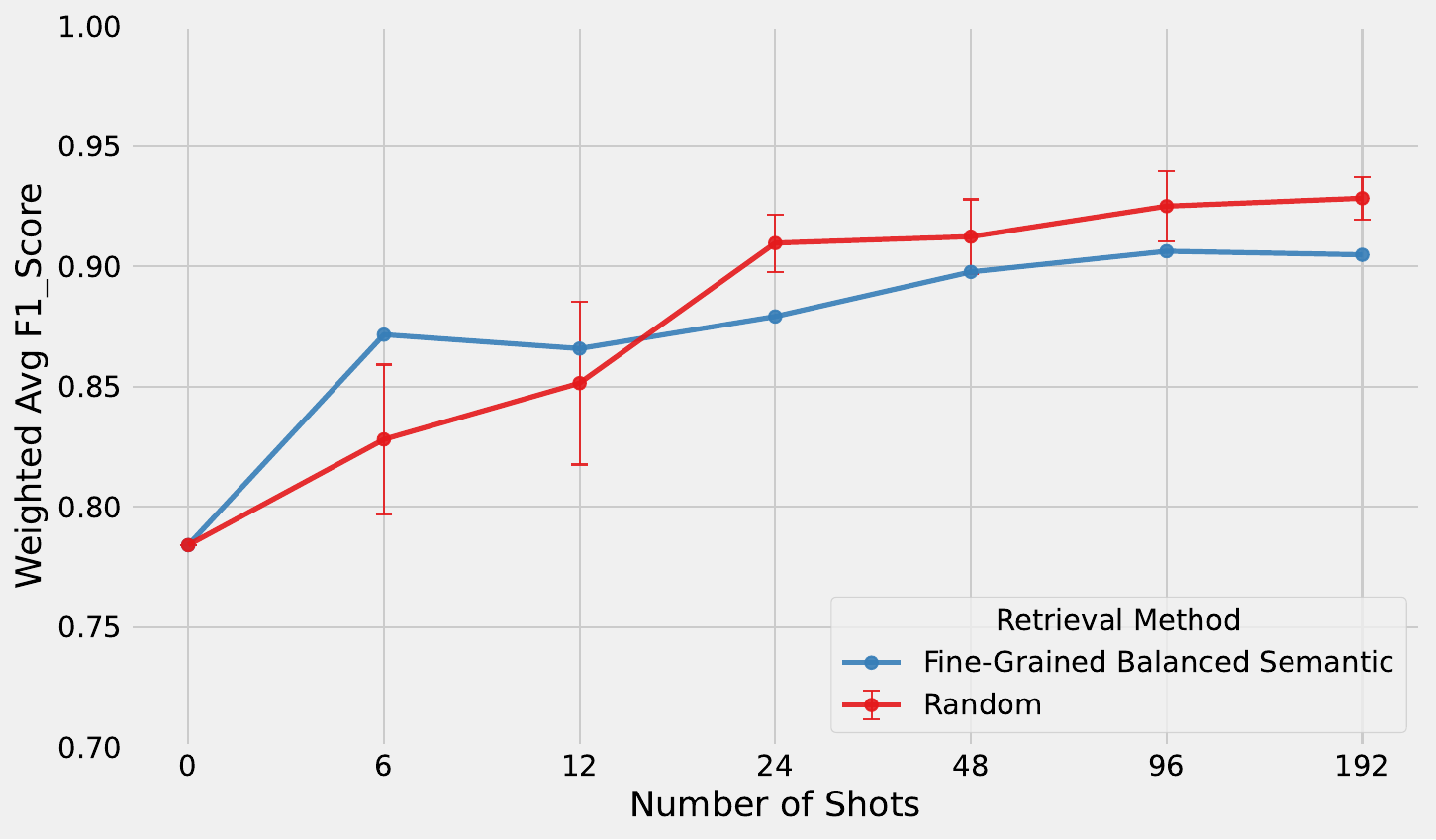}
        \caption{Mistral}
    \end{subfigure}
    \begin{subfigure}{0.32\textwidth}
        \centering
        \includegraphics[width=\textwidth]{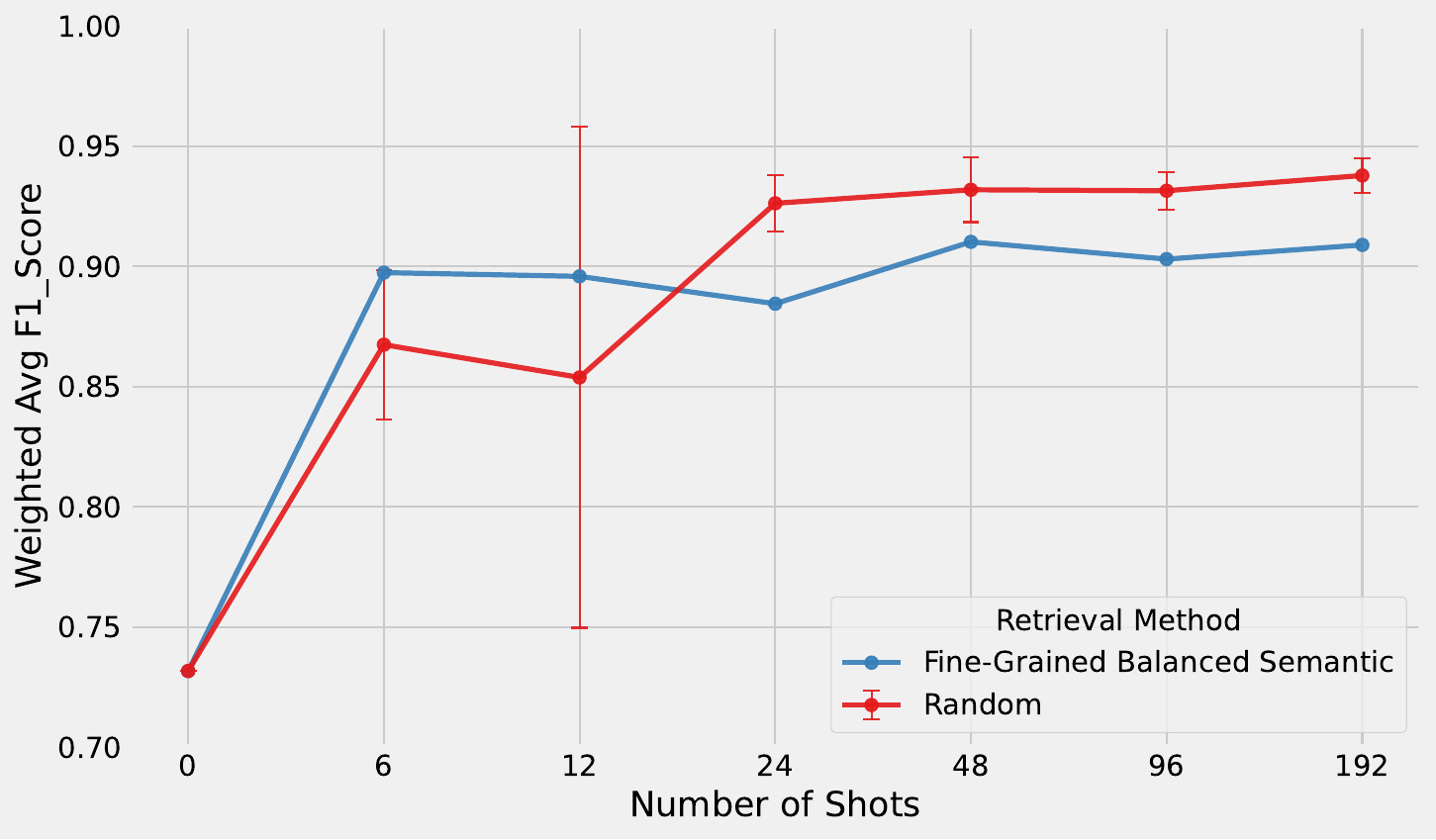}
        \caption{Qwen}
    \end{subfigure}
    \caption{The overall performance of Multi-task Multi-class ICL}
    \label{fig:multi-task_multi-class}
\end{figure}

\begin{figure}
    \centering
    \begin{subfigure}{0.45\linewidth}
        \centering
        \includegraphics[width=\linewidth]{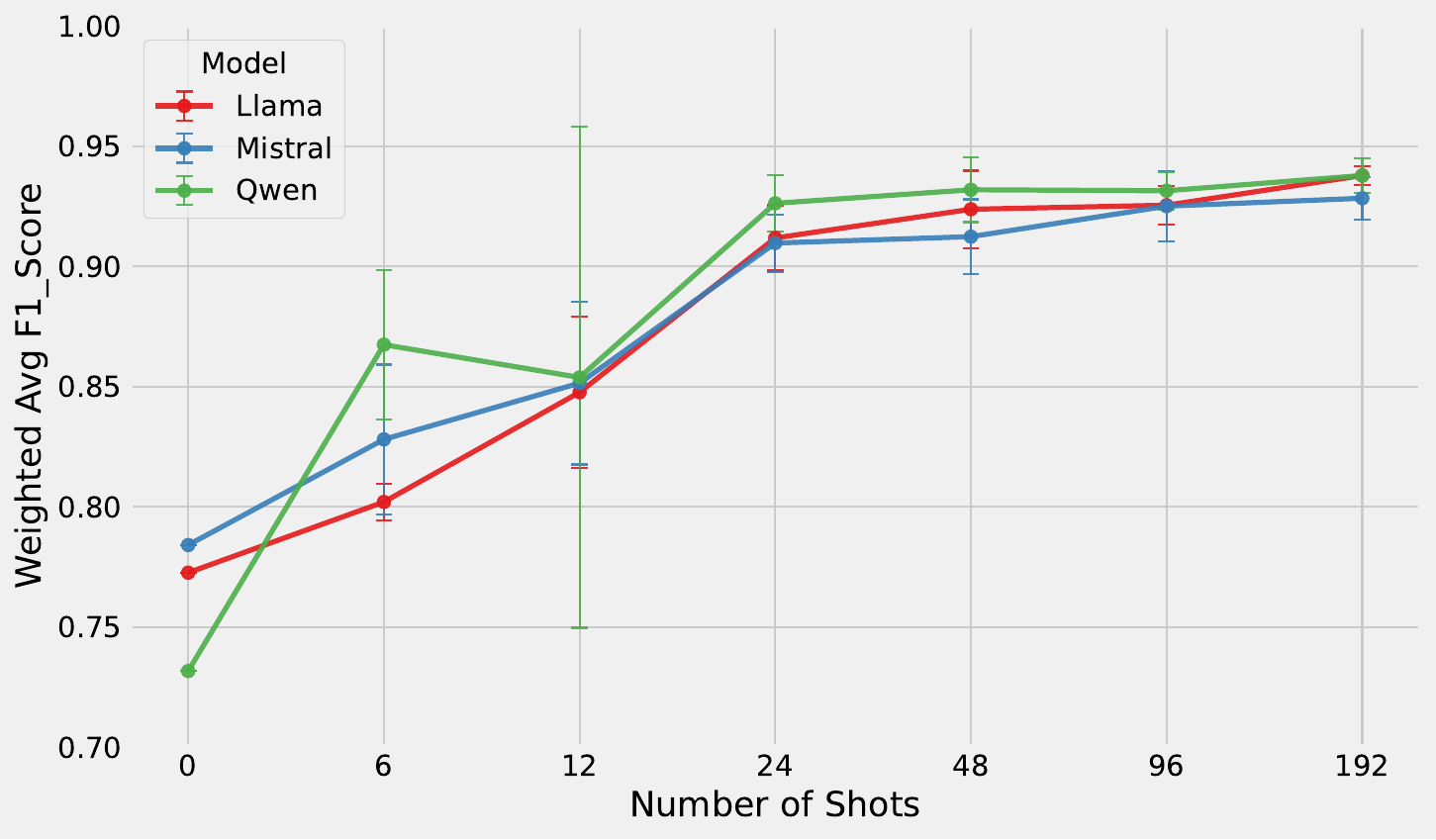}
        \captionsetup{width=\linewidth}
        \caption{Different LLMs. The retrieval strategy is set as random.}
        \label{fig:multi-task_multi-class_random}
    \end{subfigure}
    \hfill
    \begin{subfigure}{0.45\linewidth}
        \centering
        \includegraphics[width=\linewidth]{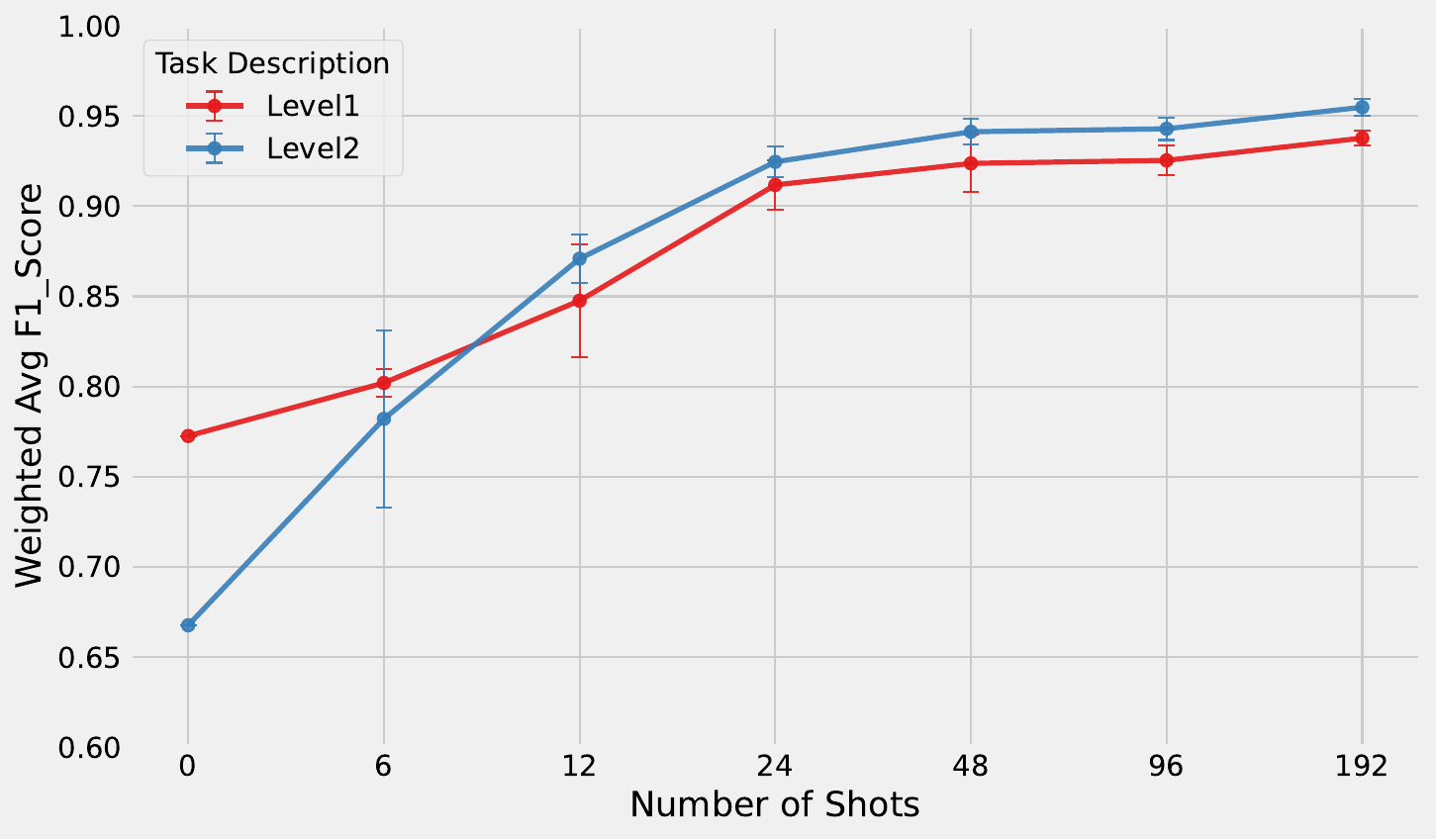}
        \captionsetup{width=\linewidth}
        \caption{Level 1  and level 2 prompts with Llama and random.}
        \label{fig:multi-task_multi-class_llama_f1_l12}
    \end{subfigure}
    \caption{The impact of key ICL parameters on multi-task multi-class performance.}
    \label{fig:multi-task_multi-class_icl_params}
\end{figure}

\subject{Comparison between different models and retrieval strategies.}  
As shown in Figure~\ref{fig:multi-task_multi-class}, we observe a consistent trend with the binary case: Fine-grained Balanced Semantic performs better than Random at small shot counts, while Random surpasses it as the number of shots increases and then stabilizes.  
Since we are primarily concerned with peak performance, we compare models under the Random strategy (Figure~\ref{fig:multi-task_multi-class_random}). Qwen generally achieves slightly higher weighted F1-scores than Llama and Mistral, except at $k=0$ and $k=192$. Both Qwen-Random and Llama-Random reach the highest weighted F1-score 0.938 at 192 shots. Given Llama’s consistently strong performance across tasks, we choose Llama-Random, rather than Qwen-Random, as the representative configuration for further analysis.

\begin{figure}[ht]
    \centering
    \begin{subfigure}{0.32\textwidth}
        \centering
        \includegraphics[width=\textwidth]{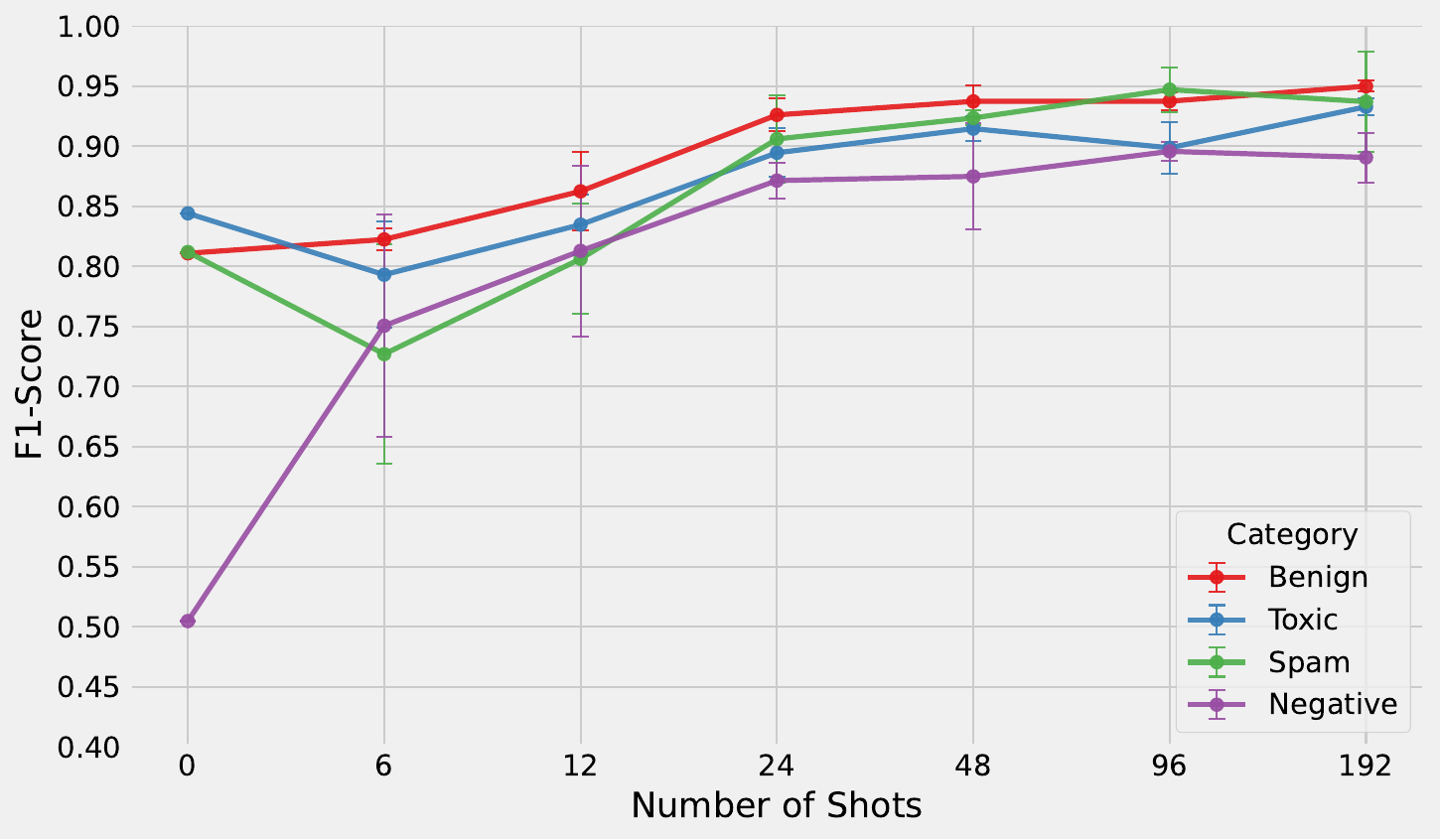}
        \caption{F1-score}
        \label{fig:multi-task_multi-class_llama_category_f1}
    \end{subfigure}
    \begin{subfigure}{0.32\textwidth}
        \centering
        \includegraphics[width=\textwidth]{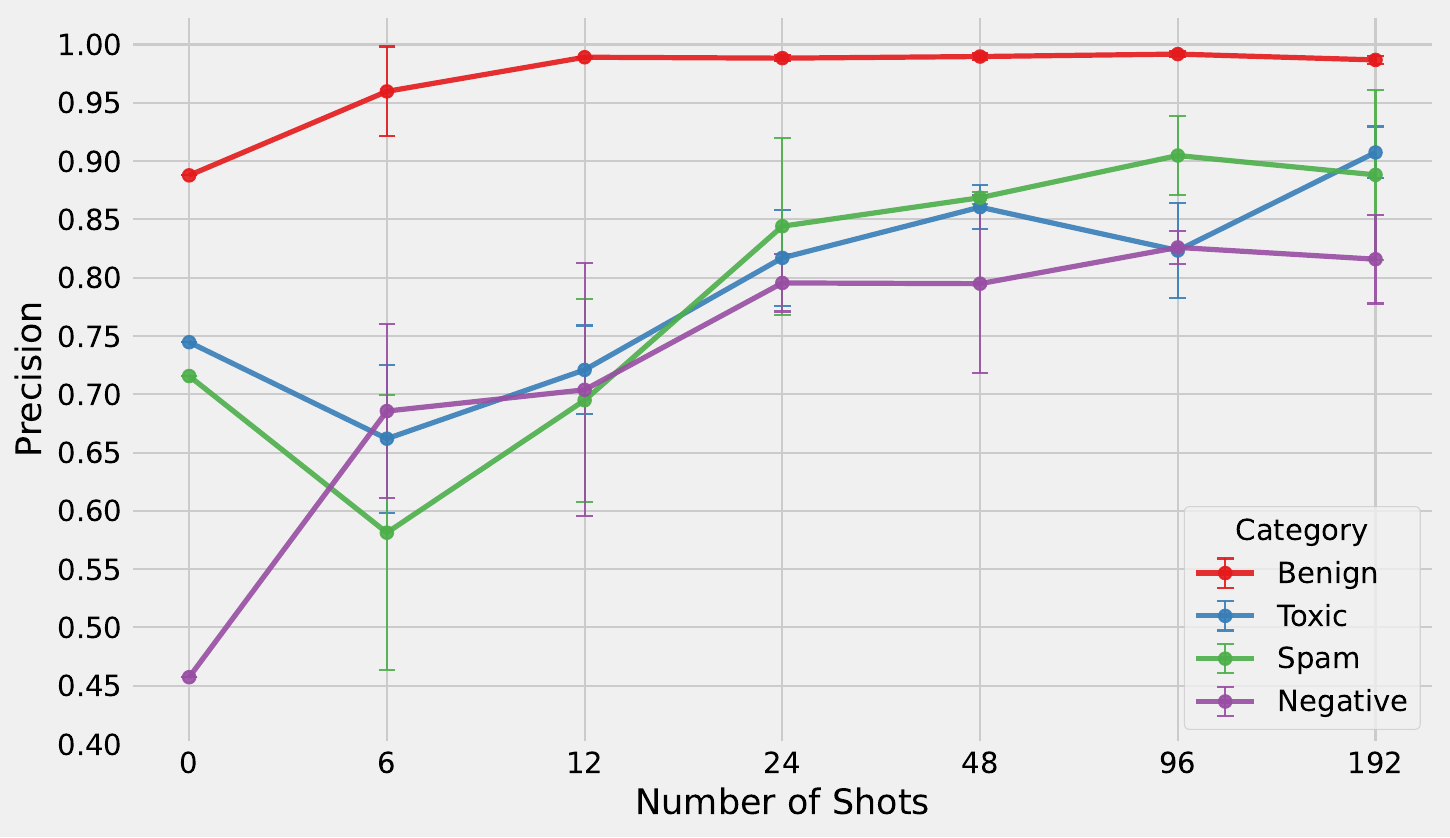}
        \caption{Precision}
        \label{fig:multi-task_multi-class_llama_category_precision}
    \end{subfigure}
    \begin{subfigure}{0.32\textwidth}
        \centering
        \includegraphics[width=\textwidth]{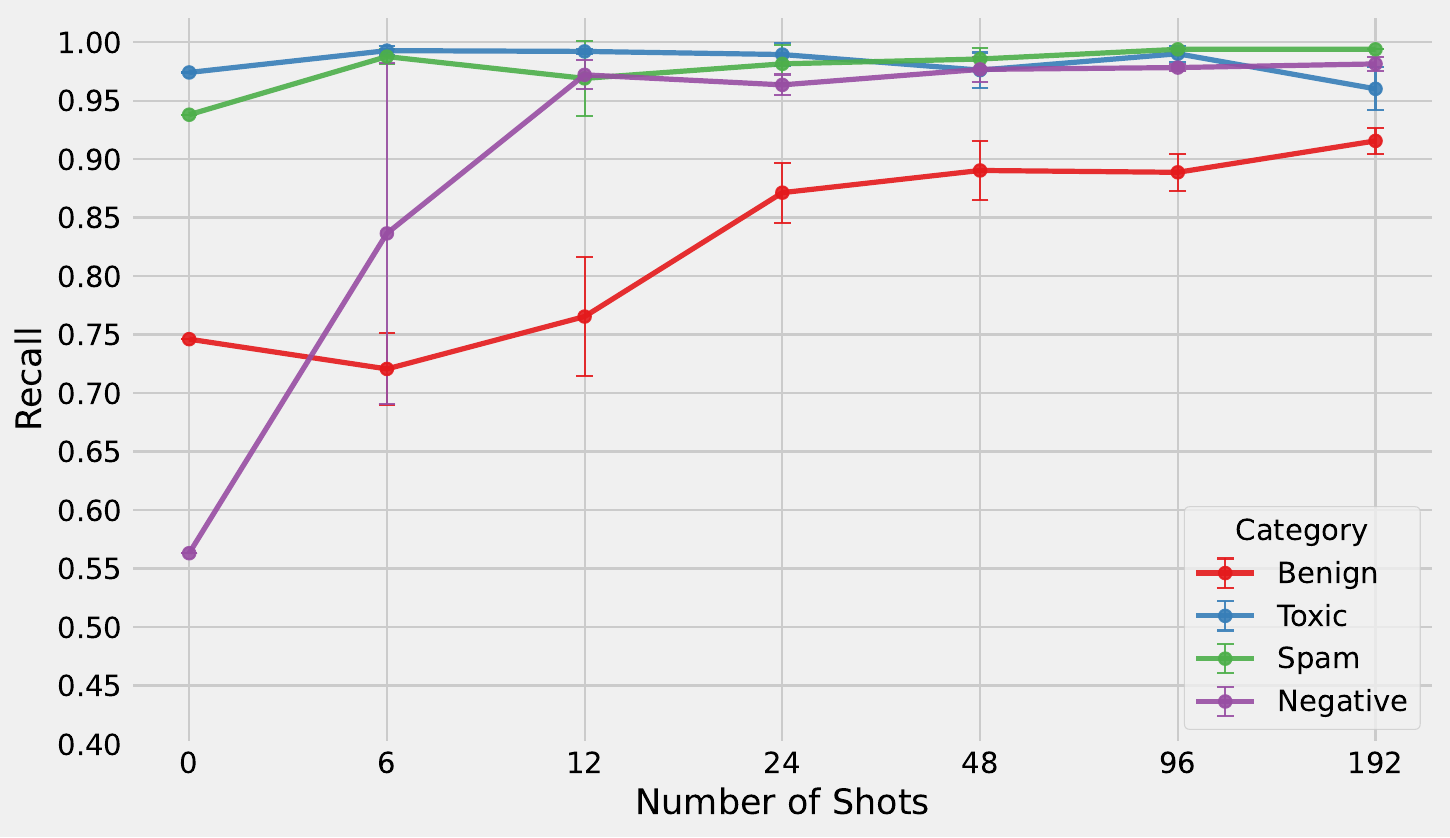}
        \caption{Recall}
        \label{fig:multi-task_multi-class_llama_category_recall}
    \end{subfigure}
    \caption{The subcategory-wise performance of the best-performing ICL setup (Llama-Random) in Multi-task Multi-class ICL. }
    \label{fig:multi-task_multi-class_llama_category}
\end{figure}

\begin{figure}[ht]
    \centering
    \begin{subfigure}{0.49\textwidth}
        \centering
        \includegraphics[width=\textwidth]{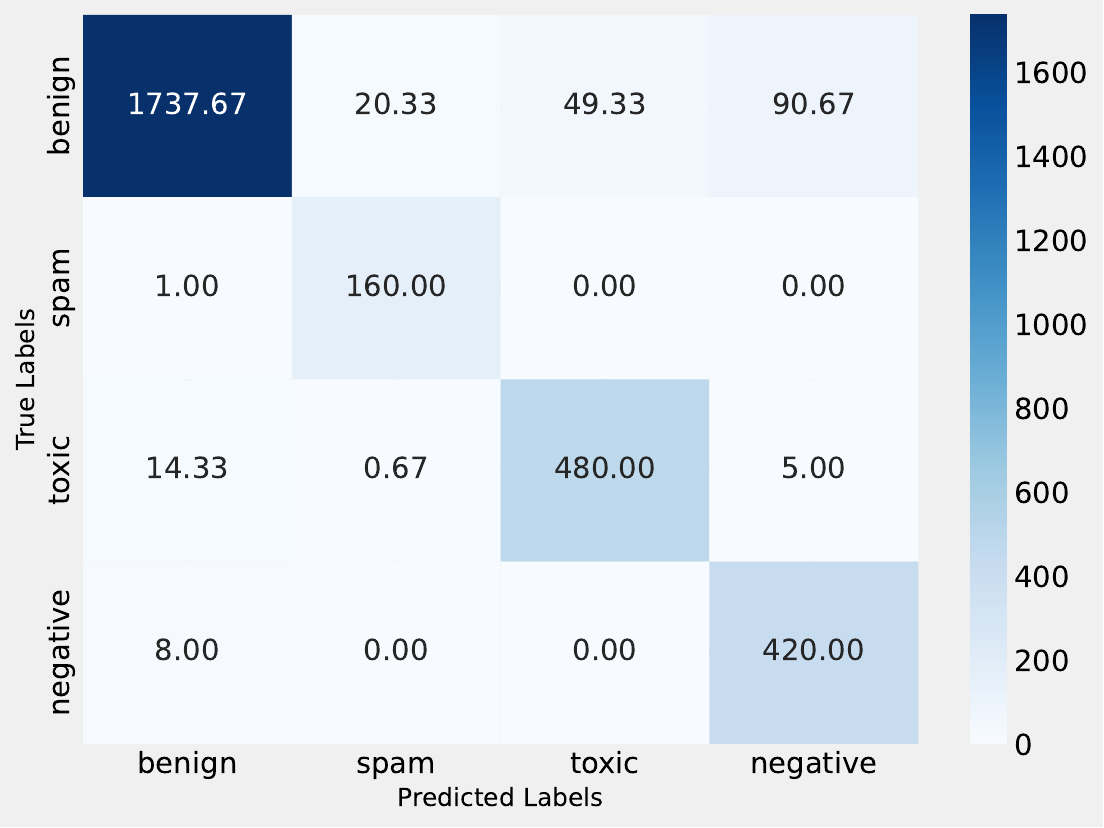}
        \caption{Level 1 Prompt}
    \end{subfigure}
    \begin{subfigure}{0.49\textwidth}
        \centering
        \includegraphics[width=\textwidth]{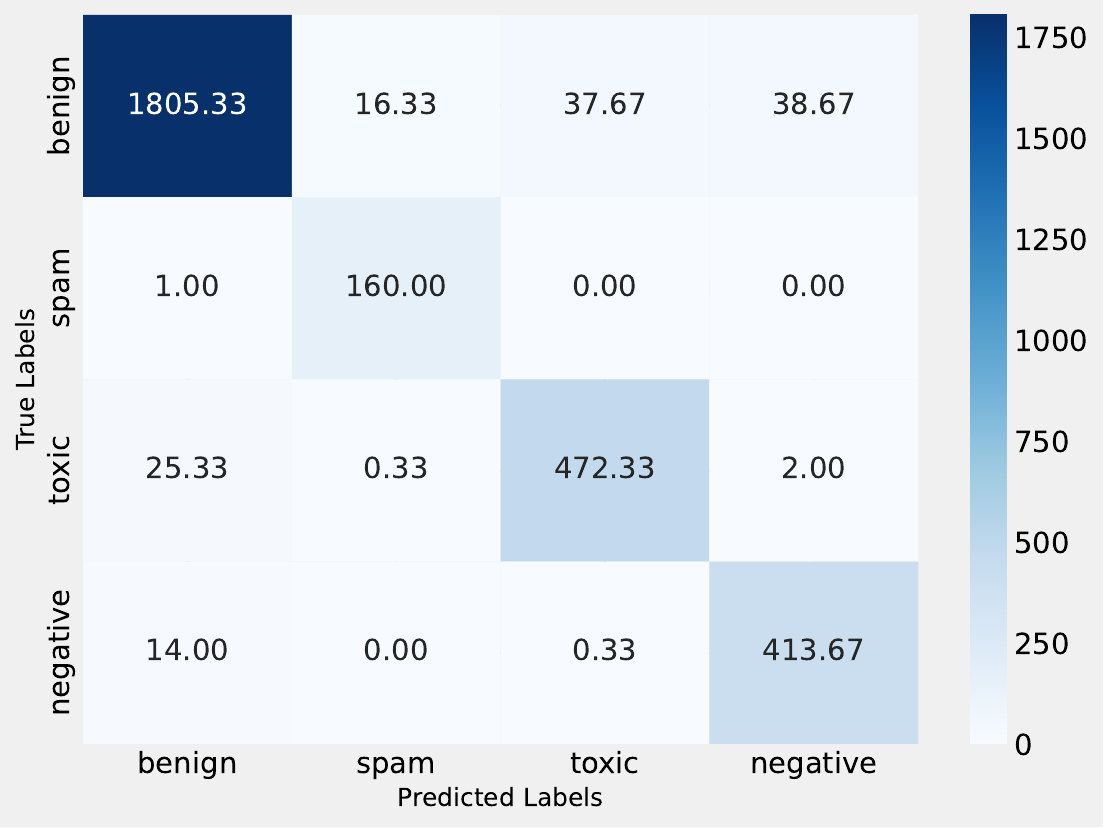}
        \caption{Level 2 Prompt}
    \end{subfigure}
    \caption{The Confusion Matrix of Llama-Random (192 shots) in Multi-task Multi-class ICL}
    \label{fig:multi-task_multi-class_llama_cm192}
\end{figure}

\subject{Fine-grained analysis of the best-performing ICL setup: Llama-Random.}  
Figure~\ref{fig:multi-task_multi-class_llama_category_f1} shows that F1-scores for all categories improve with larger shot counts and eventually stabilize. However, the \textit{negative} category lags behind others. From Figures~\ref{fig:multi-task_multi-class_llama_category_precision} and \ref{fig:multi-task_multi-class_llama_category_recall}, we attribute this to its low precision. In contrast, the benign category shows very high precision but relatively low recall.  
Examining the confusion matrix (Figure~\ref{fig:multi-task_multi-class_llama_cm192}), we find that misclassified harmful texts (toxic, spam, negative) are more often predicted as benign rather than another harmful type. This indicates that the model effectively distinguishes different harmful categories once harmfulness is recognized, but it struggles more in deciding whether a text is harmful at all. 

Moreover, more than half of the misclassified benign texts are predicted as negative, which explains the low precision of the negative category. This aligns with the intuition that negative sentiment is less saliently harmful compared to overt toxicity or spam, making it more semantically ambiguous and confusable with benign content. This inherent challenge underscores the difficulty of fine-grained harm classification and motivates the need for the more flexible multi-label formulation and a more strictly curated dataset we explore in Section~\ref{sec:wild_data}.

\subject{Comparison between Level 1 and Level 2 task descriptions.}  
Building on the success of Level 2 prompts in the binary setting, we evaluate their impact on the multi-class task. As shown in Figure 13, incorporating Level 2 task descriptions significantly boosts ICL performance when \(k \geq 12\).
The weighted average F1-score improves from 0.938 (192 shots) with Level 1 to 0.955 (192 shots) with Level 2. Comparing the confusion matrices in Figure~\ref{fig:multi-task_multi-class_llama_cm192}, we find that Level 2 task descriptions notably reduce the false positive rate, at the cost of a slight increase in false negatives. Importantly, the model still maintains strong discrimination among harmful categories. This suggests that richer label definitions help the model refine its decision boundary between harmful and benign texts, thereby improving robustness in the multi-class setting.


\section{Personalization with ICL}
\label{sec:personalization}



Having established the strong multi-task performance of ICL for harmful content detection, we now investigate its potential for \textbf{personalization}. A key limitation of centralized moderation systems is their "one-size-fits-all" approach, which fails to accommodate diverse user preferences and contexts. We therefore address a critical question: \textit{Can ICL be efficiently personalized to align with individual user preferences for what constitutes harmful content, without compromising its core detection capabilities on non-personalized categories?}

Affirmative answers would position ICL as a cornerstone for transparent, user-controlled, and on-device moderation systems. To this end, we design and evaluate three personalization scenarios that simulate real-world user interactions: blocking a new category of harmful texts (Section~\ref{subsec:blocking-new-category}), unblocking a known category (Section~\ref{subsec:unblocking-known-category}), and blocking semantic variations of a specific harmful instance (Section~\ref{subsec:blocking-variations}).

\subject{General Evaluation Setup.} In these experiments, the \textit{Multi-Task Binary ICL} setting is adopted, which most closely reflects real deployment conditions. Based on the results in Section~\ref{sec:multi-task_binary}, we adopt Llama-Random as the representative configuration for all experiments. The overall setup is the same as in the multi-task binary case, with minor adjustments tailored to each personalization scenario.

\subsection{Blocking a New Category of Harmful Texts}
\label{subsec:blocking-new-category}

\subject{Motivation.} We first consider the scenario where users encounter a new harmful category not present in the training pool. To simulate this, we remove the TextDetox dataset from the training stage and treat it as a new harmful category encountered at test time. This allows us to examine whether ICL can rapidly adapt to unseen harmful types with minimal supervision. For Level 1 prompts, we directly inject a small number of new examples ($k_2 \in \{0, 1,2,4,8\}$). For Level 2 prompts, we additionally extend the task description with a definition of the new harmful type.

\subject{Experimental Setup.} The experimental setup is designed to simulate a user introducing a previously unseen harmful category into an existing moderation system.

\begin{itemize}
    \item \textbf{Datasets and Simulation:} We treat toxicity (the \textbf{TextDetox} dataset) as the \textit{new, user-defined harmful category}. To simulate a system that has not encountered this category before, we \textbf{remove all TextDetox samples from the main demonstration pool $D^r_k$}. The test set $\mathcal{X}$ remains the union of all three datasets ($D^{TextDetox}_{test} \cup D^{UCI}_{test} \cup D^{SST2}_{test}$). The main demonstration pool is thus $D^r_k \subseteq D^{UCI}_{train} \cup D^{SST2}_{train}$. The personalization is achieved by injecting a small number of TextDetox harmful samples into the prompt.
    \item \textbf{Shots:} The total number of shots is $k = k_1 + k_2$. Here, $k_1$ (drawn from UCI and SST2) represents the \textit{base system knowledge} ($k_1 \in \{0,4,8,16,32,64,128\}$), while $k_2$ (drawn from TextDetox harmful samples) represents the \textit{user-provided personalization examples} ($k_2 \in \{0,1,2,4,8\}$).
\end{itemize}

\begin{figure}[ht]
    \centering
    \begin{subfigure}{0.49\textwidth}
        \centering
        \includegraphics[width=\textwidth]{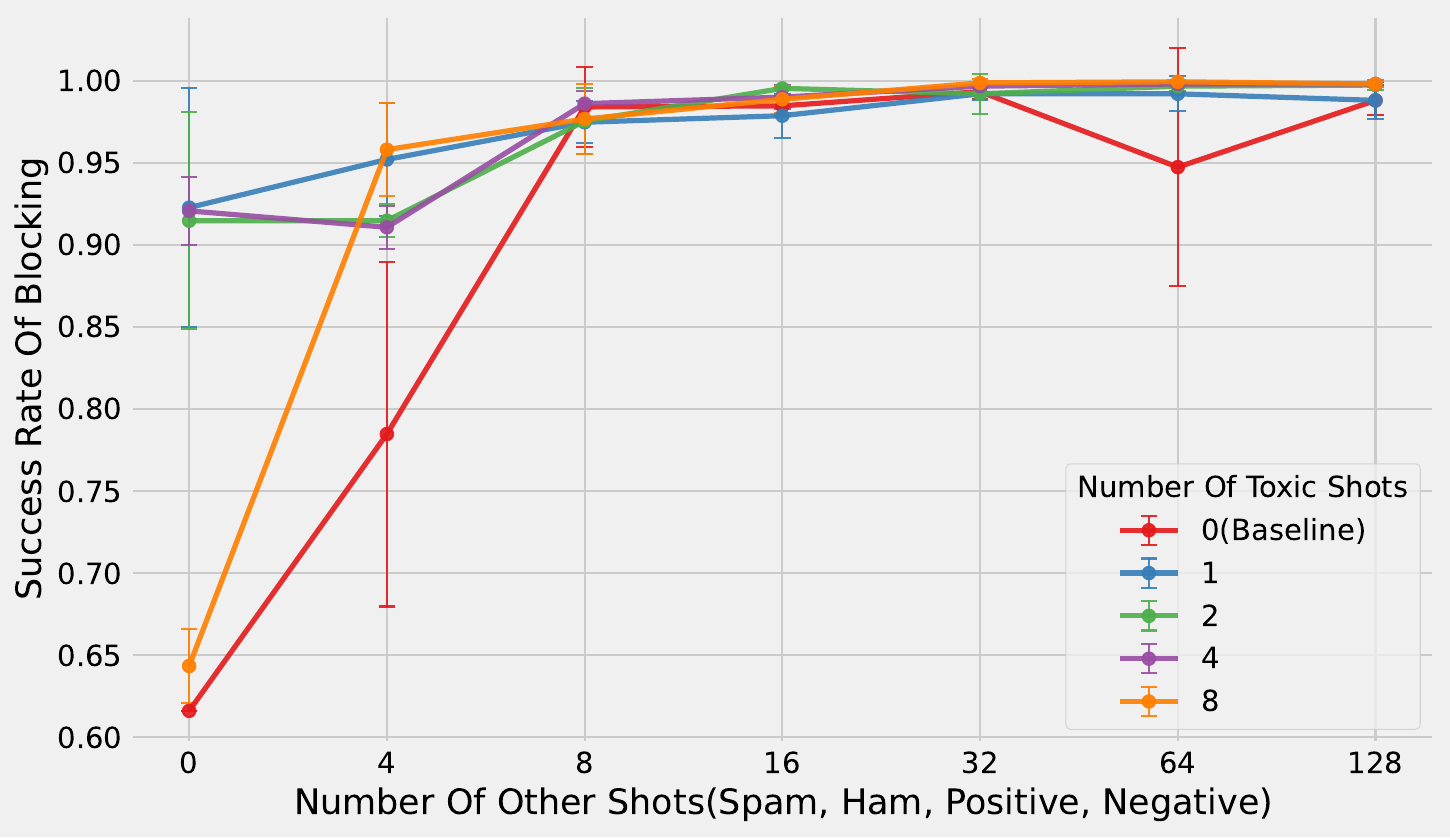}
        \caption{Level 1 Prompt}
        \label{fig:per_block_new_succ_rate_l1}
    \end{subfigure}
    \begin{subfigure}{0.49\textwidth}
        \centering
        \includegraphics[width=\textwidth]{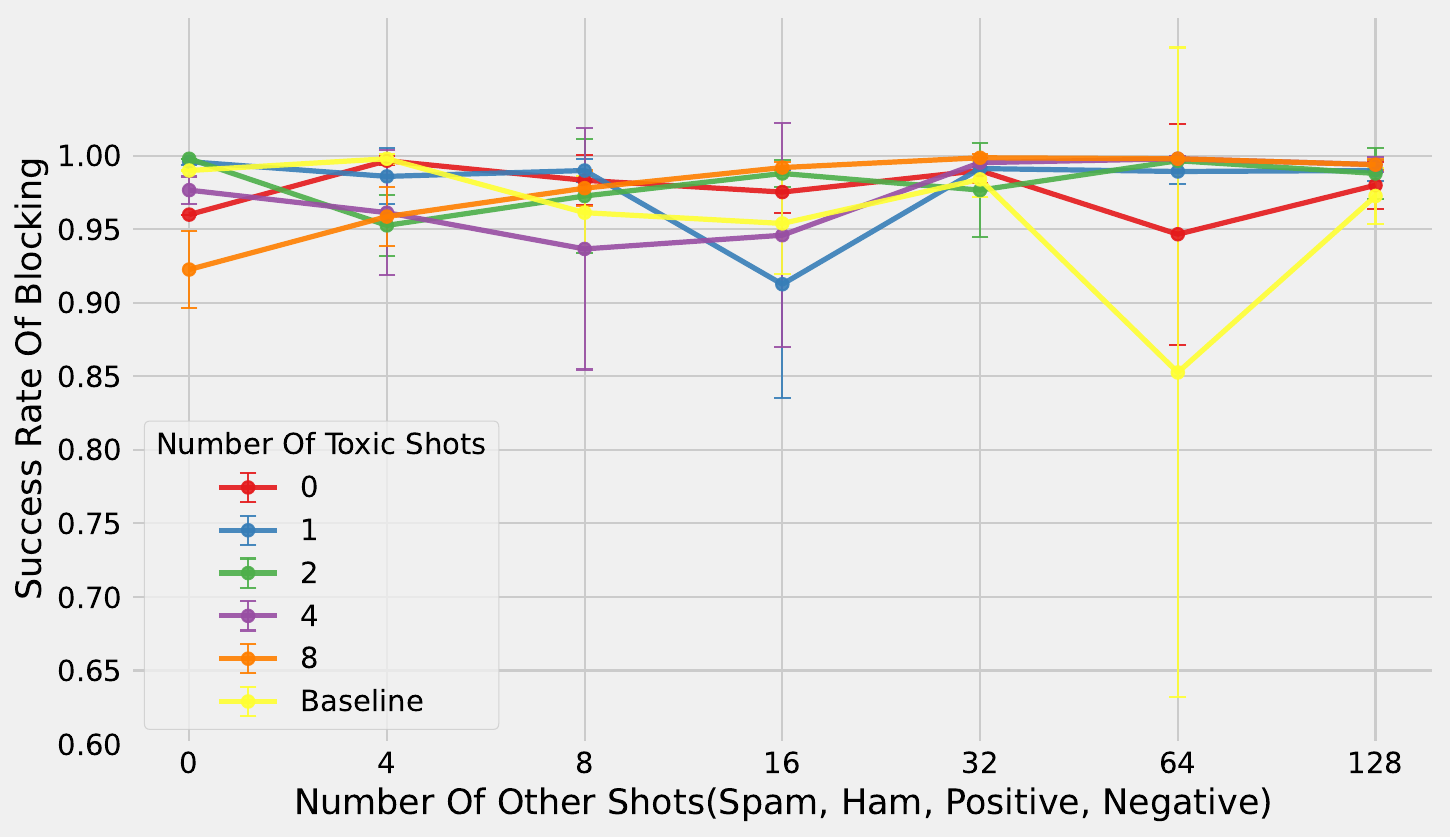}
        \caption{Level 2 Prompt}
        \label{fig:per_block_new_succ_rate_l2}
    \end{subfigure}
    \caption{\textbf{The Performance of Blocking New Harmful Texts.} Note: \textbf{Baseline} represents use original Multi-Task Binary ICL(UCI, SST2) to detect Textdetox.}
    \label{fig:per_block_new_succ_rate}
\end{figure}

\subject{Results.}  
As shown in Figure~\ref{fig:per_block_new_succ_rate_l1}, under Level 1 prompts, the model already achieves high blocking success ($>$0.98) when $k_1$ is sufficiently large, even without any TextDetox demonstrations. This indicates that ICL generalizes well across harmful types by exploiting shared harmfulness cues. However, success rates fluctuate more when no TextDetox shots are provided, highlighting the stabilizing role of even a single example. Adding 1--2 examples yields a sharp improvement when $k_1$ is small, but further increases ($k_2=4$ or $8$) bring only marginal gains.  
Level 2 prompts (Figure~\ref{fig:per_block_new_succ_rate_l2}) further enhance performance when $k_1$ is small, since explicit definitions compensate for limited demonstrations. Once $k_1 \geq 16$, the effect of definitions diminishes, as the model already learns sufficient cross-category generalizations.

\begin{figure}[ht]
    \centering
    \begin{subfigure}{0.49\textwidth}
        \centering
        \includegraphics[width=\textwidth]{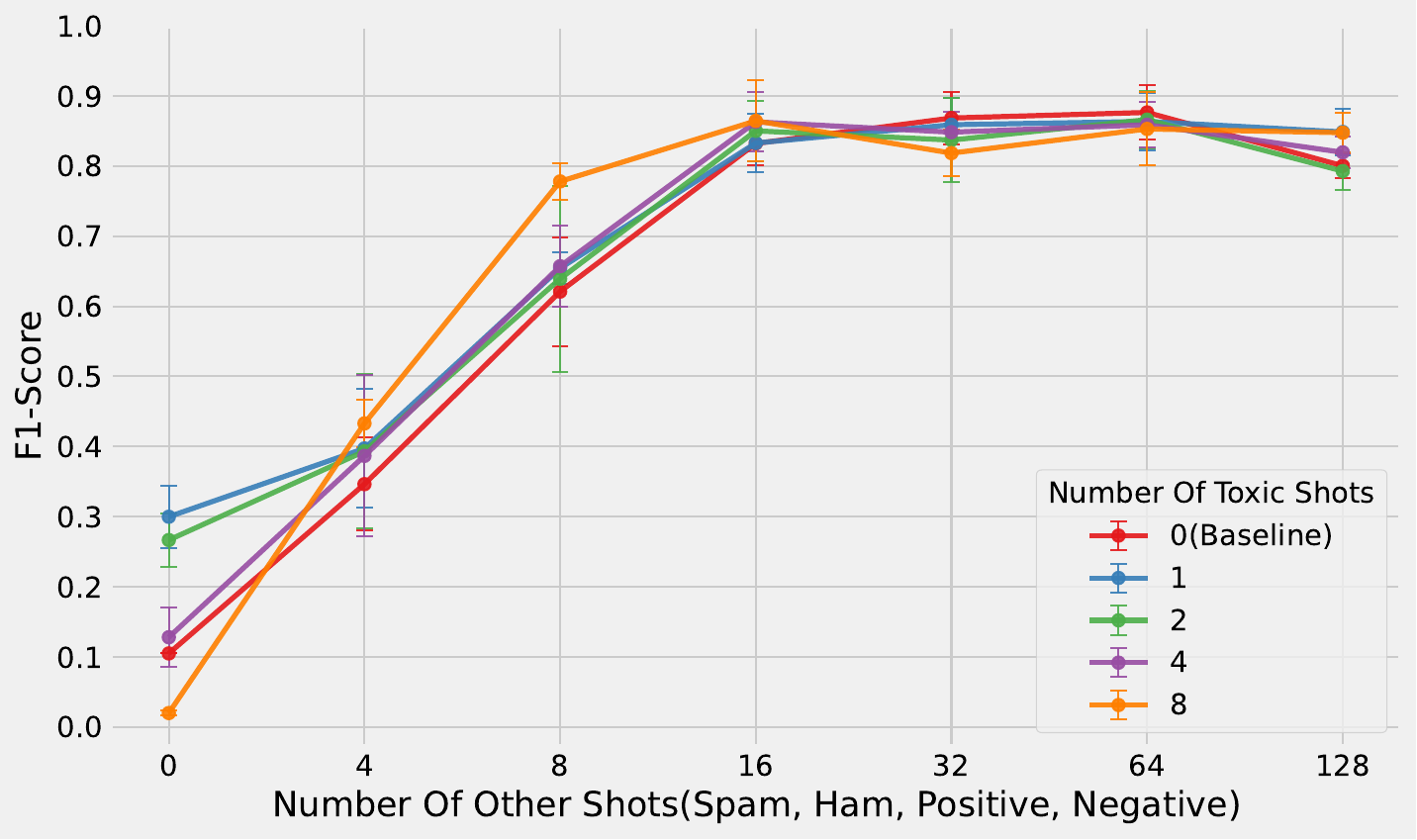}
        \caption{Level 1 Prompt}
        \label{fig:per_block_new_origin_data_l1}
    \end{subfigure}
    \begin{subfigure}{0.49\textwidth}
        \centering
        \includegraphics[width=\textwidth]{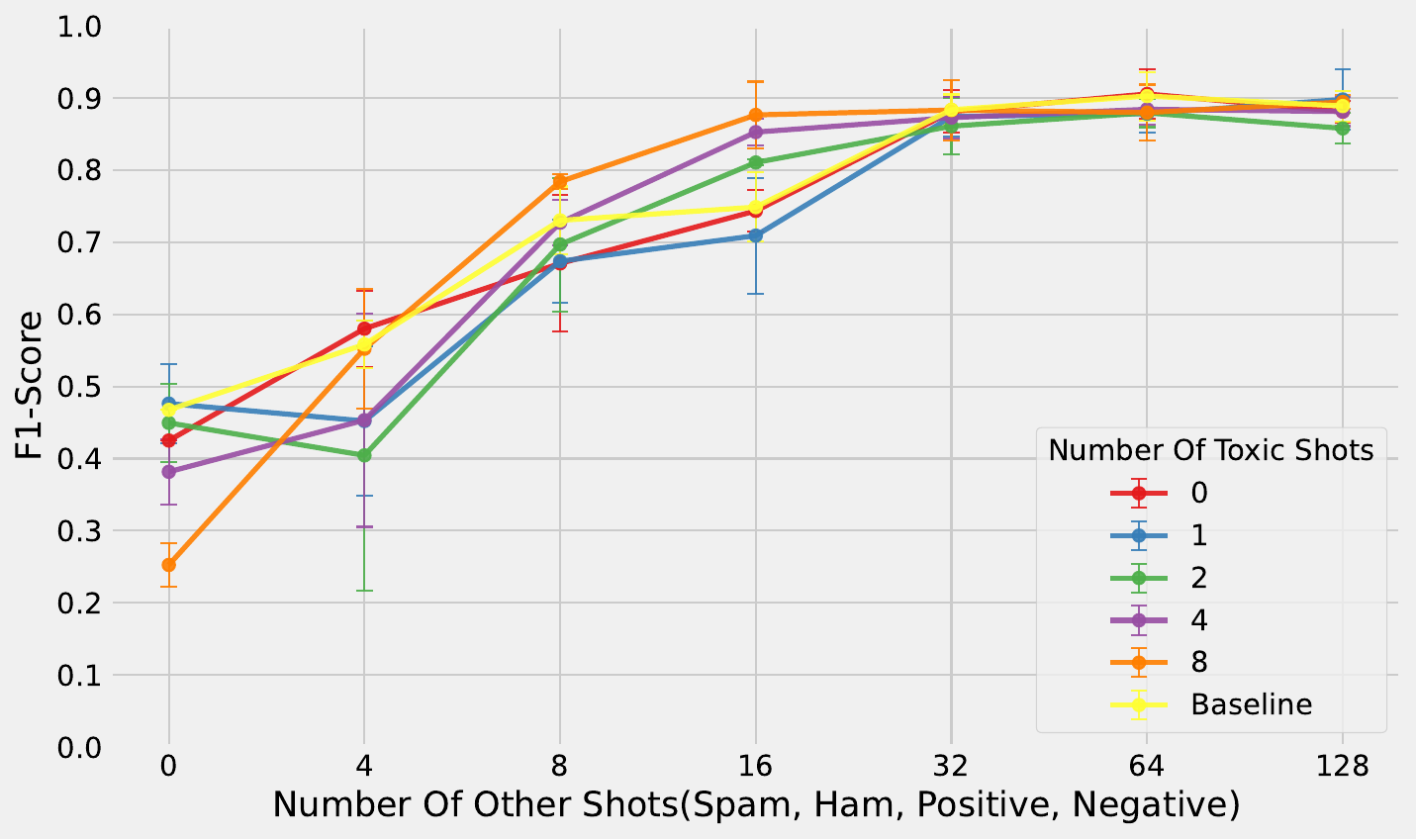}
        \caption{Level 2 Prompt}
        \label{fig:per_block_new_origin_data_l2}
    \end{subfigure}
    \caption{The performance of ICL on known categories of harmful texts, when a new category is added. Here, the known categories are set as spam and negative sentiment, while the newly added category is toxicity. }
    \label{fig:per_block_new_origin_data}
\end{figure}

\subject{Impact on Known Categories.}  
Figure~\ref{fig:per_block_new_origin_data} shows that performance on UCI and SST2 remains stable regardless of TextDetox additions or task description modifications, especially when $k_1 \geq 32$. This suggests that personalization to block new harmful categories does not degrade recognition of original categories.

\subject{Summary.}  
ICL exhibits strong generalization to new harmful categories, requiring minimal supervision to achieve stable performance. Definitions in prompts are particularly useful when shot counts are small, but their impact diminishes once sufficient demonstrations are provided. Importantly, personalization does not compromise performance on existing datasets.

\subsection{Unblocking a Known Category of Harmful Texts}
\label{subsec:unblocking-known-category}
\subject{Motivation.} Different users may not fully agree on what constitutes harmful content. To simulate this, we treat TextDetox as a harmful type that certain users wish to \textit{unblock}. For Level 1 prompts, we add a small number of TextDetox samples labeled as benign. For Level 2 prompts, we additionally redefine toxicity as benign in the task description.

\subject{Experimental Setup.}  Below summarizes the experimental setup: 
\begin{itemize}
    \item \textbf{Datasets:} Same as before, but TextDetox is redefined as benign. Only harmful samples are used.  
    \item \textbf{Shots:} $k = k_1 + k_2$, where $k_1 \in \{0,4,8,16,32,64,128\}$ (UCI, SST2) and $k_2 \in \{0,1,2,4,8\}$ (TextDetox, re-labeled as benign).
\end{itemize}

\begin{figure}
    \centering
    \begin{subfigure}{0.49\textwidth}
        \centering
        \includegraphics[width=\textwidth]{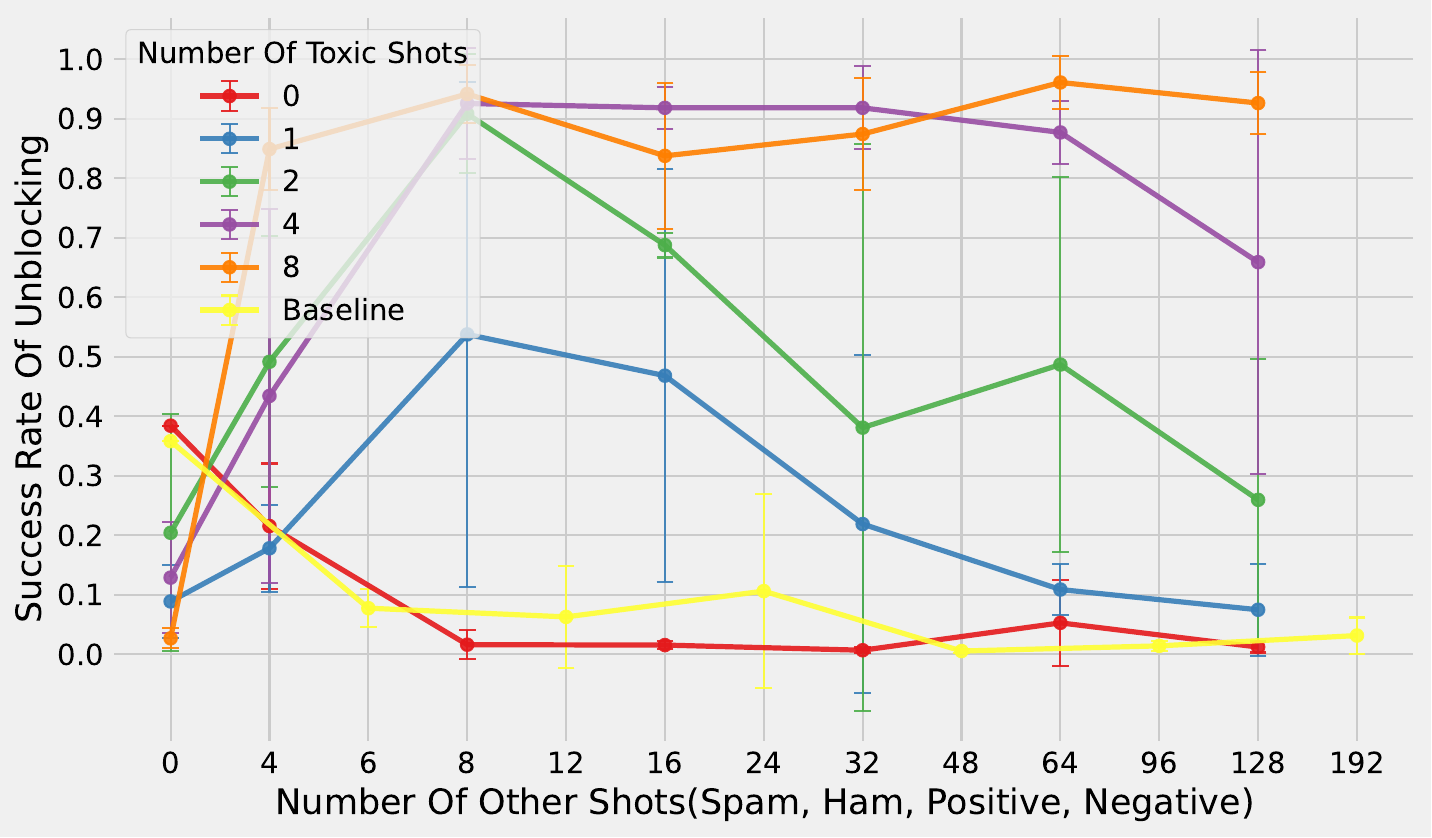}
        \caption{Level 1 Prompt}
        \label{fig:per_unblock_succ_rate_l1}
    \end{subfigure}
    \begin{subfigure}{0.49\textwidth}
        \centering
        \includegraphics[width=\textwidth]{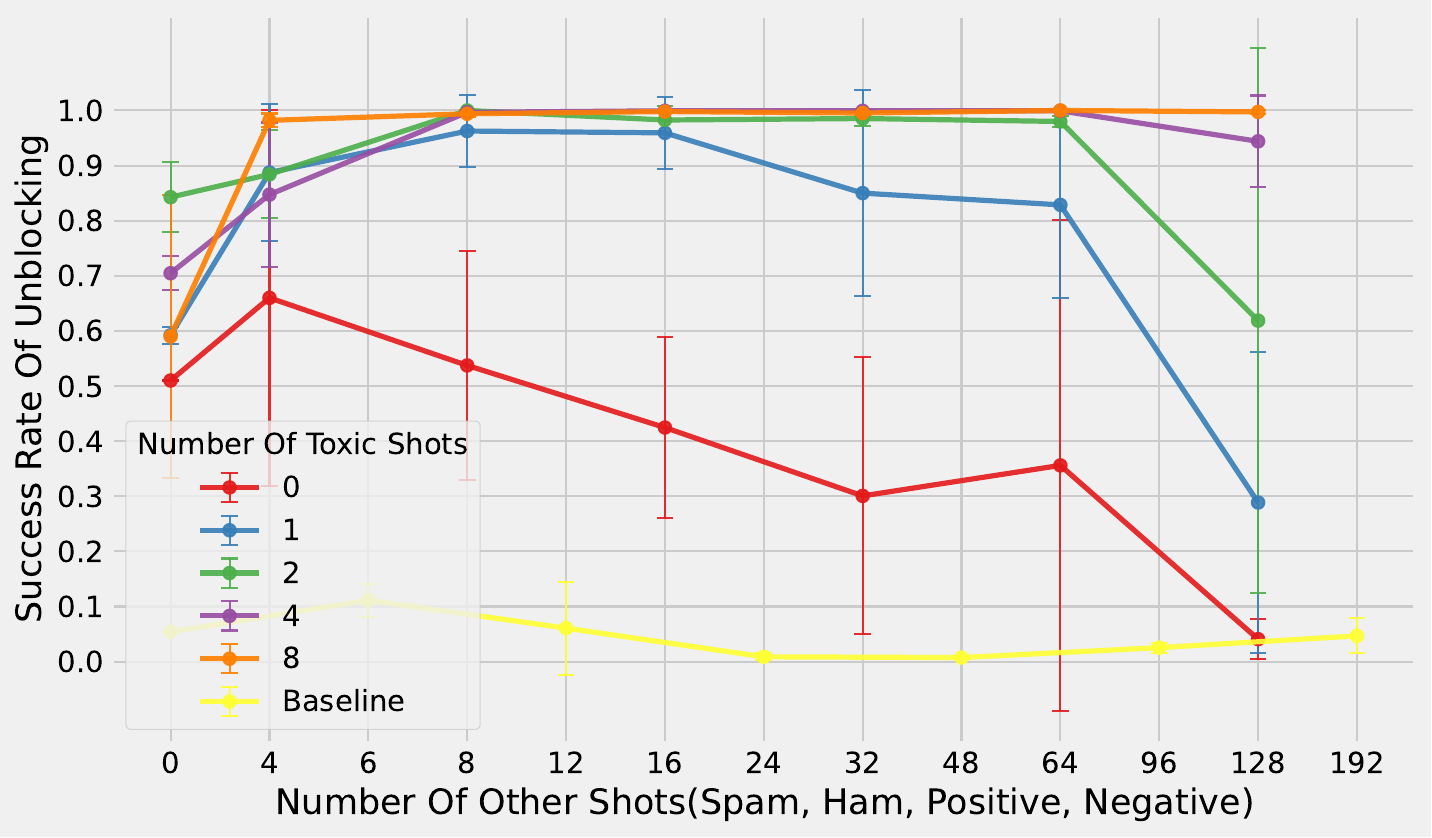}
        \caption{Level 2 Prompt}
        \label{fig:per_unblock_succ_rate_l2}
    \end{subfigure}
    \caption{\textbf{The Performance of unblocking a known category of harmful texts.} Note: \textbf{Baseline} represents using original Multi-Task Binary ICL(Textdetox, UCI, SST2) to unblock Textdetox, wherein in-context toxic demonstrations are labeled as positive (harmful). }
    \label{fig:per_unblock_succ_rate}
\end{figure}

\subject{Results.}  As shown in Figure~\ref{fig:per_unblock_succ_rate_l1}, the model's baseline behavior (with $k_2=0$) is to block toxic content, correctly classifying it as harmful based on its pre-existing knowledge. Introducing a small number of re-labeled examples ($k_2=1$ or $2$) successfully \textit{unblocks} the category when the base knowledge $k_1$ is small. However, as $k_1$ grows, the model is exposed to more examples reinforcing the original "toxicity = harmful" association, which begins to overwhelm the sparse personalization signal, causing the unblocking success rate to drop. This indicates a \textbf{tension between general harmfulness knowledge and specific user overrides}. Crucially, with a stronger personalization signal ($k_2 \geq 4$), the model reliably unblocks the category, maintaining a high success rate (e.g., $>0.93$ at $k_1=128$), demonstrating that persistent user input can effectively recalibrate the model's behavior.

Level 2 prompts (Figure~\ref{fig:per_unblock_succ_rate_l2}) alleviate this issue further by explicitly redefining toxicity. Specifically, Level 2 prompt performs better than level 1 prompt in the following aspects: Firstly, due to the explicit definition of toxicity and the clear indication that toxic texts should be considered benign, ICL achieves a certain level of unblocking effectiveness even without the inclusion of toxic shots—outperforming the level 1 prompt with one toxic text added. Then, when the same number of toxic texts is introduced, the success rate of unblocking is consistently higher for level 2 than for level 1. Moreover, the decline in the success rate with increasing other shots is less pronounced. Even when $k1$ reaches 128 and $k2$ = 8, the success rate reaches approximately 0.99.

\begin{figure}
    \centering
    \begin{subfigure}{0.49\textwidth}
        \centering
        \includegraphics[width=\textwidth]{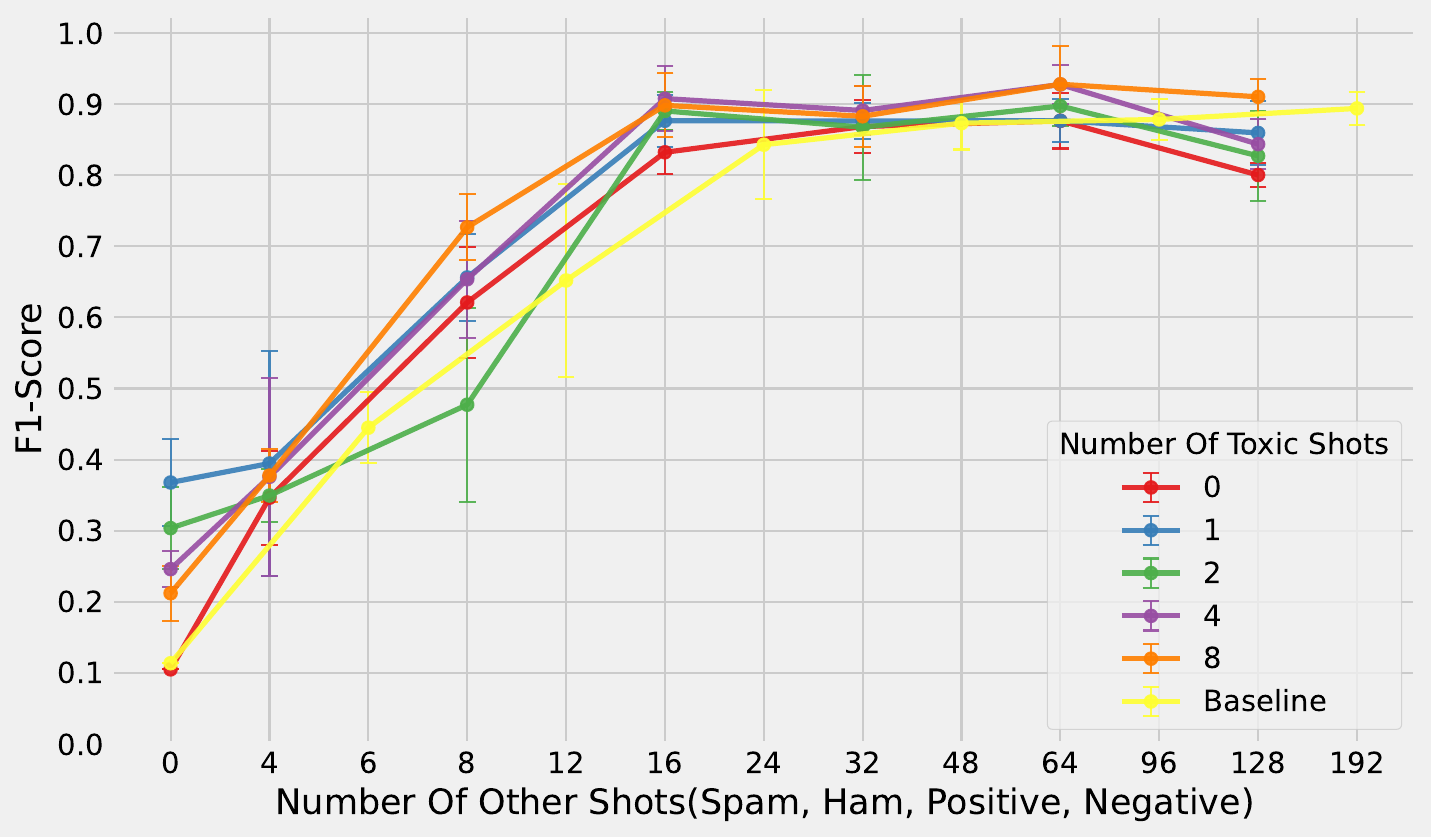}
        \caption{Level 1 Prompt}
        \label{fig:per_unblock_origin_data_l1}
    \end{subfigure}
    \begin{subfigure}{0.49\textwidth}
        \centering
        \includegraphics[width=\textwidth]{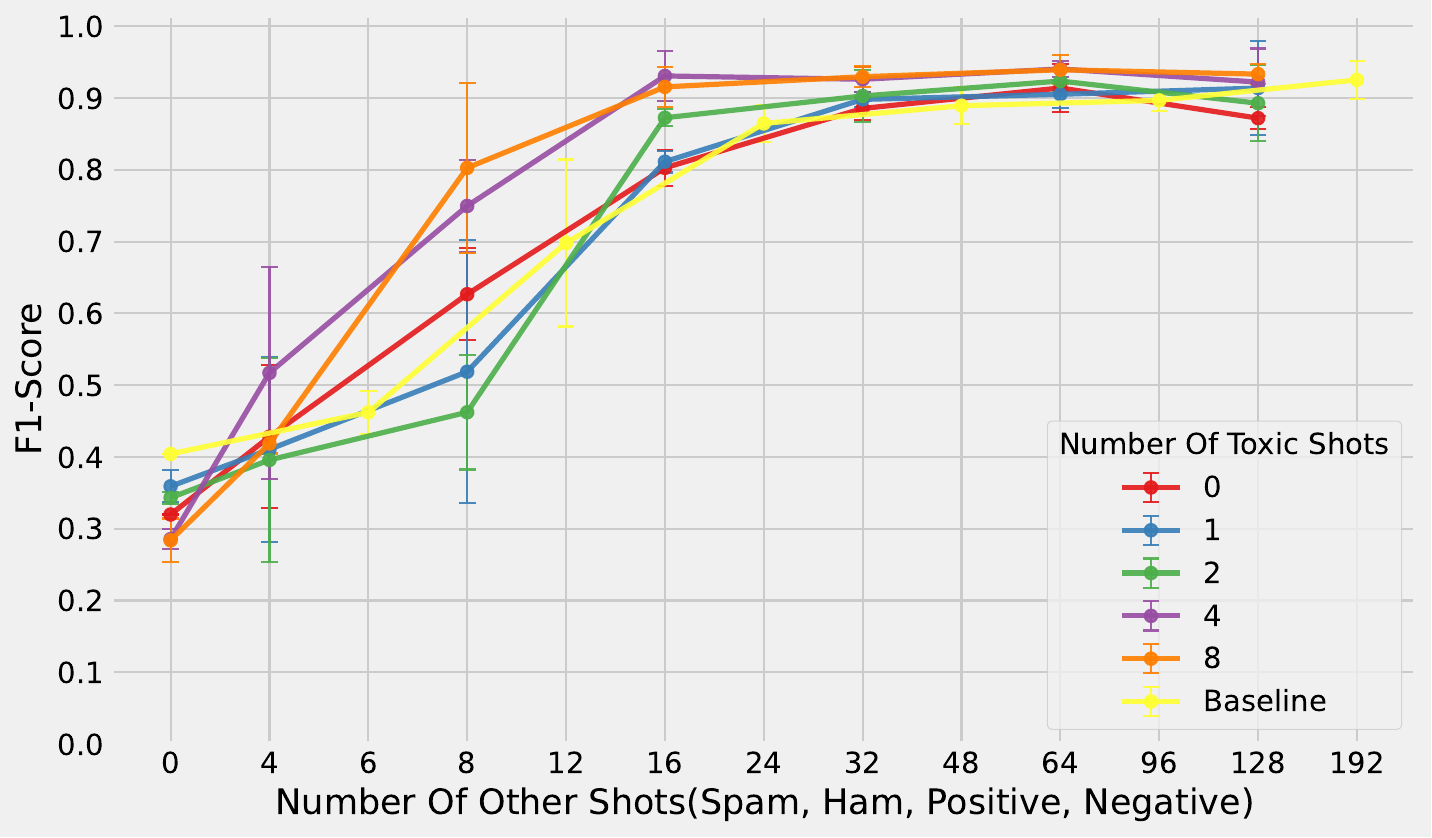}
        \caption{Level 2 Prompt}
        \label{fig:per_unblock_origin_data_l2}
    \end{subfigure}
    \caption{The F1-score Performance of ICL on the left blocking categories, when unblocking a known category of harmful texts. }
    \label{fig:per_unblock_origin_data}
\end{figure}

\begin{figure}
    \centering
    \begin{subfigure}{0.49\textwidth}
        \centering
        \includegraphics[width=\textwidth]{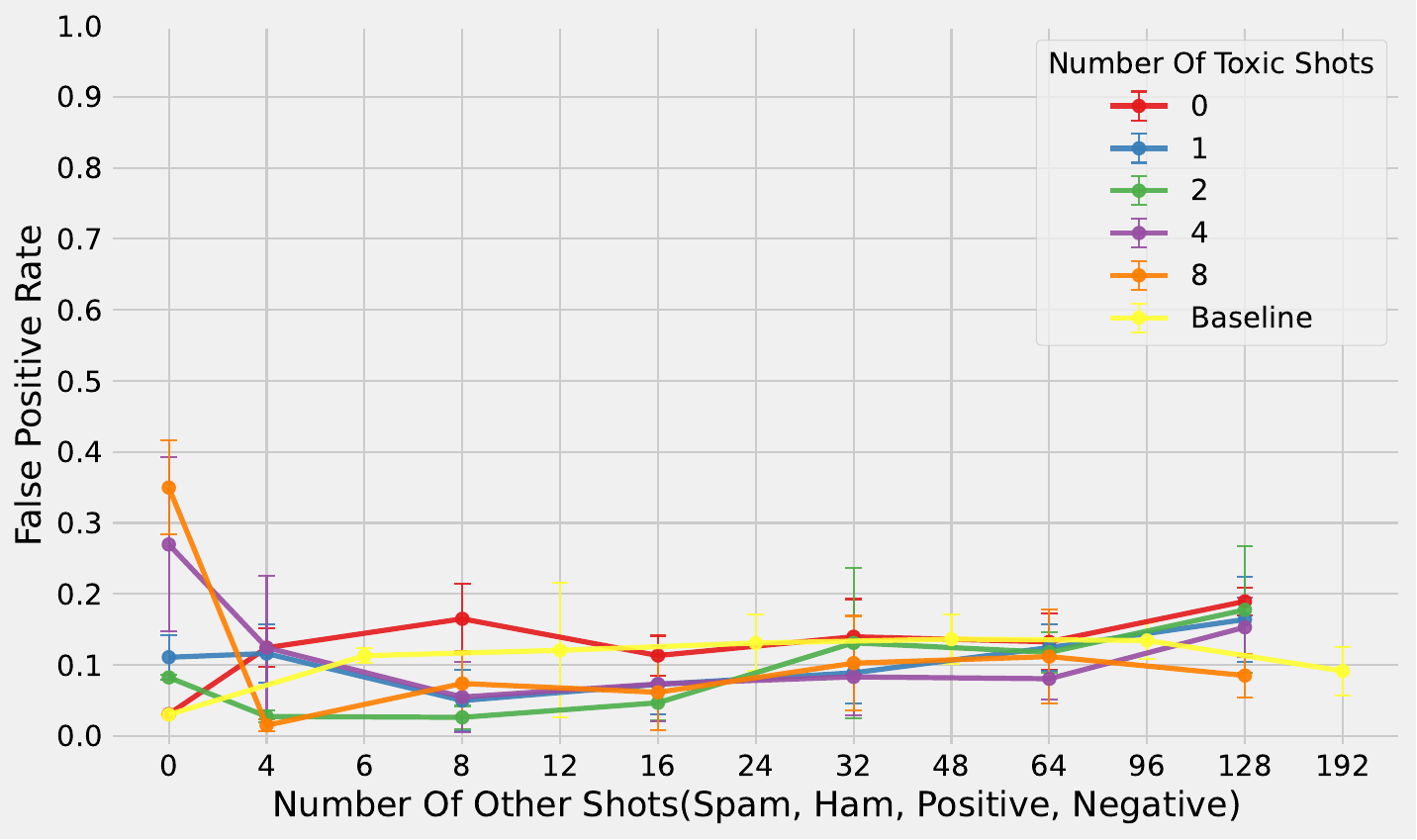}
        \caption{Level 1 Prompt}
        \label{fig:per_unblock_fpr_l1}
    \end{subfigure}
    \begin{subfigure}{0.49\textwidth}
        \centering
        \includegraphics[width=\textwidth]{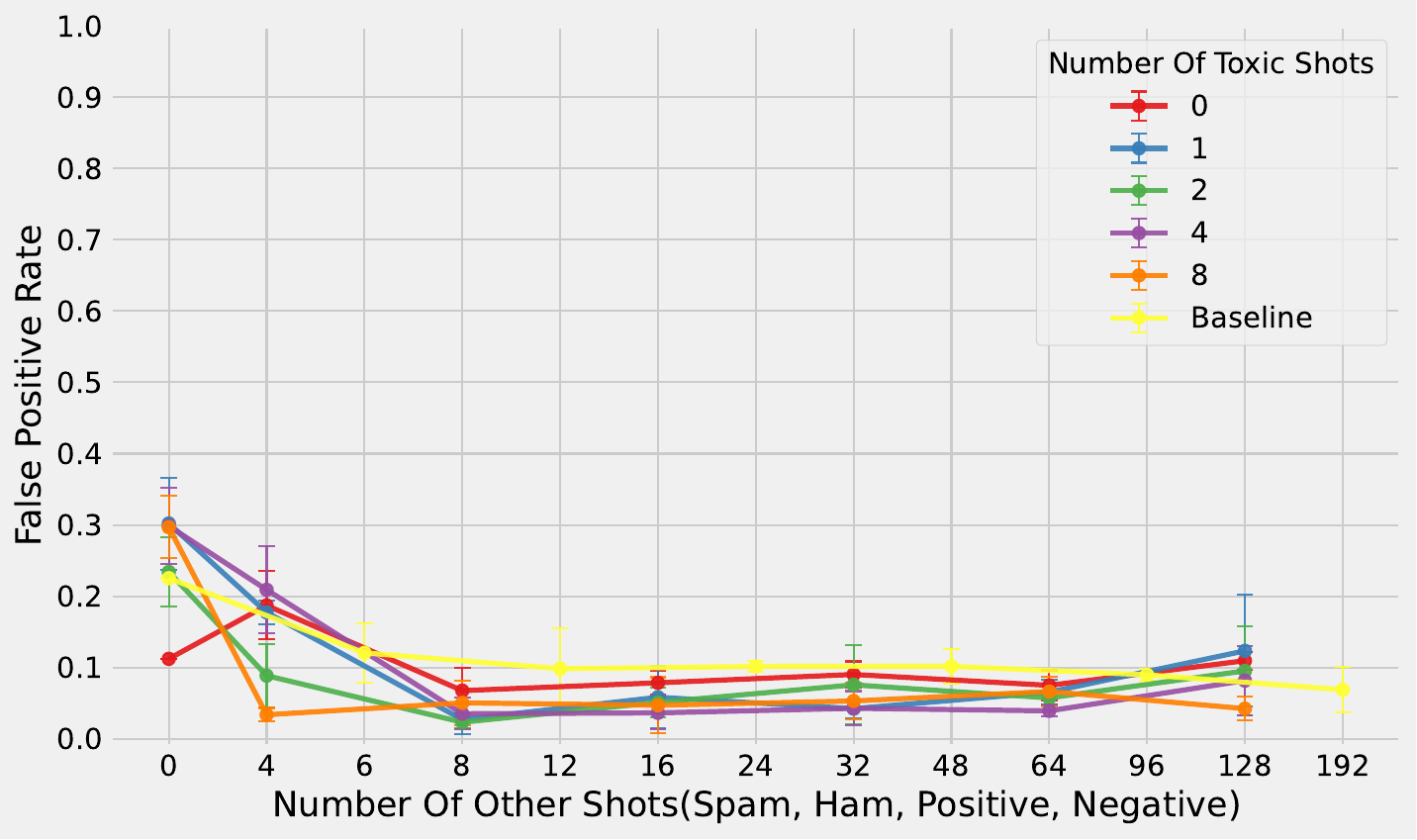}
        \caption{Level 2 Prompt}
        \label{fig:per_unblock_fpr_l2}
    \end{subfigure}
    \caption{The false positive rate of ICL on the left blocking categories, when a known category of harmful texts is unblocked.}
    \label{fig:per_unblock_fpr}
\end{figure}

\subject{Impact on the Left Blocking Categories.}  
As shown in Figure~\ref{fig:per_unblock_origin_data_l1}, performance on UCI and SST2 is preserved or even slightly improved when TextDetox is re-labeled as benign.  We hypothesize that this is due to the following reason: In Multi-Task Binary ICL, we previously noted that a key limitation of ICL is its high false positive rate. The addition of toxic texts labeled as benign encourages ICL to adopt a more conservative decision-making strategy for harmful classification, thereby reducing false positives. This is confirmed by Figure~\ref{fig:per_unblock_fpr}.

\subject{Summary.}  
Without affecting the performance on the original dataset (the left blocking categories), ICL can be flexibly adapted to user preferences for unblocking harmful categories, using only a handful of re-labeled demonstrations or explicit definitions. This demonstrates a form of \textbf{contextual unlearning}, where the model suppresses a previously learned association within the specific context of a user's prompt.

\subsection{Blocking Variations of a Newly Annotated Harmful Text}
\label{subsec:blocking-variations}
\subject{Motivation.}  
In many cases, users cannot articulate a new harmful category but simply wish to block future texts resembling a specific harmful instance they encountered. To simulate this, we randomly select 100 harmful texts from each dataset and generate perturbed variants using EDA~\cite{wei2019edaeasydataaugmentation},  a technique for data augmentation (via TextAttack\footnote{https://textattack.readthedocs.io}). For each original text, we generate 12 perturbed variants \emph{per perturbation level} ($0.1$–$0.5$), resulting in a total of $60$ variants across five levels. At each level, four variants are used for training (i.e., demonstration selection) and eight for testing, forming the dataset $D^v$. The original text itself is always included in the prompt.

\subject{Experimental Setup.}  Below are the experimental setup for blocking variations of a newly annotated harmful text:
\begin{itemize}
    \item \textbf{Datasets:} $\mathcal{X} = D_{test}^{TextDetox} \cup D_{test}^{UCI} \cup D_{test}^{SST2} \cup D^v_{test}$, $D_k \subseteq D_{train}^{TextDetox} \cup D_{train}^{UCI} \cup D_{train}^{SST2} \cup D^v_{train}$.  
    \item \textbf{Shots:} $k = k_1 + k_2 + k_3$, where $k_1=196$ (original datasets), $k_2=1$ (the original harmful text), and $k_3 \in \{0,1,2,3,4\}$ (perturbed variants).
\end{itemize}

\subject{Results.}  We evaluate the success rate of blocking these perturbed variants. We define two key conditions: The "\textbf{Baseline}" condition, where the prompt contains \textit{neither} the original text nor its variants ($k_2=0, k_3=0$), tests the model's inherent ability to generalize from its base knowledge. The "\textbf{0-perturbation}" condition, where the original text is included ($k_2=1, k_3=0$), tests the utility of a single canonical example.

\begin{figure}
    \centering
    \begin{subfigure}{0.33\textwidth}
        \centering
        \includegraphics[width=\textwidth]{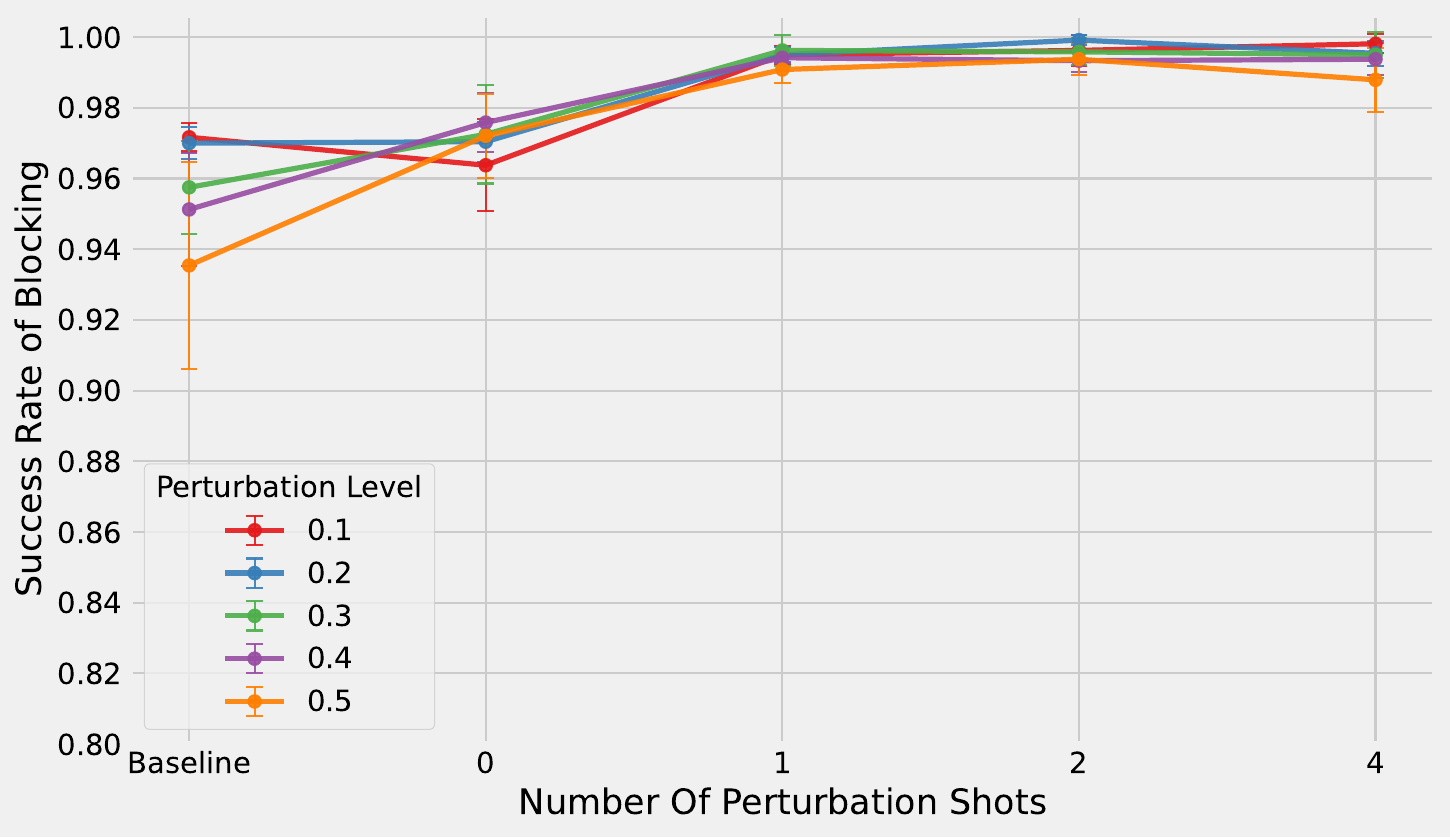}
        \caption{Toxic}
        \label{fig:per_block_var_textdetox}
    \end{subfigure}
    \begin{subfigure}{0.33\textwidth}
        \centering
        \includegraphics[width=\textwidth]{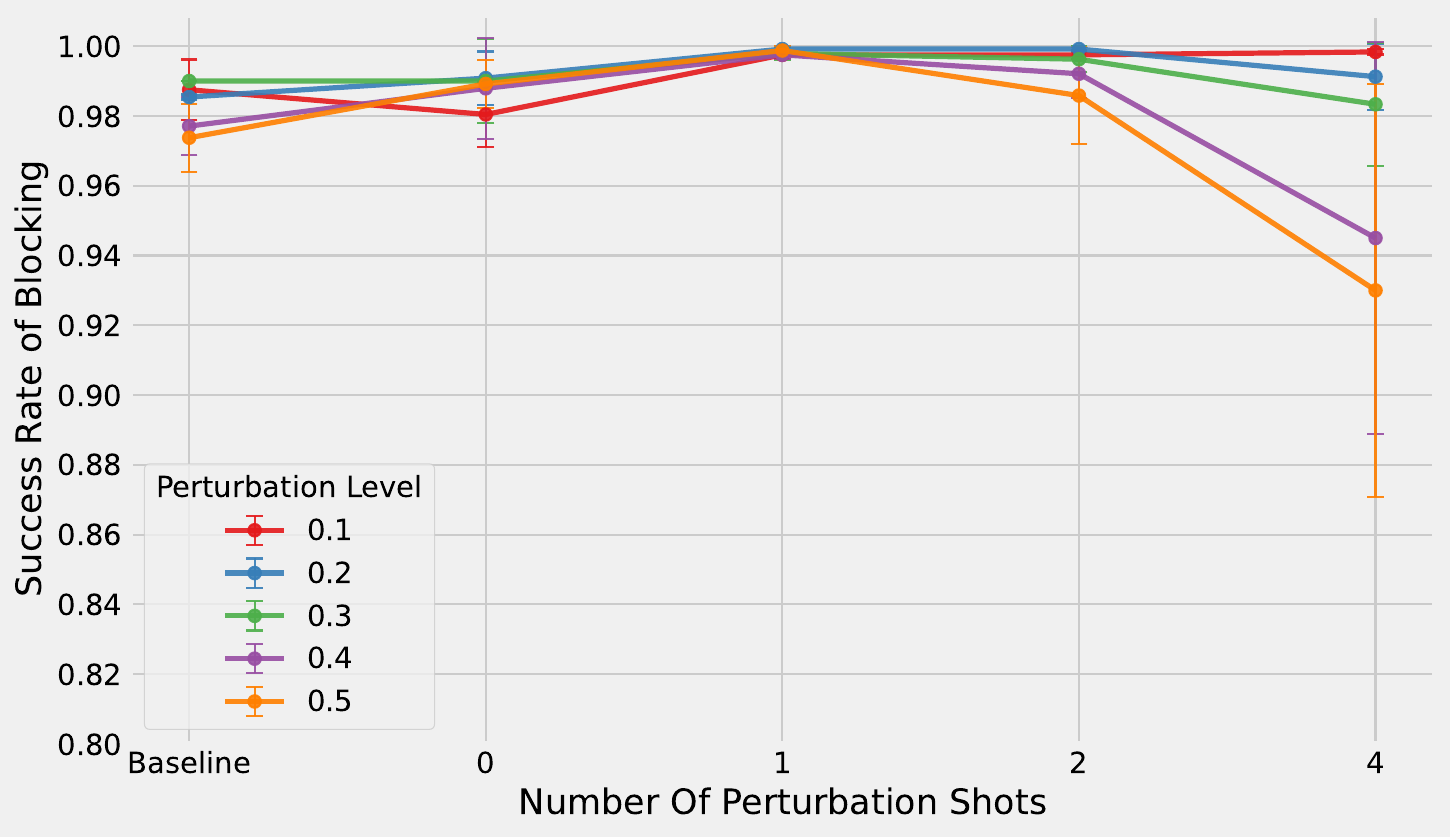}
        \caption{Spam}
        \label{fig:per_block_var_uci}
    \end{subfigure}
    \begin{subfigure}{0.33\textwidth}
        \centering
        \includegraphics[width=\textwidth]{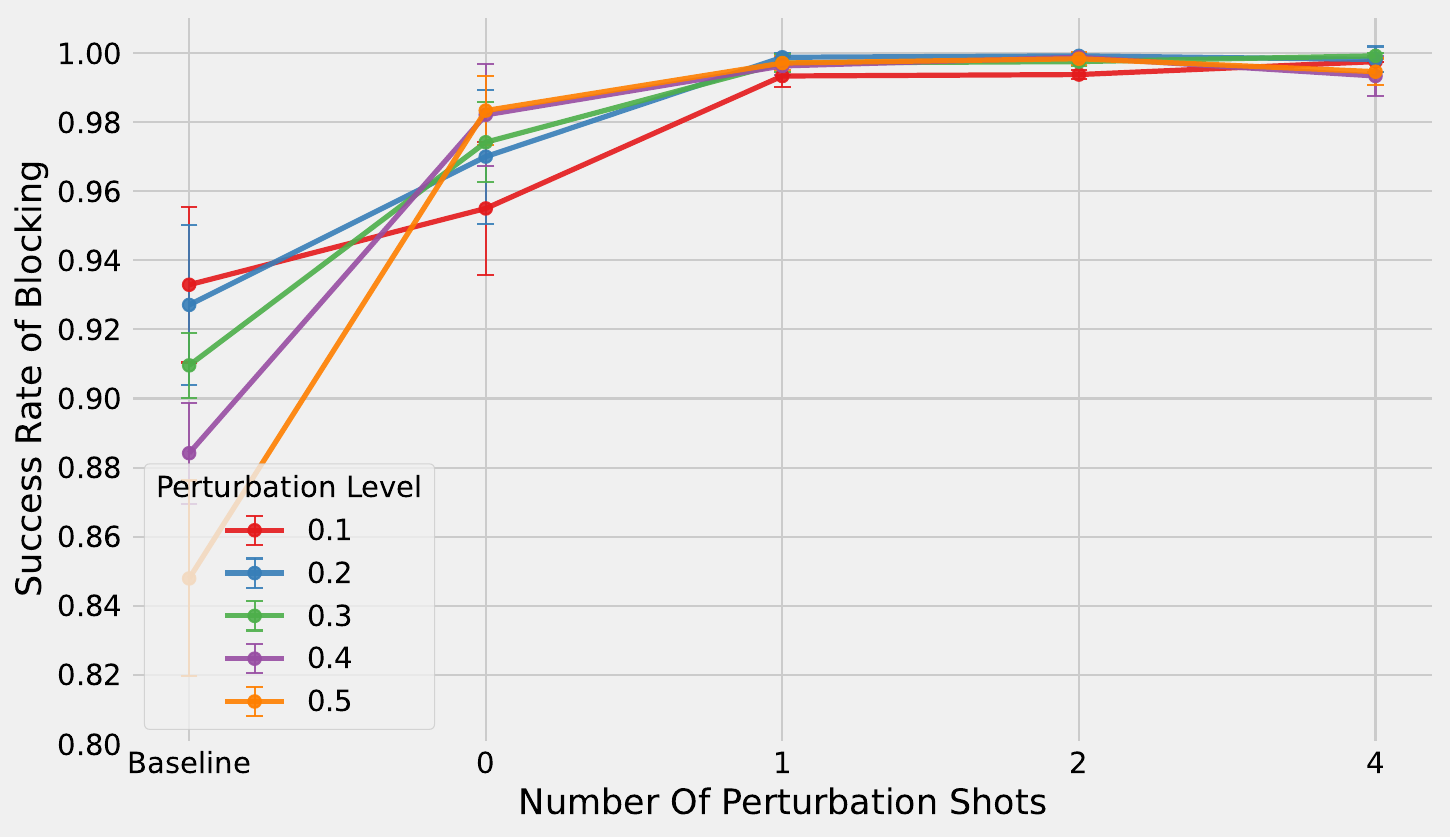}
        \caption{Negative}
        \label{fig:per_block_var_sst2}
    \end{subfigure}
    \caption{The success rate of blocking variants of specific harmful texts. \textbf{Baseline}: The original harmful text is \textit{not} in the prompt. \textbf{0-perturbation}: The original text is included, but no perturbed variants are.}
    \label{fig:per_block_var}
\end{figure}

For \textbf{toxic} texts (Figure~\ref{fig:per_block_var_textdetox}), the baseline achieves good performance. For example, when the perturbation level = 0.1, the success rate of blocking is about 0.97. However, as the perturbation level increases, the success rate begins to decrease, and when perturbation = 0.5, it drops to 0.93. But once a perturbed variant is added to the prompt, the success rate quickly rises above $0.99$ across all perturbation levels, and additional variants (2–4) do not bring further gains. This indicates that toxic texts are highly amenable to instance-level personalization, requiring minimal supervision.

For \textbf{negative} texts (Figure~\ref{fig:per_block_var_sst2}), the baseline struggles more severely with highly perturbed variants, with success rates as low as $0.85$ at perturbation level 0.5. However, similar to the toxic case, including just one perturbed example dramatically improves performance across all perturbation levels, pushing success rates above $0.99$. Interestingly, when perturbed examples are added, higher perturbation levels (0.4–0.5) sometimes become \emph{easier} to block than lower ones. A possible explanation is that heavy perturbations disrupt the original semantic structure, and leaving behind distinctive lexical or surface-level cues. These cues may actually make the perturbed texts more distinguishable from benign content, thereby allowing the model to recognize and block them more reliably.

For \textbf{spam} texts (Figure~\ref{fig:per_block_var_uci}), the baseline performance is better than toxic and negative and achieves about a blocking success rate of 0.98. The blocking rate can be further increased when including either the original test sample or one perturbation. However, there is an abnormal phenomenon that the addition of two or more perturbed variants can actually degrade performance, especially at higher perturbation levels. For instance, with 4 perturbation shots at level 0.5, the success rate drops below $0.90$. We hypothesize this is because, different perturbed spam samples often diverge significantly from each other in both semantics and lexical form, especially at higher perturbation levels, introducing inconsistent patterns that confuse the model. As a result, instead of reinforcing stable spam features, excessive perturbation examples may amplify noise and even mislead ICL.

\subject{Summary.}  
Overall, these results demonstrate that ICL can be effectively personalized to block variations of harmful texts with minimal effort. As little as one perturbed example enables near-perfect blocking for harmful texts across all perturbation levels and more perturbed examples do not mean better results. 

\subsection{Takeaways for Personalization with ICL}

Across the three personalization scenarios, we derive the following key insights:

\begin{itemize}
    \item \textbf{Effortless Expansion:} ICL can seamlessly incorporate \textbf{new harmful categories} with minimal supervision—often a single example or definition suffices. This expansion occurs without catastrophic forgetting, as performance on pre-existing categories remains intact.
    \item \textbf{Flexible Re-calibration:} Users can \textbf{unblock or re-define} categories they find acceptable. While a tension exists between general knowledge and user override, a handful of consistent examples or a clear definition enables reliable personalization, effectively implementing \textbf{contextual unlearning}.
    \item \textbf{Granular Control:} Personalization operates effectively at the \textbf{instance level}. Providing a single harmful example (and optionally, its perturbed variants) allows the model to construct a robust "concept" that generalizes to future, similar instances, enabling precise, user-driven content filtering.
\end{itemize}

\noindent Taken together, these results highlight that ICL is a powerful and flexible framework for personalization. With only minimal supervision—ranging from a handful of examples to natural-language definitions—ICL can adapt to new user requirements while maintaining strong performance on established tasks. This positions ICL as a practical solution for real-world harmful content detection where personalization and customization are essential.

\section{Evaluation on Wild Data}
\label{sec:wild_data}

Thus far, our evaluation of ICL has focused on multi-task harmful content detection and personalization using well-established public benchmarks. While such datasets provide a controlled environment for systematic comparison, they may not fully capture the complexity and diversity of real-world scenarios. 

To bridge this gap, we further examine ICL performance on \textit{wild data}—unseen, uncurated examples collected from open platforms—which often feature noisier distributions, domain shifts, and ambiguous labeling. This section investigates the generalizability and robustness of ICL under such unconstrained, naturally occurring conditions.

We consider three task formulations: multi-task binary, multi-task multi-class, and multi-task multi-label classification. The first two mirror the settings introduced earlier, while the multi-task multi-label formulation is newly incorporated here to better reflect the compositional nature of real-world texts, where content frequently spans multiple categories (e.g., spam messages expressed in a toxic manner).


\subsection{Data Source}
We construct our wild dataset from Mastodon, a decentralized social media platform characterized by diverse communities and open discourse. Between December 2024 and February 2025, we collected 8,998,738 posts, of which 3,948,831 were unique English-language entries.

Since benign content dominates real-world platforms, we employ a two-stage filtering strategy to obtain a more balanced distribution. First, we randomly sample 15{,}000 English-language posts. Next, we apply the Llama model with 48-shot random demonstrations—identified as strong in our benchmark experiments—for a preliminary harmfulness prediction. This yields 12,616 posts predicted as benign and 2,384 as harmful. From each group, we then randomly select 1,500 posts, resulting in a balanced subset of 3,000 instances for manual annotation. 

\subsection{Annotation Protocol}
To accommodate the multiple granularities and inherent overlaps of harmfulness in real-world data, we designed a hierarchical annotation protocol where each post received three types of labels:
\begin{enumerate}
    \item \textbf{Binary:} A single label of either \textit{benign} or \textit{harmful}. This supports the evaluation of multi-task binary classification.
    \item \textbf{Multi-class:} A single, mutually exclusive label from the set \{\textit{benign}, \textit{toxic}, \textit{spam}, \textit{negative}\}. Harmful posts are assigned one of the three harmful sub-categories. This supports multi-task multi-class classification.
    \item \textbf{Multi-label:} One or more labels from the set \{\textit{benign}, \textit{toxic}, \textit{spam}, \textit{negative}\}. This formulation explicitly captures overlaps (e.g., \textit{toxic} + \textit{negative}), borderline cases (e.g., \textit{benign} + \textit{spam}), and composite scenarios, providing a more faithful representation of real-world content for multi-task multi-label classification.
\end{enumerate}

This hierarchical scheme enables evaluation under binary, multi-class, and multi-label settings, providing a rich testbed for analyzing ICL robustness in complex, real-world contexts.

\subsection{Label Distribution}
Table~\ref{tab:label_distribution_Mastodon} summarizes the label composition. Under the multi-class setting, 1,202 of the 3,000 samples are harmful: 755 \texttt{negative} (62.8\%), 259 \texttt{toxic} (21.6\%), and 188 \texttt{spam} (15.6\%). This distribution aligns with expectations: Mastodon users frequently express personal emotions or frustrations, making \texttt{negative} content dominant among harmful posts.

In the multi-label setting, 980 texts (32.7\%) carry two or more labels. Common overlaps include \texttt{negative + toxic} (339 texts), \texttt{benign + negative} (249 texts), and \texttt{benign + spam} (184 texts). We also observe 32 triple-label cases (e.g., \texttt{benign + negative + toxic}). Notably, some posts simultaneously bear \texttt{benign} and \texttt{toxic}/\texttt{spam} labels. These overlaps may seem counterintuitive at first glance, but closer inspection reveals their plausibility. Below we provide illustrative examples:  

\begin{itemize}
    \item \textbf{Benign + Spam.} These texts lack overtly malicious intent, but their link-heavy structure or excessive use of hashtags makes them spam-like in appearance.  
    \begin{quote}\small
    \texttt{case1: obligatory provided link monstrosity https://www.theguardian.com/[...]} \\
    \texttt{case2: Depression. Written by Satan \& Lee Wilson. \#bookblog \#readallthebooks \#MSWL \#WhatToRead \#Writing \#LeeWilson \#BookMarketing \#WritingCommunity \#amquerying \#Reading Block Filter - \#UGhhYmFyZw https://satan-speaks.com/[...]}
    \end{quote}

    \item \textbf{Benign + Toxic.} These posts are primarily expressive or informative, yet contain profanity, offensive language, or sarcastic tone that warrant a toxic label.  
    \begin{quote}\small
    \texttt{case1: FUCK this show is so good} \\
    \texttt{case2: LIVEBLOG: Israeli Army Kills Scores in Gaza, Assassinates Police Chief [...] “At least 11 displaced Palestinians were killed, and others injured, after Israeli occupation forces targeted tents in the Al-Mawasi area, west of Khan Yunis [...]” \#Press \#Gaza \#Palestine \#Resistance \#Israel \#Genocide \#Terrorism \#WarCrimes \#Hamas \#Zionism \#Barbarism \#BloodLust} \\
    \texttt{case3: Nothing but clap for MY jeb!sident}
    \end{quote}
\end{itemize}

These examples demonstrate that what might initially appear contradictory under rigid single-label schemes often reflects the fuzzy and compositional nature of real-world discourse. Consequently, multi-label formulations are more faithful in capturing the nuanced semantics of wild data. 

\begin{table}
    \centering
    \caption{Label distribution of Mastodon Wild Data.}
    \label{tab:label_distribution_Mastodon}

    \begin{subtable}[t]{0.20\textwidth}
        \centering
        \caption{Multi-class Label}
        \label{tab:label_distribution_Mastodon_multi-class}
        \begin{tabular}{lc}
            \toprule
            Label & Count \\
            \midrule
            Benign  & 1798 \\
            Negative & 755 \\
            Toxic & 259 \\
            Spam & 188 \\
            \bottomrule
        \end{tabular}
    \end{subtable}
    \hfill
    \begin{subtable}[t]{0.74\textwidth}
        \centering
        \caption{Multi-label Label}
        \label{tab:label_distribution_Mastodon_multi-label}
        \begin{tabular}{lclclc}
            \toprule
            Labels   & Count & Labels & Count & Labels & Count \\
            \midrule
            Benign & 1437 & Benign, Negative & 249 & Benign, Negative, Spam & 11  \\
            Negative & 517 & Benign, Spam & 184 & Benign, Negative, Toxic & 13 \\
            Spam & 60 & Benign, Toxic & 8 & Benign, Spam, Toxic & 5 \\
            Toxic & 6 & Negative, Spam & 10 & Negative, Spam, Toxic & 3 \\
            - & - & Negative, Toxic & 339 & - & - \\
            - & - & Spam, Toxic & 98 & - & - \\
            \midrule
            Sum & 2020 & - & 948 & - & 32 \\
            \bottomrule
        \end{tabular}
    \end{subtable}
\end{table}

\subsection{Multi-Task Binary on Wild Data}

\subject{Experimental Setup} Most of the experimental settings remain the same as those previously described, with only minor modifications. In this section, we detail the parts that differ.

\begin{itemize}
    \item \textbf{Models, Selection Strategies, and Number of Shots}  
    We adopt the best-performing model-strategy combination identified on the public benchmarks: \texttt{Llama-Random}. Regarding the number of shots, although using 192 shots yields the highest performance (F1-score: 0.918), the 48-shot setting already achieves competitive results (F1-score: 0.887). To balance effectiveness and efficiency, we fix the number of shots at 48.

    \item \textbf{Datasets}  
    We use the 3,000 annotated examples introduced earlier as the test set $D_{test}$. To evaluate the generalization capability of ICL, we first construct demonstrations ($D_k$) from the training sets of public datasets, i.e., $D_k^r \subseteq D_{train}^{\text{TextDetox}} \cup D_{train}^{\text{UCI}} \cup D_{train}^{\text{SST2}}$. However, since these public datasets differ from our wild data in both source and style, we also explore a more realistic scenario: we annotate an additional 1,000 wild data samples following the same annotation protocol, and use them as $D_{train}$ for constructing in-domain demonstrations. This allows us to assess the performance of ICL when a certain amount of real-world annotated data is available.

    \item \textbf{Prompt Template} Initially, we used the same prompt template as in the previous experiments. However, inspired by later empirical findings (to be discussed in detail later), we experimented with a modified prompt format: when presenting demonstration examples in the prompt, we not only include the corresponding labels but also provide a reason for why the text belongs to that label. The updated prompt template is illustrated in Figure~\ref{fig:prompt_template_reason}. The \textit{reason} for a demonstration is generated by using DeepSeek-V3 under the assumption that the label is known. Let $\mathcal{R}$ denote the reason space. Under this setting, $D_k^r = \{ (x_1, r_1, y_1), (x_2, r_2, y_2), \dots, (x_k, r_k, y_k) \}$.

\end{itemize}

\begin{figure}[ht]
    \centering
        \begin{tcolorbox}[colback=white,colframe=black,width=0.9\textwidth]
        \texttt{<Description of the specific task> \\
        == \\
        Query:<sentence 1> \\
        Answer:\{"reason": <reason>, "label": <label>\} \\
        == \\
        Query:<sentence 2> \\
        Answer:\{"reason": <reason>, "label": <label>\} \\
        == \\
        ... \\
        Query:<predicted text> \\
        Answer:
        }
        \end{tcolorbox}
    \caption{Prompt Template with Reason for ICL.}
    \label{fig:prompt_template_reason}
\end{figure}

\subject{Results} As shown in Table~\ref{tab:performance_mul_binary_wild}, when using demonstrations sampled from public datasets, ICL achieves an accuracy of 0.829 and an F1-score of 0.795 on the wild test set. Although the recall is relatively high at 0.883, the false positive rate (FPR) is also high (0.203), resulting in a relatively low precision of 0.724. Compared with the performance on public test sets (F1-score = 0.887), there is a clear performance drop. We hypothesize that this is due to the domain gap between the public datasets and the wild data—differences in language style, expression patterns, and content distribution likely cause a mismatch, making it difficult for the model to generalize when demonstrations come from a different source.

To test this hypothesis, we replaced all demos with samples drawn from the wild dataset and repeated the experiment. This setup leverages one of the strengths of ICL: the ability to rapidly adapt to new domains given only a few labeled examples, without further training. Surprisingly, however, the performance did not improve—in fact, it deteriorated. The FPR increased from 0.203 to 0.224, and the recall dropped from 0.883 to 0.836. This counterintuitive result suggests that although the demos now match the domain of the test set, their effectiveness is compromised. A possible explanation is that the wild data is inherently noisy and exhibits uneven quality, making it difficult for the model to extract reliable decision patterns from the raw labels alone.


The failure of in-domain wild demos suggests that raw labels alone are insufficient for the model to disambiguate noisy, real-world examples. We hypothesize that providing \textit{interpretive signals} could guide the model towards more robust decision patterns. To test this, we augment the demonstration format to include a natural-language rationale, transforming each demo from $(x_i, y_i)$ to $(x_i, r_i, y_i)$, where $r_i$ is a reason explaining why $x_i$ belongs to class $y_i$. These rationales are generated automatically by DeepSeek-V3, conditioned on the ground-truth label.

Experimental results confirm this intuition. After adding the rationale, although recall dropped slightly to 0.800, the FPR significantly decreased to 0.019. As a result, the precision rose dramatically to 0.966, and accuracy increased to 0.909. These results indicate that providing interpretive signals during in-context learning can greatly reduce overprediction and improve overall robustness, especially in noisy, real-world scenarios.

\begin{table}x
    \centering
    \caption{Performance of Multi-Task Binary ICL on Wild Data with Different Demonstration Settings}
    \label{tab:performance_mul_binary_wild}
    \begin{tabular}{lccccc}
        \toprule
        Demos Source (Prompt Template) & Precision & Recall & FPR & F1-Score & Accuracy \\
        \midrule
        Public Datasets (No Reason) & 0.724 & 0.883 & 0.203 & 0.795 & 0.829 \\
        Wild Data (No Reason)       & 0.714 & 0.836 & 0.224 & 0.770 & 0.800 \\
        Wild Data (With Reason)     & 0.966 & 0.800 & 0.019 & 0.875 & 0.909 \\
        \bottomrule
    \end{tabular}
\end{table}

\subsection{Multi-Task Multi-Class}

After evaluating ICL under the multi-task binary setting, we now shift our focus to the more fine-grained \textit{multi-task multi-class} setting. 

\subject{Experimental Setup} We adopt the same three configurations as in the previous section to investigate the impact of demo source and prompt format: (1) using public datasets as demos without reasoning, (2) using wild data as demos without reasoning, and (3) using wild data with both labels and corresponding reasoning. Apart from the expanded label space, the experimental setup remains largely consistent with the multi-task binary setting.

\begin{figure}
    \centering
    \begin{subfigure}{0.32\textwidth}
        \centering
        \includegraphics[width=\textwidth]{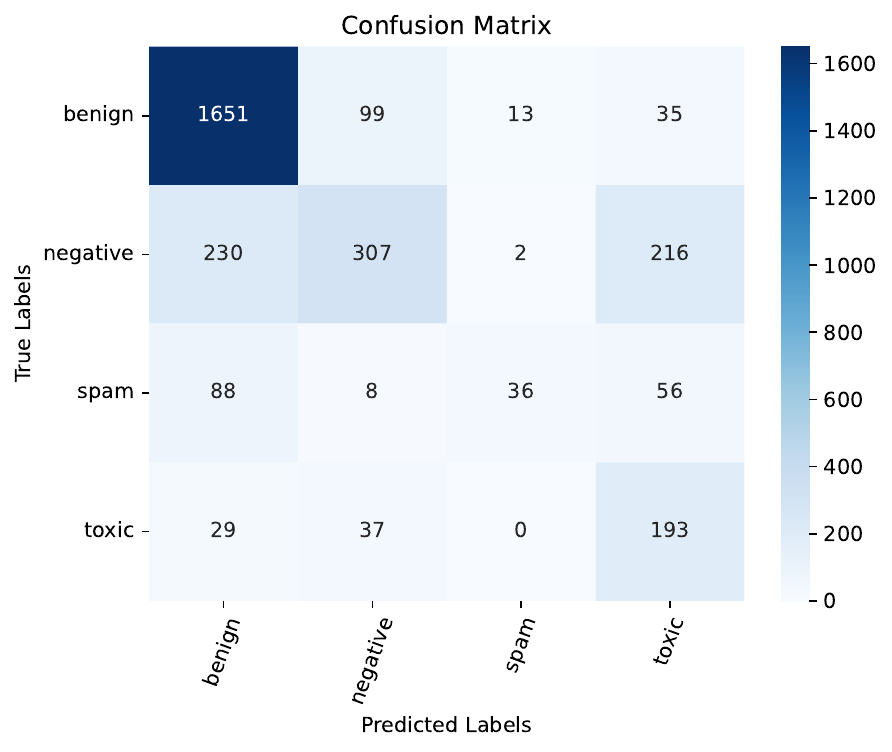}
        \caption{Public Datasets(No Reason)}
    \end{subfigure}
    \begin{subfigure}{0.32\textwidth}
        \centering
        \includegraphics[width=\textwidth]{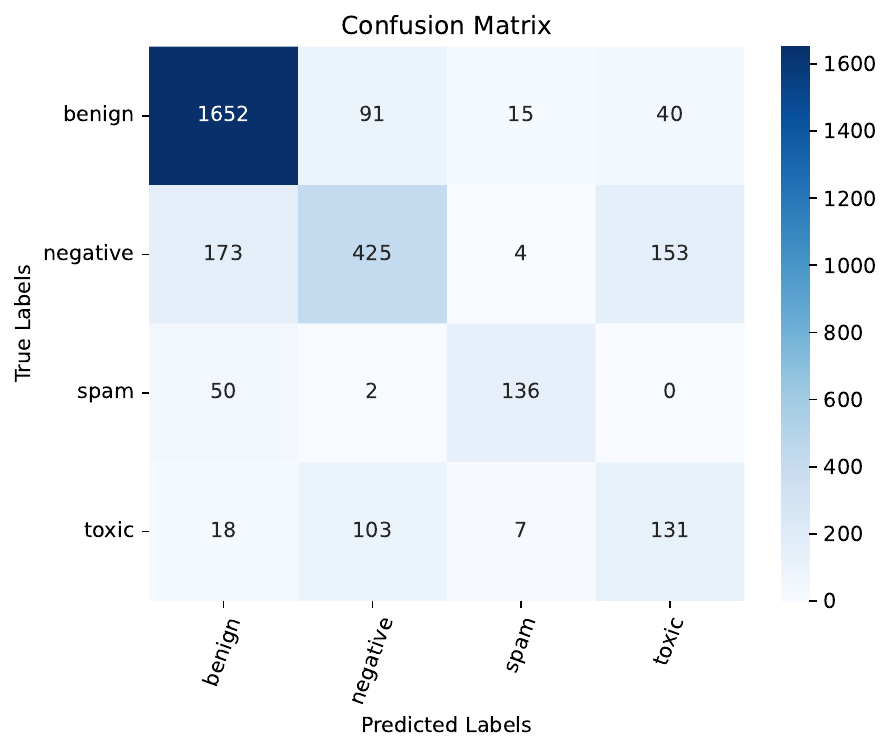}
        \caption{Wild Data(No Reason)}
    \end{subfigure}
    \begin{subfigure}{0.32\textwidth}
        \centering
        \includegraphics[width=\textwidth]{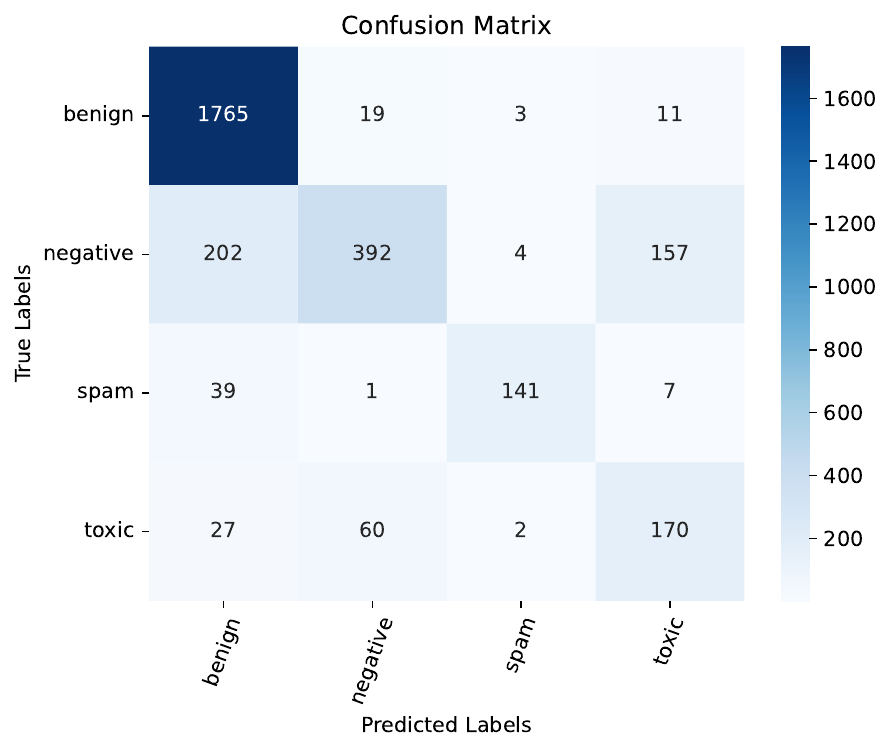}
        \caption{Wild Data(Reason)}
    \end{subfigure}
    \caption{Confusion Matrix of Multi-Task Multi-Class ICL on Wild Data with Different Demonstration Settings.}
    \label{fig:wild_multi-class_confusion}
\end{figure}

\subject{Results} Performance improves consistently across configurations (Table~\ref{tab:f1-score_mul_class_wild}). Weighted F1 increases from 0.712 (public demos) to 0.814 (wild demos + reason). The largest gain occurs for \texttt{spam} (F1: 0.301 $\to$ 0.834). Despite these overall improvements, the performance on the \textit{negative} and \textit{toxic} categories remains suboptimal. As visualized in the confusion matrix (Figure~\ref{fig:wild_multi-class_confusion}), a considerable number of \textit{negative} samples are misclassified as \textit{benign} or \textit{toxic}, while many \textit{toxic} samples are predicted as \textit{negative}. Additionally, a significant portion of \textit{spam} instances are incorrectly labeled as \textit{benign}.

We hypothesize that these confusions are partly due to the multi-label nature of some examples in our dataset. Based on our annotation experience, certain texts may simultaneously exhibit characteristics of multiple categories (e.g., a toxic comment that also expresses negative emotion), which makes the single-label classification setting inherently challenging for these classes.

\begin{table}[h]
    \centering
    \caption{F1-Score of Multi-Task Multi-Class ICL on Wild Data with Different Demonstration Settings}
    \label{tab:f1-score_mul_class_wild}
    \begin{tabular}{lccccc}
        \toprule
        Demos Source (Prompt Template) & Benign & Negative & Spam & Toxic & Weighted Avg \\
        \midrule
        Public Datasets (No Reason) & 0.870 & 0.509 & 0.301 & 0.509 & 0.712 \\
        Wild Data (No Reason)       & 0.895 & 0.618 & 0.777 & 0.449 & 0.779 \\
        Wild Data (With Reason)     & 0.921 & 0.639 & 0.834 & 0.563 & 0.814 \\
        \bottomrule
    \end{tabular}
\end{table}

\subsection{Multi-Task Multi-Label}

While the previous section focuses on multi-task multi-class classification, many real-world harmful content detection scenarios involve overlapping or co-occurring categories---that is, a single piece of content may simultaneously exhibit characteristics of multiple classes, such as being both \textit{toxic} and \textit{negative}. To better accommodate such complexity, we now consider the more realistic and challenging \textbf{multi-task multi-label} setting, where the model must predict one or more labels per input instance.

This formulation is particularly relevant for open-domain harmful content detection, where rigidly assigning a single label may overlook important semantic nuances. For example, a message attacking someone in a hostile and angry tone may reasonably be classified as both \textit{negative} and \textit{toxic}. As a result, this setting imposes higher demand on the model's ability to capture multi-faceted cues within each input and distinguish subtle overlaps between label semantics.

In this section, we evaluate how in-context learning (ICL) performs under this multi-label setup, and analyze the ICL’s strengths and limitations when handling overlapping categories.

\subject{Experimental Setup} Since previous experiments have demonstrated that using demonstrations from wild data along with explicit reasoning consistently yields better performance in both multi-task binary and multi-task multi-class settings, we adopt this configuration exclusively in our evaluation of the multi-task multi-label setup.

Given the multi-label nature of this task, the prompt template is slightly modified to instruct the model to output multiple labels per instance. An illustration of the updated prompt format is shown in Figure~\ref{fig:prompt_template_label_reason}. All other experimental settings remain consistent with multi-task binary apart from the expanded label space.

\begin{figure}[ht]
    \centering
        \begin{tcolorbox}[colback=white,colframe=black,width=\textwidth]
        \texttt{<Description of the specific task> \\
        == \\
        Query:<sentence 1> \\
        Answer:\{"reason": <reason>, "is\_benign": True/False,  "is\_negative": True/False, "is\_toxic": True/False, "is\_spam": True/False,\} \\
        == \\
        Query:<sentence 2> \\
        Answer:\{"reason": <reason>, "is\_benign": True/False,  "is\_negative": True/False, "is\_toxic": True/False, "is\_spam": True/False,\} \\
        == \\
        ... \\
        Query:<predicted text> \\
        Answer:
        }
        \end{tcolorbox}
    \caption{ICL Prompt Template for Multi-Task Multi-Label Classification with Reason.}
    \label{fig:prompt_template_label_reason}
\end{figure}


\subject{Results} The Multi-Task Multi-Label ICL model achieves promising results on wild data, with a \textbf{Subset Accuracy} of 0.707, indicating that 70.7\% of samples have exactly matched label sets between prediction and ground truth. This is a relatively high score in the multi-label setting, where strict exact-match criteria are often hard to satisfy. To evaluate per-label prediction quality, we report two common metrics: \textbf{Hamming Loss}\footnote{https://scikit-learn.org/stable/modules/generated/sklearn.metrics.hamming\_loss.html} and \textbf{Jaccard Score}\footnote{https://scikit-learn.org/stable/modules/generated/sklearn.metrics.jaccard\_score.html}. Hamming Loss is defined as the fraction of incorrectly predicted labels over the total number of labels, and the lower the better. Our model achieves a Hamming Loss of 0.090, which means only 9.0\% of all possible label predictions are incorrect, demonstrating low per-label error. Jaccard Score, also known as Intersection over Union (IoU), is calculated as the size of the intersection divided by the size of the union between predicted and true label sets, and reflects the degree of partial correctness. Our model achieves a Jaccard Score of 0.831, suggesting strong alignment between predicted and actual label combinations.

Detailed per-class results are summarized in Table~\ref{tab:performance_mul_label_wild}. The model performs particularly well on \textit{benign} and \textit{negative} categories. \textit{Benign} has very high precision (0.917), while \textit{negative} achieves an impressive recall of 0.916, showing that most truly negative samples are successfully detected. Performance on \textit{spam} is slightly weaker, with a recall of 0.836 and F1-score of 0.795; we hypothesize this may be due to the presence of link-heavy or hashtag-laden content on platforms like Mastodon, which could inflate false positives. The \textit{toxic} class shows a good balance between precision (0.845) and recall (0.829), leading to an F1-score of 0.837. Overall, the model attains a \textbf{weighted average F1-score} of 0.867, indicating reliable and balanced performance across categories without major bias. These results demonstrate that ICL is effective in the multi-task multi-label setting, performing well both at the instance level and the label level.

\begin{table}
    \centering
    \caption{Performance of Multi-Task Multi-Label ICL on Wild Data}
    \label{tab:performance_mul_label_wild}
    \begin{tabular}{lcccc}
        \toprule
        Label & Count & Precision & Recall & F1-Score \\
        \midrule
        Benign   & 1907 & 0.917 & 0.874 & 0.895 \\
        Negative & 1202 & 0.810 & 0.916 & 0.859 \\
        Spam     &  371 & 0.758 & 0.836 & 0.795 \\
        Toxic    &  532 & 0.845 & 0.829 & 0.837 \\
        \midrule
        Weighted Avg & 4012 & 0.861 & 0.877 & 0.867 \\
        \bottomrule
    \end{tabular}
\end{table}

\subsection{Summary}
Our evaluation on Mastodon wild data highlights three key insights.  
First, models that achieve strong performance on clean public benchmarks exhibit clear degradation when directly applied to wild data, underscoring the challenge of domain shifts and noisy distributions. This confirms that results obtained solely on curated datasets may overestimate real-world robustness, and highlights the necessity of complementary wild data evaluation.  
Second, augmenting demonstrations with explicit rationales proves highly effective: this design substantially reduces false positives and improves robustness, showing that interpretive signals can compensate for label noise in real-world contexts.  
Third, both annotation observations and experimental results confirm that single-label formulations are often insufficient in open-domain harmful content detection. Many posts simultaneously display characteristics of multiple categories, and the multi-label setting provides a more faithful and informative modeling paradigm.  

Taken together, these findings establish wild data evaluation as an essential step for validating the practical utility of ICL. Compared with public datasets, wild data not only reveals previously hidden weaknesses (e.g., domain sensitivity, high FPR), but also motivates new design choices (reason-enriched prompts, multi-label-aware formulations) that significantly improve performance. These results demonstrate that robust deployment of ICL for harmful content detection requires going beyond benchmarks to embrace real-world complexity.

\section{Discussion}
\label{sec:discussion}

\subject{Cross-Task Generalization of ICL.} Our results show that in-context learning (ICL) exhibits strong cross-task generalization for harmful content detection. Across binary, multi-class, and multi-label formulations, ICL consistently achieves competitive or superior performance compared to supervised baselines, despite requiring only a small number of demonstrations. This highlights ICL’s ability to flexibly adapt to diverse task formulations without retraining, a critical advantage in real-world moderation where task boundaries often shift dynamically. Importantly, the robustness of ICL across these heterogeneous setups distinguishes our work from prior studies that evaluate harmful content detection only in single-task settings.

\subject{Personalization.}
Beyond cross-task robustness, our study highlights the promise of ICL for personalization. By designing three personalized scenarios, we show that models can rapidly adapt to blocking or unblocking specific categories of harmful behaviors with minimal supervision. Moreover, ICL also allows for targeted defenses against adversarial modifications, ensuring that even highly altered harmful content is effectively captured. Together, these findings demonstrate that personalization in ICL is not limited to dataset-level adjustments but can extend to user- or community-specific harmfulness definitions, offering a lightweight alternative to costly retraining pipelines.

\subject{Nuances of Wild Data Evaluation.}
The wild data experiments underscore the importance of moving beyond curated benchmarks. Mastodon posts often display overlapping or borderline labels (e.g., \texttt{benign+spam}, \texttt{negative+toxic}), which are poorly represented in traditional datasets. Our multi-label evaluation demonstrates that ICL can capture these compositional cases, but also exposes limitations in single-label formulations that mask the inherent fuzziness of real-world content. These findings reinforce the necessity of adopting more nuanced evaluation protocols and richer datasets that account for overlap, ambiguity.

\subject{Limitations and Future Directions.}  
Despite these contributions, several limitations remain. First, our study currently covers only three representative harmful categories (toxicity, spam, negative sentiment). Future research should incorporate a broader taxonomy, including misinformation, violent content, etc., to achieve more comprehensive coverage. Second, while our multi-class and multi-label settings already capture a range of realistic scenarios, the taxonomy itself remains relatively coarse. Future work could explore finer-grained subdivisions within each category. For example, \textit{toxicity} can be broken down into subtypes such as threats, insults, or personal attacks; \textit{spam} can be subdivided into promotional spam, phishing attempts, or link farming; and \textit{negative sentiment} can be differentiated into emotions such as sadness, anger, or frustration. Such granularity would allow for more precise modeling of harmful behaviors as they manifest in diverse forms. Third, our evaluation of reason-augmented prompts focuses solely on classification outcomes without systematically assessing the quality, faithfulness, or potential biases of the generated explanations. Fourth, although Mastodon provides a valuable wild testbed, it cannot capture the full diversity of harmful content across platforms, domains, and user demographics. Beyond text, multimodal harmful content (e.g., hateful memes, doctored videos, multimodal spam) poses additional challenges. Finally, our experiments are restricted to English, leaving open questions about the effectiveness of ICL in multilingual or code-switching scenarios.

\subject{Code and Data Availability.} 
To foster reproducibility and future research, we publicly release our codebase and the annotated Mastodon dataset. The codebase includes implementations for all ICL configurations, retrieval strategies, and evaluation scripts. The dataset comprises the 3,000 annotated posts with binary, multi-class, and multi-label labels, providing a valuable benchmark for robust harmful content detection. Access links are available in the Abstract for reference.
\section{Conclusion}
\label{sec:conclusion}
We have presented a comprehensive study of in-context learning for harmful content detection, systematically evaluating its capabilities across cross-task generalization, user personalization, and robustness in the wild. Our findings demonstrate that ICL, when configured with appropriate models, retrieval strategies, and prompt designs, achieves strong performance competitive with fine-tuned models across binary, multi-class, and multi-label tasks, all without task-specific training.

Crucially, we showed that ICL is not only powerful but also highly adaptable. It enables \textbf{lightweight personalization}, allowing users to block new categories, unblock existing ones, and defend against adversarial variations with minimal examples. Furthermore, our evaluation on a new Mastodon wild dataset revealed the limitations of benchmark-only evaluation and identified \textbf{rationale-augmented prompts} as a key technique for restoring robustness against noisy, real-world data and mitigating false positives.

In summary, this work moves \textbf{beyond one-size-fits-all} moderation by establishing ICL as a practical, privacy-preserving, and user-centric pathway for content safety. By bridging the gap between controlled benchmarks and the complexities of the wild, we hope to inspire future work towards more robust, explainable, and adaptable online safety systems.

\clearpage

\bibliographystyle{IEEEtranS}
\bibliography{bibliography}

\appendix
\renewcommand{\thesection}{Appendix \Alph{section}}
\section{Additional Details on Model Selection}
\label{appendix:model_selection}

As discussed in Section~\ref{subsec:icl-setting}, we adopt three instruction-tuned foundation models available by July 2024: Llama-3.1-8B-Instruct, Mistral-7B-Instruct-v0.3, and Qwen2-7B-Instruct. 
Here we provide additional details on subsequent releases between July 2024 and July 2025, and explain why we don't update models.  

Table~\ref{tab:model_timeline} summarizes the newer models. For the Llama family, multiple versions were released, including lightweight models (1B/3B), large multimodal models (11B/90B), a single 70B text-only model, and the Llama 4 MoE series (109B/400B). These models are either too small to serve as strong baselines or too large to evaluate efficiently in our ICL setup, and thus fall outside our experimental scope.  

\begin{figure}
    \centering
    \begin{subfigure}{0.49\textwidth}
        \centering
        \includegraphics[width=\textwidth]{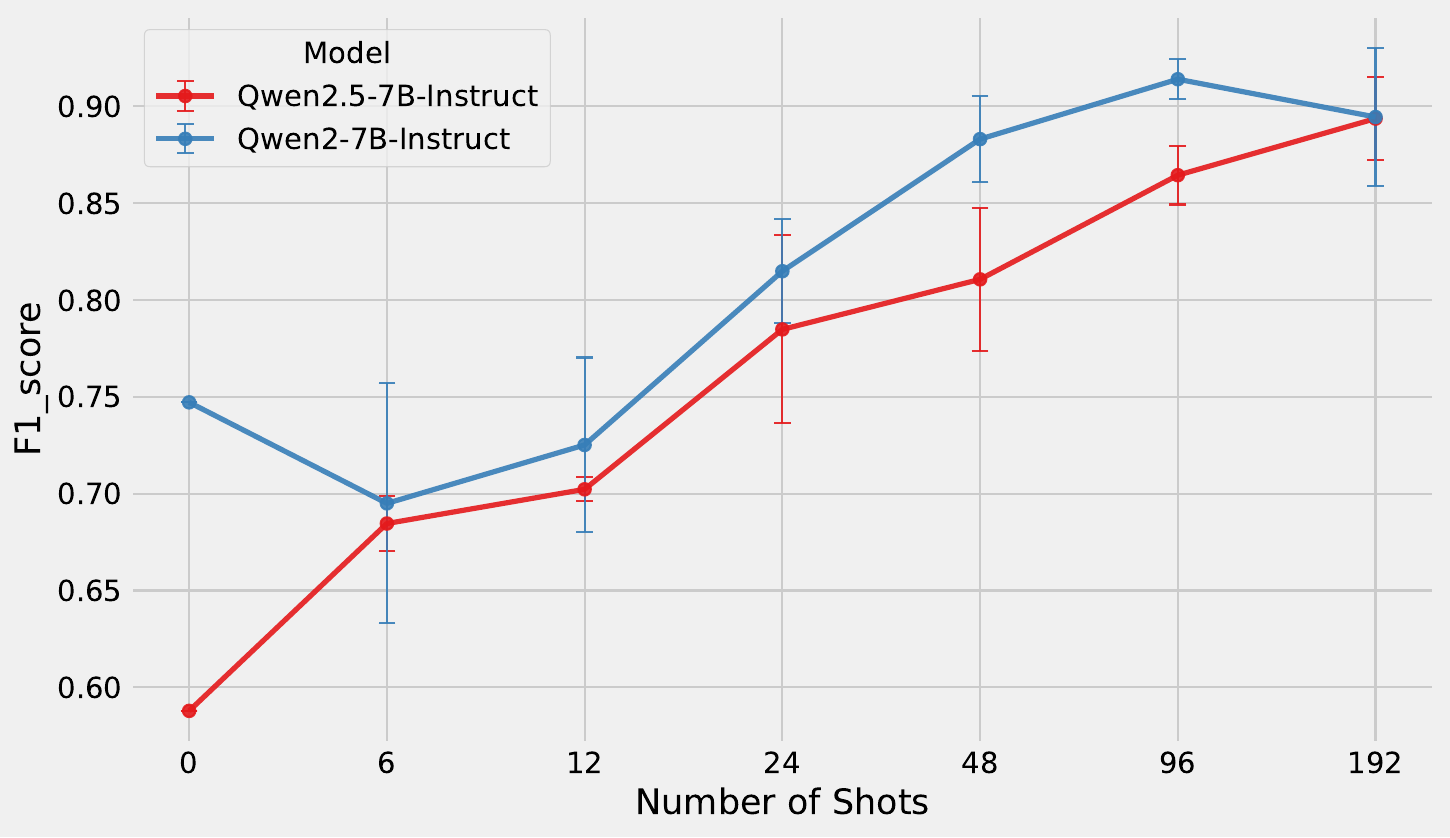}
        \caption{Qwen}
        \label{fig:model_comparison_qwen}
    \end{subfigure}
    \begin{subfigure}{0.49\textwidth}
        \centering
        \includegraphics[width=\textwidth]{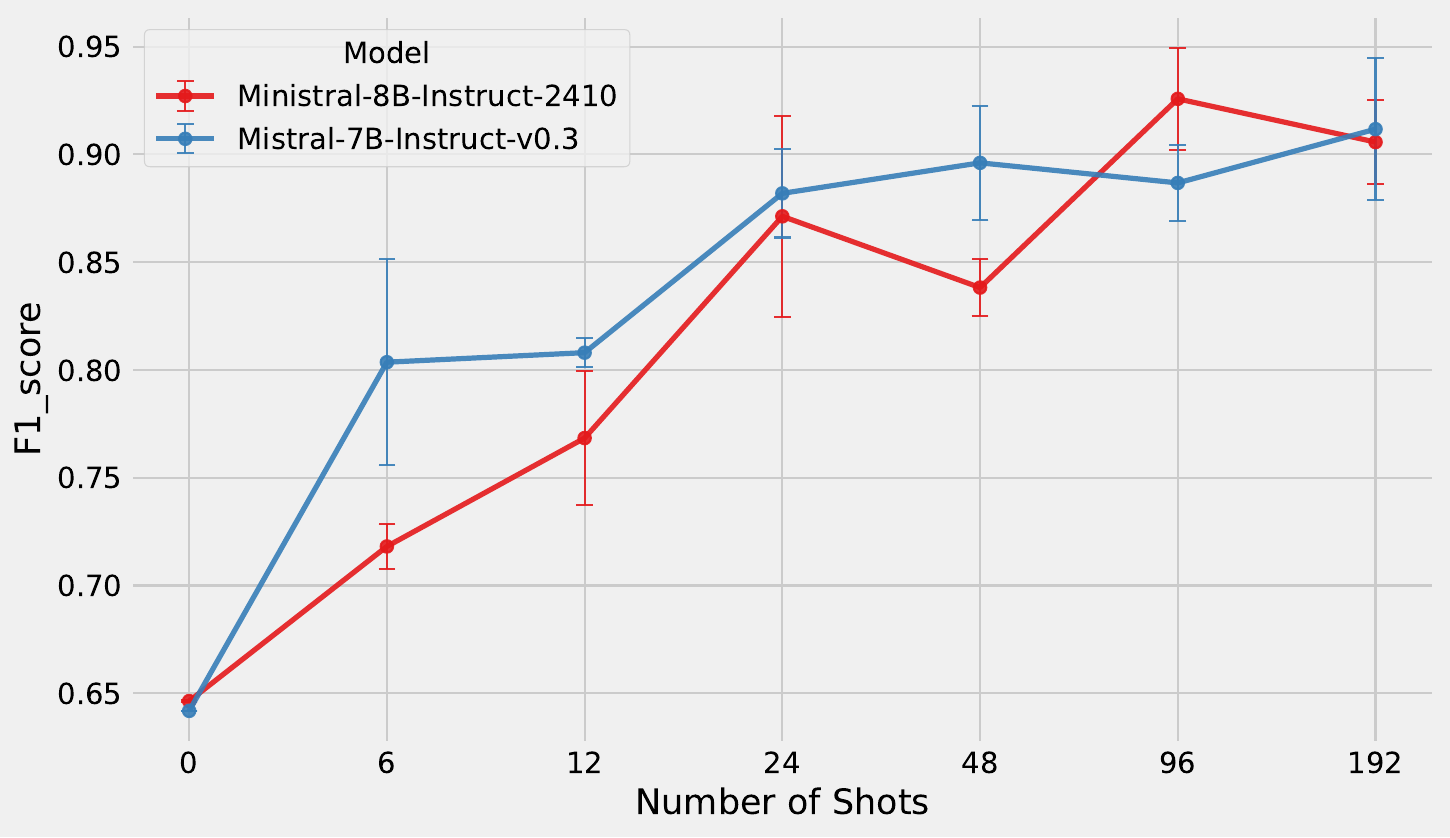}
        \caption{Mistral}        
        \label{fig:model_comparison_mistral}
    \end{subfigure}
    \caption{Performance comparison between adopted baselines and newly released models on the multi-task binary task (same configurations).}
    \label{fig:model_comparison}
\end{figure}

For Qwen, two new generations have been released. Qwen2.5 provides a range of dense models (0.5B–72B) and task-specific variants (Coder and Math). Qwen3 introduces both MoE and dense models, spanning from 0.6B to 235B. Among them, only Qwen2.5-7B-Instruct is comparable in scale and task orientation to our baseline. However, as shown in Figure~\ref{fig:model_comparison_qwen}, Qwen2.5-7B-Instruct performs similarly to, or in some cases worse than, Qwen2-7B-Instruct under our evaluation.  

For Mistral, many new models have appeared. The suitable variant is Ministral-8B-Instruct-2410. Yet, Figure~\ref{fig:model_comparison_mistral} demonstrates that this model does not yield clear improvements over Mistral-7B-Instruct-v0.3 in the multi-task binary setting.  

In summary, although a wide range of newer models have been released between July 2024 and July 2025, they are either mismatched in scale, specialized for unrelated domains (e.g., multimodal, programming, math), or do not provide measurable advantages over our adopted baselines. We therefore retain Llama-3.1-8B-Instruct, Qwen2-7B-Instruct, and Mistral-7B-Instruct-v0.3 as the representative models throughout our study.

\begin{table}[h]
\centering
\small
\begin{tabular}{lll}
\toprule
\textbf{Family} & \textbf{Previously Used Model} & \textbf{New Releases (2024.7--2025.7)} \\
\midrule
Llama 
& Llama-3.1-8B-Instruct 
& Llama 3.2 (1B/3B, 11B/90B multimodal); \\
& & Llama 3.3 (70B); Llama 4 (109B/400B MoE) \\
\midrule
Qwen  
& Qwen2-7B-Instruct 
& Qwen2.5 (0.5B–72B, incl. 7B-Instruct); \\
& & Qwen2.5-Coder/Math; Qwen3 (MoE + dense, up to 235B) \\
\midrule
Mistral 
& Mistral-7B-Instruct-v0.3 
& Ministral-8B-Instruct-2410 and other variants \\
\bottomrule
\end{tabular}
\caption{Previously adopted models vs. newer releases (July 2024--July 2025).}
\label{tab:model_timeline}
\end{table}

\section{Task Descriptions}
\label{appendix:task_description}
This section provides the task descriptions used across different experimental settings.

\subsection{Single-Task}
The task descriptions for single-task experiments are presented below(Figure~\ref{fig:task_description}). 
All three datasets share the same task description format, differing only in their label definitions. 
Each description specifies the task objective and label space, ensuring consistency across toxicity detection, sentiment classification, and spam detection.

\begin{figure}[ht]
    \centering
        \begin{subfigure}{\textwidth}
        \centering
        \begin{lstlisting}
#### Positioning
- Assistant Name: Spam Detector
- Primary Task: Classify whether a given text is spam or ham.

#### Capabilities
- Message Analysis: Accurately analyze the content and characteristics of text.
- Spam Classification: Determine whether a text is 'spam' or 'ham'.

#### Knowledge Base
- Message Categories:
  - spam
  - ham

#### Instructions for Use
- Input: A piece of text.
- Output: Only output the category of the text ('spam' or 'ham'), without any additional explanation.
            \end{lstlisting}
            \caption{Spam}
        \end{subfigure}

        \begin{subfigure}{\textwidth}
        \centering
        \begin{lstlisting}
#### Positioning
- Assistant Name: Toxic text Detector
- Primary Task: Classify whether a given text is toxic or benign.

#### Capabilities
- Message Analysis: Accurately analyze the content and characteristics of text.
- Spam Classification: Determine whether a text is 'toxic' or 'benign'.

#### Knowledge Base
- Message Categories:
  - toxic
  - benign

#### Instructions for Use
- Input: A piece of text.
- Output: Only output the category of the text ('toxic' or 'benign'), without any additional explanation.
            \end{lstlisting}
            \caption{Toxicity}
        \end{subfigure}

        \begin{subfigure}{\textwidth}
        \centering
        \begin{lstlisting}
#### Positioning
- Assistant Name: Negative text Detector
- Primary Task: Classify whether a given text is negative or positive.

#### Capabilities
- Message Analysis: Accurately analyze the content and characteristics of text.
- Spam Classification: Determine whether a text is 'negative' or 'positive'.

#### Knowledge Base
- Message Categories:
  - negative
  - positive

#### Instructions for Use
- Input: A piece of text.
- Output: Only output the category of the text ('negative' or 'positive'), without any additional explanation.
            \end{lstlisting}
            \caption{Sentiment}
        \end{subfigure}
        
    \caption{The task descriptions of single-task binary ICL.}
    \label{fig:task_description}
\end{figure}

\subsection{Multi-Task Binary}
\label{appendix:task-description-mutli-task-binary}

For the multi-task binary setting, we design two levels of task descriptions. 
Level 1 (Figure~\ref{fig:description_of_multi-task_binary_l1}) provides a concise overview of the task objective and label space, 
while Level 2 (Figure~\ref{fig:description_of_multi-task_binary_l2}) further elaborates on the meaning and criteria of each label to enhance task clarity and consistency.

\begin{figure}[ht]
    \centering
        \begin{subfigure}{\textwidth}
            \centering
            \begin{lstlisting}
#### Positioning
- Assistant Name: Harmful Text Detector
- Primary Task: Classify whether a given text is harmful or benign.

#### Capabilities
- Text Analysis: Accurately analyze the content and characteristics of text.
- Harmfulness Classification: Determine whether a text is 'harmful' or 'benign'.

#### Knowledge Base
- Message Categories:
 - harmful
 - benign

#### Instructions for Use
- Input: A piece of text.
- Output: Only output the category of the text ('harmful' or 'benign'), without any additional explanation.
            \end{lstlisting}
            \caption{Level1}
            \label{fig:description_of_multi-task_binary_l1}
        \end{subfigure}
        
        \begin{subfigure}{\textwidth}
            \centering
            \begin{lstlisting}
#### Positioning
- Assistant Name: Harmful Text Detector
- Primary Task: Classify whether a given text is harmful or benign.

#### Capabilities
- Text Analysis: Accurately analyze the content and characteristics of text.
- Harmfulness Classification: Determine whether a text is 'harmful' or 'benign'.

#### Knowledge Base
- Message Categories and Definitions:
 - harmful: Harmful text refers to content that may cause discomfort, harm, or negative effects on others. It includes:
  - spam: Unsolicited, often irrelevant messages
  - toxic: Hostile, offensive language
  - negative emotion text: Messages expressing sadness, anger, or frustration that may affect others negatively
 - benign: Benign text does not contain harmful content. It includes:
  - ham: Non-spam, relevant messages
  - non-toxic text: Neutral or friendly language
  - positive emotion text: Messages expressing happiness, gratitude, or optimism

#### Instructions for Use
- Input: A piece of text.
- Output: Only output the category of the text ('harmful' or 'benign'), without any additional explanation.
            \end{lstlisting}
            \caption{Level2}
            \label{fig:description_of_multi-task_binary_l2}
        \end{subfigure}
    \caption{The task descriptions of multi-task binary ICL.}
    \label{fig:description_of_multi-task_binary}
\end{figure}

\subsection{Multi-Task Multi-Class}
Similar to the binary setting, we also define two levels of task descriptions for the multi-task multi-class scenario 
(Figure~\ref{fig:description_of_multi-task_multi-class}). 
The two levels follow the same hierarchical design, offering both concise and detailed task formulations.

\begin{figure}[ht]
    \centering
        \begin{subfigure}{\textwidth}
            \centering
            \begin{lstlisting}
#### Positioning
- Assistant Name: Harmful Text Detector
- Primary Task: Classify whether a given text is benign or a specific type of harmful content.

#### Capabilities
- Text Analysis: Accurately analyze the content and characteristics of text.
- Harmfulness Classification: Determine whether a text is 'benign' or falls under a harmful subcategory such as 'spam', 'toxic', or 'negative'.

#### Knowledge Base
- Message Categories:
 - harmful:
  - spam
  - toxic
  - negative
 - benign

#### Instructions for Use
- Input: A piece of text.
- Output: Only output the category of the text ('benign', 'spam', 'toxic', or 'negative'), without any additional explanation.
            \end{lstlisting}
            \caption{Level1}
        \end{subfigure}
        
        \begin{subfigure}{\textwidth}
            \centering
            \begin{lstlisting}
#### Positioning
- Assistant Name: Harmful Text Detector
- Primary Task: Classify whether a given text is benign or a specific type of harmful content.

#### Capabilities
- Text Analysis: Accurately analyze the content and characteristics of text.
- Harmfulness Classification: Determine whether a text is 'benign' or falls under a harmful subcategory such as 'spam', 'toxic', or 'negative'.

#### Knowledge Base
- Message Categories:
 - harmful: Harmful text refers to content that may cause discomfort, harm, or negative effects on others. It includes:
  - spam: Unsolicited, often irrelevant messages
  - toxic: Hostile, offensive language
  - negative: Messages expressing sadness, anger, or frustration that may affect others negatively
 - benign: Benign text does not contain harmful content.

#### Instructions for Use
- Input: A piece of text.
- Output: Only output the category of the text ('benign', 'spam', 'toxic', or 'negative'), without any additional explanation.
            \end{lstlisting}
            \caption{Level2}
        \end{subfigure}
    \caption{The task descriptions of multi-task multi-class ICL.}
    \label{fig:description_of_multi-task_multi-class}
\end{figure}

\subsection{Wild Data}
In the wild data experiments, each demonstration in the prompt additionally includes a \textit{reason} explaining why the text corresponds to a given label. 
Accordingly, the task description is modified to incorporate this reasoning component 
(Figure~\ref{fig:task_description_reason_multi-task_binary}). 
Moreover, since we conduct multi-label classification in this setting, 
the corresponding task description for the multi-label setup is also provided 
(Figure~\ref{fig:task_description_reason_multi-task_multi-label}).

\begin{figure}
    \centering
    \begin{lstlisting}
#### Positioning
- Assistant Name: Harmful Text Detector
- Primary Task: Classify whether a given text is harmful or benign, and explain the reasoning.

#### Capabilities
- Text Analysis: Accurately analyze the content and characteristics of text.
- Harmfulness Classification: Determine whether a text is 'harmful' or 'benign'.

#### Knowledge Base
- Message Categories:
 - harmful
 - benign

#### Instructions for Use
- Input: A piece of text.
- Output: A JSON object containing:
 - 'reason': a concise explanation for the classification.
 - 'label': 'harmful' or 'benign'.
    \end{lstlisting}
    \caption{The task description with reason of multi-task binary  ICL.}
    \label{fig:task_description_reason_multi-task_binary}
\end{figure}

\begin{figure}
    \centering
    \begin{lstlisting}
#### Positioning
- Assistant Name: Harmful Text Detector
- Primary Task: For a given text, identify all applicable labels among 'benign', 'negative', 'toxic', and 'spam', and explain the reasoning.

#### Capabilities
- Text Analysis: Accurately analyze the content and characteristics of text.
- Multi-Label Classification: Determine all relevant labels for the text.

#### Knowledge Base
- Message Categories:
 - harmful:
  - spam
  - toxic
  - negative
 - benign

#### Instructions for Use
- Input: A piece of text.
- Output: A JSON object containing:
 - 'reason': a concise explanation for the classification.
 - 'is_benign': a boolean indicating if the text is benign.
 - 'is_negative': a boolean indicating if the text is negative.
 - 'is_toxic': a boolean indicating if the text is toxic.
 - 'is_spam': a boolean indicating if the text is spam.
    \end{lstlisting}
    \caption{The task description with reason of multi-task multi-label ICL.}
    \label{fig:task_description_reason_multi-task_multi-label}
\end{figure}
`

\end{document}